\documentclass[opre,nonblindrev]{informs3_modified}
\RequirePackage{tgtermes}
\RequirePackage{newtxtext}
\RequirePackage{newtxmath}
\RequirePackage{bm}
\RequirePackage{endnotes}

\OneAndAHalfSpacedXII

\usepackage{natbib}
 \bibpunct[, ]{(}{)}{,}{a}{}{,}%
 \def\bibfont{\small}%

\TheoremsNumberedThrough
\ECRepeatTheorems

\EquationsNumberedThrough    


\usepackage{graphics}
\usepackage{pgfplots}
\pgfplotsset{compat=1.16}
\usepgfplotslibrary{fillbetween}
\usepackage{endnotes}
\usepackage{xcolor}
\usepackage{booktabs}
\usepackage{enumerate}
\usepackage{commath}
\usepackage{algorithm}
\usepackage{float}
\usepackage{algorithmic}
\usepackage{listings}
\usepackage{cancel}
\usepackage[normalem]{ulem}
\usepackage{thmtools}
\usepackage{thm-restate}
\usepackage{caption}
\usepackage{subcaption}
\usepackage{hyperref}
\usepackage{cleveref}
\let\footnote=\endnote
\usepackage{mathtools}

\DeclarePairedDelimiter{\floor}{\lfloor}{\rfloor}

\usepackage{multicol}
\usepackage{multirow}
\usepackage{clipboard}

\definecolor{blue}{rgb}{0,0,0.9}
\definecolor{red}{rgb}{0.9,0,0}
\definecolor{green}{rgb}{0,0.9,0}

\newcommand{\bs}{\boldsymbol s}

\newcommand{\bx}{\boldsymbol x}
\newcommand{\by}{\boldsymbol y}
\newcommand{\btheta}{\boldsymbol \theta}
\newcommand{\R}{\mathbb R}
\newcommand{\N}{\mathbb N}
\newcommand{\I}{\mathbb I}

\newcommand{\V}{\mathbf V}

\newcommand{\EE}{\mathbb{E}}

\newcommand{\PP}{\mathbb{P}}

\newcommand{\cR}{\mathcal{R}}
\newcommand{\cI}{\mathcal{I}}
\newcommand{\cF}{\mathcal{F}}

\newcommand{\cO}{\mathcal{O}}

\newcommand{\cL}{\mathcal{L}}
\newcommand{\cN}{\mathcal{N}}

\newcommand{\ind}{\mathbf{1}}
\newcommand{\QED}{\ \hfill\rule[-2pt]{6pt}{12pt} \medskip}

\newcommand{\trace}{\text{trace}}
\newcommand{\RR}{\mathbb{R}}

\usepackage{tikz}
\usepackage{amsmath}
\usepackage{algorithm}
\usepackage{algorithmic}
\usepackage{appendix}

\def \E {\mathbb{E}}

\def \R {\mathbb{R}}
\def \N {\mathbb{N}}

\def \ind {\mathbb{I}}

\usepackage{natbib}

\bibpunct[, ]{(}{)}{,}{a}{}{,}%
\def\bibfont{\small}%
\begin{document}
\RUNAUTHOR{}
\RUNTITLE{}
\TITLE{Revenue Maximization and Learning in Product Ranking
}

\ABSTRACT{
We consider the revenue maximization problem for an online retailer who plans to display in order a set of products differing in their prices and qualities.
Consumers have attention spans, i.e., the maximum number of products they are willing to view, and inspect the products sequentially before purchasing a product or leaving the platform empty-handed when the attention span gets exhausted.
Our framework extends the well-known cascade model in two directions: random attention spans of a representative customer are captured and the firm maximizes revenues instead of clicking probabilities.
We show a nested structure of the optimal product ranking as a function of the attention span when the attention span is fixed.
Using this fact, we develop an approximation algorithm when only the distribution of the attention spans is given. Under mild conditions, it achieves $1/e$ of the revenue of the clairvoyant case when the realized attention span is known. We also show that no algorithms can achieve more than 0.5 of the revenue of the same benchmark. The model and the algorithm can be generalized to the ranking problem when consumers make multiple purchases.
When the conditional purchase probabilities are not known and may depend on consumer and product features, we devise an online learning algorithm that achieves $\tilde{\cO}(\sqrt{T})$ regret relative to the approximation algorithm, despite the censoring of information: the attention span of a customer who purchases an item is not observable.
Numerical experiments demonstrate the outstanding performance of the approximation and online learning algorithms.
}

 \ARTICLEAUTHORS{%
 		\AUTHOR{Ningyuan Chen
		\\
		 {\scriptsize Rotman School of Management, University of Toronto,  ningyuan.chen@utoronto.ca}
		 \\
		 Anran Li \\
				 {\scriptsize Department of Decisions, Operations and Technology, The Chinese University of Hong Kong,
anranli@cuhk.hk
}
\\
		 Shuoguang Yang
		 \\
 		{\scriptsize  Department of Industrial Engineering and Decision Analytics, The Hong Kong University of Science and Technology, Clear Water Bay, Hong Kong SAR, China, yangsg@ust.hk, }
		}
 	} 

\maketitle


\section{Introduction}\label{sec:intro}
Online retailing has seen steady growth over the last decade.
According to the data from the US Census Bureau\footnote{\url{https://www.census.gov/retail/ecommerce.html}}, e-commerce sales accounts for 16.4\% of the total sales in the fourth quarter of 2024.
For an e-commerce platform, one of the most important decisions is the products' display positioning as it plays a crucial role in shaping customers' shopping behavior.
Empirical evidence abounds. \citet{JEMS:JEMS234} find that a consumer's likelihood of purchasing from a firm is strongly related to the order in which the firm is listed on a webpage by a search engine. In the online advertising industry, it has been widely observed that ads placed higher on a webpage attract more clicks from consumers \citep{agarwaletal2009,GooseY2009}.
So the products displayed at the top positions can naturally enjoy higher exposure and therefore higher revenue. Given the importance of product ranking positions, the key question for an online platform is how to rank the products to maximize the total revenue.

To answer the question, it is crucial to characterize and quantify how exactly customers react to products ranked in different positions.
As one of the most popular behavioral models for product ranking, the cascade model has proved its salience and robustness in extensive experimental studies.
In the cascade model, a customer with a reserved utility threshold sequentially views the displayed products from top to bottom.
Had the customer viewed an item ranked in a certain position, if the product provides a net utility higher than her threshold,
then the customer purchases the product and leaves right away.
Otherwise, the customer moves on to view the product ranked in the next position.
The model succinctly captures the one-way substitutability top-ranked products impose on bottom-ranked products.
In a large experimental study, \citet{craswell2008experimental} find that the cascade model best explains the consumer behavior among a number of competing hypotheses.
The cascade model captures the well-documented behavior studied in psychology called \emph{satisficing} \citep{simon1955behavioral,caplin2011search}:
The customer views products sequentially from the top to the bottom and purchases a product whenever the utility of the product exceeds an acceptable threshold without ``optimizing''.
The remaining products are skipped even though they may provide higher values.
The empirical success and theoretical tractability have made the cascade model the most widely used model in the literature (see, e.g., \citealt{kempe2008cascade}).

However, the cascade model doesn't provide a sufficient explanation to why a large fraction of customers may leave without purchasing a product in our application.
Indeed, the cascade model is originally motivated by clicks on search results and leaving typically happens when none of the results are attractive enough for a click.
As a result, when using the cascade model for product ranking,
``leaving'' only occurs when a satisfactory product has been found or all products have been exhausted.
In contrast, eye-tracking experiments show that consumers are more likely to examine the items near the top of the list and may leave without purchasing any product before reaching the bottom of the page \citep{FengJ07}.
This phenomenon can be explained by the limited attention span of customers: if a customer arrives with an intrinsic attention span $n$, then she may leave after viewing at most $n$ products regardless of whether a purchase has been made.
According to \citet{FengJ07}, consumers' attention to an item decreases exponentially with its distance to the top, indicating that the distribution of attention span decays rapidly.

In this paper, we introduce limited attention spans to the cascade model.
In our model, the attention span of a customer determines the maximum number of items she is willing to view.
However, for a representative customer, the platform cannot perfectly predict her realization of attention span. Therefore, we capture the limited attention by assigning each customer a random attention span sampled from a distribution. Moreover, although we cannot pin down the realization of attention span for each arrival, we can learn its distribution from past interactions with customers.
When a customer arrives, she sequentially views the displayed products decided by the e-commerce platform (also referred to as the ``firm'' or the ``retailer'') according to the cascade model.
Once the attention span is reached or a satisfactory product is found, the customer leaves the firm.
Thus, in our model, customers leave for two reasons. First, she may purchase a product in the display whose utility exceeds the threshold for the first time and leave, even though there could be another product with a higher utility within the rest of her attention span.
Second, a customer may have exhausted her attention before finding anything satisfactory, in which case she leaves without buying anything or viewing more products.

Within the model, we study the firm's optimal display order of products in order to maximize the potential revenue.
When some of the information is unknown, such as the distribution of the attention span and the attractiveness of the products, we propose a learning algorithm that extracts the information from past observations and maximizes revenues simultaneously.
Our contribution is threefold.
\begin{itemize}
    \item We incorporate attention spans and revenue maximization into the well-known cascade model.
        In the literature on cascade models, the objective of the firm is typically to maximize the purchase rate of customers, which renders the optimal ranking decision trivial. Indeed, the firm only needs to rank the products in the descending order of their attractiveness.
            To maximize the revenue, however, the prices of the products also have to be considered.
    When the attention span is fixed, we develop a dynamic programming algorithm to find the optimal ranking leveraging the fact that the products displayed at the bottom would not cannibalize the demand from the top.
        We show a nested structure of the optimal ranking and the marginal revenue decreases as the attention span increases.
        That is, if we compare the optimal assortment and ranking of $x$ products tailored for customers with an attention span of $x$ against the optimal assortment and ranking of $x+1$ products for customers with an attention span of $x+1$, the assortment for $x$ products is not only a subset of the $x+1$ products but also maintains the same relative order in ranking.
        We also provide sufficient and necessary conditions when the optimal ranking is a prefix as the attention span increases, which is a special case of the nested structure.
    \item Despite the structure of the optimal ranking under fixed attention spans, the computational cost can still be prohibitively high when the attention span is random.
        We develop a novel approximation algorithm for ranking optimization for a representative customer.
        It leverages the concavity of the optimal revenue as the attention span increases, thanks to the nested structure of the optimal ranking.
        When the random attention spans have an increasing failure rate (IFR), an assumption satisfied by many distributions,
        the resulting revenue ratio is $1/e$ relative to a clairvoyant who can access the realized attention span of each customer.
        We show that no ranking can achieve a revenue higher than 1/2 of the same clairvoyant. So the optimality gap of our algorithm is at most $1/2-1/e \approx 0.13.$
    We also look at special cases, for example, when the optimal ranking is a prefix as the attention span increases or when the attention span is geometrically distributed, under which we can provide polynomial-time optimal algorithms. More interestingly, we show that the model and the algorithm can be generalized to the ranking optimization problem when consumers make multiple purchases in a single visit.
    \item When the customers' behavior is not known a priori, including the distribution of the attention spans and the attractiveness of the products, the firm may have to learn the unknown information and maximize the revenue simultaneously.
        What makes the problem challenging is the presence of \emph{consumer features} and \emph{product features},
        two defining characteristics of online retailing.
        We consider the interaction (outer product) between both features that determines the product attractiveness.
        In addition, when customers leave the firm after making a purchase, the attention span information is censored.
        We propose an online learning algorithm by constructing unbiased estimators for the distribution of the attention span and combining them with the UCB approach to learn the product attractiveness.
        It achieves good performance ($\tilde{\cO}(\sqrt{T})$ regret) relative to the offline algorithm.
\end{itemize}

\section{Literature Review}\label{sec:literature}
There are two streams of literature that are closely related to this study:
(1) choice models (especially those taking the positions of products into account) and the corresponding optimal assortment planning, (2) the online learning of product ranking models.
We review the two streams separately below.

\subsection{Consumer Choice Models and Assortment Optimization}\label{sec:literature-choice-model}
Most classic discrete choice models do not take the position effects of products into account.
The assortment optimization problem determines the optimal subset of products to include in the assortment to maximize the revenue.
For example, \citet{MahajanV2001,TalluriV2004} show that under the well-known multinomial logit (MNL) model, the optimal assortment includes the products with the highest prices.
Since this paper studies product ranking, the optimization involves the display order of products.
It is significantly different from this line of literature in terms of models and algorithms.

Recently, there have been a number of studies focusing on the business case when consumers may inspect products sequentially or based on their display positions, which is more relevant in online retailing.
Next we compare our paper with these studies from the angles of model setups and results.
\citet{abeliuk2016assortment} incorporate the position bias into the deterministic utilities of the MNL model.
Their model describes the choice behavior in the population but does not provide a sequential choice model for individual customers.
In \citet{guillermoAnran20}, a customer arrives with a random attention span (similar to our model).
She then chooses the preferred product among \emph{all} the products in the span according to a choice model.
This is in contrast to our model in which customers choose products sequentially and do not \emph{recall} past products.
Therefore, the optimal assortment in \citet{guillermoAnran20} does not exhibit the nested property (Theorem~\ref{prop:nested}).
The model in \citet{alidannymnl2020} is similar to \citet{guillermoAnran20}, but focuses on the MNL choice model.
In \citet{Derakhshan2020}, the customers search products sequentially.
They first form their consideration set using an optimal stopping problem to maximize the expected surplus (due to the existence of search cost) and then choose the product inside the consideration set using the MNL model.
This model doesn't consider attention spans and focuses on the maximization of market share instead of revenues.
In \citet{chen2020position}, the authors model this problem using mechanism design when the sellers of the products have private information.
A few recent papers extend discrete choice models to multiple stages.
\citet{FLORES20191052,jakedanny19,liunanast20,gaoMa2020seq} extend the MNL model to multiple stages.
In each stage, the customer chooses a product based on the MNL model and leaves the market if a product is chosen; otherwise, the customer moves on to the next stage.
The optimization problem is usually NP-hard and the authors give approximation methods.
Although the model in this study can be treated as a special case of the models studied in these papers when each stage only contains one product, there are two important differences.
First, random attention span of a representative customer is not considered in those papers.
Second, we are able to derive richer structures and constant-bound approximation algorithms specific to our setting.
Recently, \cite{brubach2021follow} study a similar problem in the online matching framework.
They design an approximate algorithm using linear programming that obtains $1/2$-approximation relative to the LP upper bound of the optimal revenue under random attention span.
They also show that any algorithm can be arbitrarily bad relative to the \emph{clairvoyant} that foresees the realized attention span, indicating that the clairvoyant revenue is an upper bound of the LP formulation.
Our focus is on the nested structure of the optimal ranking for fixed attention spans, which allows us to drive a different approximation algorithm that achieves $1/e$-approximation relative to the \emph{clairvoyant} instead of the LP upper bound under the mild IFR assumption of attention span distribution.
Moreover, our algorithm can be generalized to the purchase of multiple products (Section~\ref{sec:multiple-purchase} in the Appendix), which has not been studied in the display optimization literature.
In fact, the well-known cascade model with position dependent multipliers (CMPDM) in \citet{kempe2008cascade} can be recast into the revenue maximization problem of product ranking we study.
They provide a fully-polynomial-time approximation scheme (FPTAS) with a $(1/4 - \epsilon)$-approximation guarantee relative to the optimal expected revenue. While we derive some structural properties for the optimal ranking and propose a more efficient algorithm with a better $1/e$ approximation guarantee relative to a more stringent benchmark, i.e, a clairvoyant who
can access the realized attention span of each customer, with the additional mild IFR assumption on the distribution of attention span.
This setup and derived results are novel in the literature on cascade models.
From the perspective of algorithmic design, the proposed approximation algorithm is primarily inspired by the nested structure (Theorem~\ref{prop:nested}).
This structure and algorithm are not known in the papers mentioned above.

The random attention span considered in this study is analogous to similar behavioral setups in the literature.
The primary motivation is the cognitive cost for customers to view and evaluate a large number of products in online retailing.
\citet{wangsahin17} build a model where the customers first form a consideration set by weighing between their expected utility and the search effort. The products in the consideration set are then thoroughly evaluated.
\citet{gallegoli17} study the operational decisions of a model with random consideration sets where customers have a fixed preference order and each product enters the consideration set with an exogenous probability, similar to the conditional purchase probability in the cascade model.
\citet{ali2019clickmnl} relax the fixed preference order assumption of the random consideration set model and focus on an assortment setting.

\subsection{Online Learning}
The multi-armed bandit framework has seen great success in modeling the exploration/exploitation trade-off for many applications.
See \cite{bubeck2012regret} for a comprehensive survey.
In particular, in the field of revenue management, there have been many studies on demand learning and price experimentation using this framework, since the early seminal works by \citet{araman2009dynamic,besbes2009dynamic,broder2012dynamic}.
Many extensions and new features have been studied, such as network revenue management \citep{besbes2012blind,ferreira2018online,chen2019nonparametric}, personalized dynamic pricing \citep{chen2018nonparametric,chen2018primal,miao2019context,ban2020personalized}, limited price experimentations \citep{cheung2017dynamic}, and non-stationarity \citep{besbes2015non}.
Readers may refer to \citet{den2015dynamic} for a review of papers in this area.

There are several papers applying the multi-armed bandit framework to assortment planning, such as \citet{rusmevichientong2010dynamic,saure2013optimal}.
\citet{agrawal2019mnl} consider the learning problem in the well-known MNL choice model under general assumptions compared to the literature. Their algorithm simultaneously explores the optimal assortment and attempts to maximize the revenue.
\citet{kallus2020dynamic} investigate a similar problem under the contextual setting, when customers of different types have various preferences.
They exploit the low-rank nature of the parameters to design efficient algorithms.
In a related paper, \citet{chen2018dynamic} assume that the contextual information can be non-stationary over time.
\citet{oh2019thompson} leverage Thompson sampling to learn the optimal assortment in the MNL model.
The methods used in the literature cannot be directly applied to our model, because customers have dramatically different behavior in a product ranking model from the MNL model.
Therefore, our algorithm differs in what to learn and how to actively experiment.

Our work is closely related to the papers studying the learning of the cascade model and product ranking models.
The first such algorithm, called ``cascading bandits'', is proposed in \cite{kveton2015cascading}.
It learns the optimal set of items to display in order to maximize the click rate,
when the firm can observe the clicking behavior of past customers.
Due to the wide range of applications in online advertising and recommendation systems of the cascade model,
the online learning algorithm for such models under generic settings and contextual information has been extensively studied by \cite{katariya2016dcm,zoghi2017online,lagree2016multiple,kveton2015combinatorial} and \cite{zong2016cascading,cheung2019thompson}.

Compared to cascading bandits, our work has three major differences in problem formulations, which require novel treatments when designing algorithms.
First, the items in the cascade model are only differentiated by their clicking probabilities.
In our setting, the items generate different revenues.
Second, the customers in the cascade model never abandon the system early, while in our model, we focus on capturing a representative customer who has a random attention span and its distribution needs to be learned.
These two changes make our problem much more challenging.
In the cascade model, the offline optimal solution is rather straightforward: finding the set of items with the highest clicking rates.
In our model, it is not clear how to solve the offline problem exactly.
Third, when a customer purchases an item and leaves the system before reaching the limit of the attention span, this information is censored to the firm.
We develop novel unbiased estimators to handle the censored observations.

Recently, the learning of product ranking models that generalize the cascade model has drawn attention in the Operations Research and Operations Management community.
\citet{ferreira2019learning} study a model in which the attention span and the clicking probabilities can be correlated.
In this dimension, their model is more general than ours.
They focus on maximizing the probability that a customer clicks at least one item.
As a result, a greedy policy that ranks products sequentially is $1/2$-optimal.
The greedy policy is their learning target, whose structure can be exploited.
In contrast, because we focus on revenue maximization, the target policy doesn't have a structure to facilitate learning.
\citet{golrezaei2021learning} propose an algorithm that learns the optimal ranking in the presence of fake users.
\citet{gao2018joint} investigate the optimal pricing of the cascade model, in which the clicking and purchasing probabilities are parametrized and need to be learned.
Their model is more structured and the nested structure doesn't seem to hold in their model.
Their major focus is on pricing and the design of the algorithm deviates significantly from ours.
\citet{niazadeh2020online} consider the online learning problem of \citet{asadpour2020ranking} and focus on the maximization of the sum of a sequence of submodular functions, subject to a permutation of the inputs.
It is closely connected to product ranking.
However, because we focus on revenue maximization instead of clicking rates, the objective function is no longer submodular.
A recent paper \citep*{zhang2022ordering} consider the joint problem of inventory and ranking under a click-then-convert model.
The online learning portion of our model is similar to \citet{cao2019dynamic,cao2019sequential}.
The major distinction is our focus on maximizing the revenue instead of clicks.
As mentioned above, the difference in the objective leads to completely different offline optimal solutions.
Therefore, the resulting learning algorithm and the analysis differ.

\section{The Model}\label{sec:model}
Let $[n]\triangleq \{1,\ldots,n\}$ denote the (unordered) set of products. There is an online retailer (he) who chooses an assortment $S\subseteq [n]$ to display and ranks them in order within $M$ slots.
We use $\sigma\in  P(S)$ to refer to a permutation $\sigma$ of $S$, where $P(S)$ is the set of all permutations of items in $S$.
More precisely, for $i\in \left\{1,\dots,|S|\right\}$, we use $\sigma(i)\in S$ to denote the product displayed in the $i$-th position.\footnote{We use $|S|$ ($|\sigma|$) to represent the number of products in the assortment (ranking).}
Equivalently, for a product $j\in S$, the firm displays it in the $\sigma^{-1}(j)$-th position.

When a representative consumer (she) arrives, she views the products sequentially.
However, the retailer cannot perfectly predict how many products she is willing to view.
Therefore, we capture it by a random attention span $X$, with distribution $g_x \triangleq \Pr(X=x)$ and tail probability denoted by $G_x\triangleq\Pr(X\ge x)$ for $x=1,2,\dots$.
For a consumer with fixed attention span $x$ drawn from $X$, she views product $\sigma(1)$ first and purchases it if it is satisfactory, i.e., the utility exceeds the no-purchase option.
If she purchases the first product, then she leaves and the firm garners revenue $r_{\sigma(1)}$.
Otherwise, she moves on to the second product and repeats the process until she finds a product satisfactory or has viewed $x$ products, after which she leaves permanently.
We assume that the probability of product $j$ being satisfactory is independent of everything else and
denote it by $\lambda_j$, i.e., conditional on viewing product $j$, the probability that she would purchase it. For the rest of the paper, we refer to it as conditional purchase probability.
The consumer keeps viewing the products till she finds one satisfactory or exhausts her attention span $x$, at which point she leaves without purchasing anything.

Therefore, each product $j\in [n]$ is associated with a conditional purchase probability $\lambda_j$ and price $r_j$.\footnote{We assume $\lambda_j\in(0,1]$ and $r_j>0$ for all $j$ because removing products with zero conditional purchase probability or zero price from the list does not affect the results.}
Both quantities are assumed to be fixed and known in the offline setting. (In Section~\ref{sec:learning}, we consider the case when $\lambda_j$ is unknown and to be learned, which may depend on customer/product features.)
We index the products such that
\begin{equation}\label{eq:product-indexing}
    r_1\ge r_2\ge \dots\ge r_n; \quad \lambda_j>\lambda_{j+1} \ \text{if}\   r_j=r_{j+1}.
\end{equation}
In other words, the products are indexed in the descending order of their prices and then in the descending order of their conditional purchase probabilities if the prices are equal.
We assume no products have identical characteristics (the combination of price and conditional purchase probability).

Given that position $k$ is within a consumer's attention span, her purchase probability of product $\sigma(k)$ is
$\prod_{i=1}^{k-1} (1-\lambda_{\sigma(i)} ) \cdot \lambda_{\sigma(k)}$.
 In other words, the consumer purchases product $\sigma(k)$ if it is satisfactory while all products displayed earlier are not. $1-\lambda_{\sigma(i)}$ is the cannibalization effect that a product at position $i$ exerts on the products displayed later.

\subsection{Revenue Maximization for the Retailer}
The firm's goal is to choose an assortment $S$ of at most $M$ products, as well as a ranking $\sigma\in P(S)$ of the assortment, in order to maximize the expected revenue garnered from a representative consumer.
From the earlier introduction we see that the expected revenue from a consumer with a fixed attention span $x$ can be expressed as
\begin{equation}\label{eq:fix-span-revenue}
    R(\sigma,x) \triangleq \sum_{k=1}^{x\wedge M} \prod_{i=1}^{k-1} (1-\lambda_{\sigma(i)} ) \cdot \lambda_{\sigma(k)} r_{\sigma(k)}.
\end{equation}

When a representative customer arrives, as the retailer does not know her attention span, he takes the expected value of $x\sim X$ in \eqref{eq:fix-span-revenue} and obtains the total expected revenue
\begin{align}
& \EE[R(\sigma, X)]= 
\sum_{x=1}^M  g_x R(\sigma,x)  \nonumber \\
& = \sum_{x=1}^M  g_x\Big ( \sum_{k=1}^{x} \prod_{i=1}^{k-1} (1-\lambda_{\sigma(i)} ) \cdot \lambda_{\sigma(k)} r_{\sigma(k)} \Big ) \notag \\
& =  \sum_{x=1}^{M} \prod_{i=1}^{x-1} (1-\lambda_{\sigma(i)} ) \cdot \lambda_{\sigma(x)} r_{\sigma(x)}  G_x. \label{eq:opt_random}
\end{align}
Therefore, the optimization problem for the firm is the joint assortment, i.e., choosing an $S \subset N$ such that $|S| \leq M$, and ranking, i.e., deciding $\sigma \in P(S)$ to maximize the total expected revenue from a random customer:
\begin{equation}\label{eq:random-span-revenue}
    \max_{S \subset N, \sigma \in P(S) } \, \EE[ R(\sigma,X)].
\end{equation}
From (\ref{eq:fix-span-revenue}) we can see that a product displayed at a later slot does not cannibalize the demand and therefore the revenue for the earlier.
As a result, the optimal ranking would occupy all the $M$ slots and we only focus on assortments such that $|S|=M$.

Note that one customer purchasing at most one product is a defining feature of discrete choice models. And for the cascade model,
    it is also primarily motivated by clicking one relevant item from online searches.
    Although the single item clicking/purchasing is widely adopted in the literature, the assumption may be limiting in online retailing:
    for example, a customer may add several products to the cart on Amazon before checking out.
    In Section~\ref{sec:multiple-purchase} in the Appendix, by relaxing the cannibalization factor to a broader range, we extend the model to capture multiple purchases.
    That is, a customer may continue viewing and selecting the products after making some purchases.
We show that most results derived in the next section still hold in that setting. In this sense, our work is more general than any of the current discrete choice model based display optimization literature.

Note that if $r_j\equiv 1$ for all $j\in[n]$, then our model encompasses the well-known classic cascade model  \citep{craswell2008experimental} as a special case.
In this case, the online retailer tries to maximize the click rate, or equivalently, the purchasing probability of at least one product.
For the cascade model, the optimal solution follows an intuitive structure: ranking the products in the descending order of their conditional purchase probabilities $\lambda_j$.
In the general case when $r_j$'s are not identical, the products are differentiated along two dimensions, and it is unclear how to trade off between their conditional purchase probabilities and prices.
For example, it is natural to rank a product with high conditional purchase probability and high price in the top position.
However, in most practical scenarios, a product with higher conditional purchase probability is often associated with a lower price.
This presents a dilemma: should a retailer prioritize displaying products that attract more customers or those that yield higher profits? This fundamental tension is at the heart of any assortment or ranking optimization challenge. The resolution of this tension is further influenced by the factors such as the available display capacity $M$ and the distribution of customers' attention spans.
We present the following example to shed some light on the complexity of these questions.
\begin{example}\label{exp:random-span}
Suppose the online retailer is selling three products, with revenues and conditional purchase probability $(r_j,\lambda_j)$ as: $(1,1)$, $(9,0.1)$ and $(1.9,0.52)$.
The capacity is $M=2$.
When the customer only views one product ($X\equiv 1$), the optimal ranking is to display product one, which has the highest conditional purchase probability, in the first position.
So an optimal ranking is $\sigma^1 = \{ 1\}$ and $R(\sigma^1,1) = 1$.

Now suppose the customer always views two products ($X\equiv 2$).
As the customer is willing to view more positions, the retailer would prefer to place a product with higher profit first, though it may have a lower conditional purchase probability.
So an optimal ranking is $\sigma^2 = \{ 2,1\}$ and $R(\sigma^2, 2) = 1.8$.

Next consider the case when the attention span is random.
When $\Pr(X=1) = 0.9$ and $\Pr(X=2) = 0.1$, it can be shown that the optimal ranking $\sigma^*$ is neither $\sigma^1 $ nor $\sigma^2$.
In fact, we have $\EE[R(\sigma^1,X)] = R(\sigma^1,1) = 1$ and $\EE[R(\sigma^2,X)] = 0.9 \cdot 9\cdot 0.1 + 0.1\cdot  R(\sigma^2,2) =  0.99$.
On the other hand, one can show that $\sigma^* = \{3,1\} $ generates
$
\EE[R(\sigma^*, X)] = 0.9\cdot  1.9\cdot 0.52 + 0.1 \cdot (1.9\cdot 0.52+0.48
\cdot 1\cdot 1)= 1.036$.
Surprisingly, product three never appears in the optimal rankings for fixed attention spans, but should be placed on the top when the attention span is random.
\hfill\QED 
\end{example}
Example~\ref{exp:random-span} demonstrates the intricacy of the optimal ranking for random attention spans.
A natural question, then, is whether the optimal ranking can be efficiently solved.
It turns out that this is a notoriously difficult problem:
It shares a similar formulation to the CMPDM model in \citet{kempe2008cascade}, which claims that it is an open question to show the NP-hardness and provide a fully-polynomial-time approximation scheme (FPTAS) with $(1/4-\epsilon)$-approximation guarantee of the optimal expected revenue.
We devote the next section to a new approximation algorithm that improves the approximation ratio against a more stringent benchmark (the clairvoyant revenue) to $1/e$ under a wide class of distributions of $X$.


\section{Best-x Algorithm for Optimal Ranking}\label{sec:app-alg}
With the complexity of the problem, we focus on developing a fast and easy-to-implement approximation algorithm for \eqref{eq:random-span-revenue} and providing a performance guarantee of the garnered revenue.
Before presenting our approximation algorithm, we first investigate the revenue maximization problem when the attention span is fixed.
It turns out that the problem has rich structures, based on which we then develop an approximation algorithm for random spans.

\subsection{Revenue Maximization for Fixed Attention Spans}\label{sec:fix-span}
Suppose the retailer is expecting a customer with fixed attention span $x\le M$.
Recall that $M$ is the number of available slots so for $x>M$ we can truncate it to $x=M$.
To maximize the revenue, the retailer would choose an assortment of size $|S|=x$.
We first provide a lemma characterizing the optimal ranking when the assortment $S$ is given.
\begin{lemma}\label{lem:fix-m-decrease-r}
    Fix attention span at $x$. Suppose the assortment $S$ with size $|S|=x$ is given and $\sigma^x$ maximizes $R(\sigma,x)$ among $\sigma\in P(S)$.
    We have that the products in $\sigma^{x}$ are displayed in increasing order of their product indices, i.e., $\sigma^{x}(i) \leq \sigma^{x}(i+1)$ for $i=1,\dots,|S|-1$.
\end{lemma}
Combined with condition \eqref{eq:product-indexing}, Lemma~\ref{lem:fix-m-decrease-r} implies that the products in the optimal ranking are displayed in the descending order of their prices and then in the descending order of their conditional purchase probabilities if there are ties in prices.
Lemma~\ref{lem:fix-m-decrease-r} looks counter-intuitive at first sight as it prioritizes the price over the conditional purchase probability.
To understand the intuition, note that we focus on a customer with a fixed attention span $x$, i.e., she would always view $x$ products or purchase the first satisfactory product.
Given an assortment and ranking, by swapping expensive products to the top, the customer would view expensive products first, which increases the expected revenue.

Although Lemma~\ref{lem:fix-m-decrease-r} reduces the problem of finding the optimal ranking under a fixed attention span to finding the optimal assortment, it is still a daunting task.
Namely, the number of assortments with given size $M$ is exponential in $M$.
It is computationally intractable to compare the expected revenues of all possible assortments/rankings and pick the optimal one. However, in light of Lemma~\ref{lem:fix-m-decrease-r}, we can develop a dynamic program to quickly solve the joint assortment/ranking problem. The dynamic program employs both the optimal ranking result from Lemma~\ref{lem:fix-m-decrease-r} and the fact that the revenue $R(\sigma,x)$ can be computed recursively, i.e.,
\begin{align}
   &  R(\sigma,x)  \triangleq \sum_{k=1}^{x} \prod_{i=1}^{k-1} (1-\lambda_{\sigma(i)} ) \cdot \lambda_{\sigma(k)} r_{\sigma(k)} \nonumber \\
 &=  (1- \lambda_{\sigma(1)}) \big( \sum_{k=2}^{x} \prod_{i=2}^{k-1} (1-\lambda_{\sigma(i)} ) \cdot \lambda_{\sigma(k)} r_{\sigma(k)} \big) \nonumber 
 \\
 & \quad +  \lambda_{\sigma(1)} r_{\sigma(1)} \nonumber \\
 &= \lambda_{\sigma(1)} r_{\sigma(1)}+(1- \lambda_{\sigma(1)})R(\sigma[2,x],x-1) \label{eq:rec3}\\
 &= R(\sigma[2,x],x-1) \nonumber 
 \\
 & \quad +\lambda_{\sigma(1)} \big( r_{\sigma(1)}-R(\sigma[2,x],x-1)\big), \nonumber
\end{align}
where we use $\sigma[2,x]$ to denote the sub-ranking of $\sigma$ from position $2$ to $x$.
Note that $R(\sigma[2,x],x-1)$ is the revenue from the display $\sigma[2,x]$ when the consumer has attention span $x-1$.
Intuitively, by the third equality in \eqref{eq:rec3}, had we fixed the product at the top position, then the optimal sub-ranking $\sigma[2,x]$ should be chosen to maximize $R(\sigma[2,x],x-1)$.
Moreover, according to Lemma~\ref{lem:fix-m-decrease-r}, if the top displayed product is $j$, the indices of the rest of the products in $\sigma[2,x]$ should not be smaller than $j$ and their prices are less than $r_j$.
This structure inspires Algorithm~\ref{AST:m} where we leverage dynamic programming and iteratively update two state variables: $x$--length of attention span and $j$--index of the most expensive product we can offer. At every state $(x,j)$, we compute $H_j^x$ and $\sigma_j^x$, the optimal revenue and ranking for customers with attention span $x$, while the products are chosen from $\{j,j+1,\dots,n\}$. When we move to state $(x,j-1)$, we know the optimal ranking should be either offering product $j-1$ at the top and then followed by $\sigma_j^{x-1}$, or excluding product $j-1$ and remaining with offering $\sigma_j^{x}$, whichever returns higher revenue. Now we are ready to present Algorithm~\ref{AST:m}, which can find an optimal assortment/ranking in $O(nx)$ computations for a customer with a fixed attention span $x$. Therefore, it would take $O(nM^2)$ computations to find optimal rankings for all customers $x \in [M]$.

\begin{algorithm}[t]\caption{Assortment Optimization Given $X=x$}\label{AST:m}
\begin{algorithmic}
\REQUIRE{$r_j$, $\lambda_j$ for $j\in [n]$, $x$}
\STATE $H^0_{j} = 0$ for all $j = 1,2,\dots,n$; $H^k_{n+1} =0$ for all $k=0,1,2,\dots,x$; $\sigma^0_j =\emptyset$ for all $j = 1,2,\dots,n$; $\sigma^k_{n+1} =\emptyset$ for all $k=0,1,2,\dots,x$
\FOR{$k = 1, \ldots, x$ }
   \FOR{$j = n, \ldots, 1$ }
   \STATE \begin{equation} \label{eq:dp_recursion}
   H^k_j\gets \max \Big \{ H^{k-1}_{j+1} +\lambda_{j}\big(r_{j}-H^{k-1}_{j+1}\big) , \, H^k_{j+1} \Big \}
   \end{equation}
    \IF {$ \Big( H^{k-1}_{j+1} +\lambda_{j}\big(r_{j}-H^{k-1}_{j+1}\big) \Big) \geq H^k_{j+1}$}
   \STATE $\quad \sigma^k_j\gets\sigma^{k-1}_{j+1} \cup \{j\}$
    \ELSE
    \STATE  $\quad \sigma^k_j\gets\sigma^k_{j+1}$
    \ENDIF
   \ENDFOR
\ENDFOR
\RETURN $H^{x}_1$ and $\sigma^x_1$
\end{algorithmic}
\end{algorithm}

With a slight abuse of notations, when $\sigma^k_j$ is optimized to be $\sigma^{k-1}_{j+1} \cup \{j\}$, Lemma~\ref{lem:fix-m-decrease-r} uniquely determines its ranking order.
Although there might be multiple optimal rankings, Algorithm~\ref{AST:m} always outputs one with a specific property.
In particular, for two rankings $|\sigma|=|\sigma'|=x$, we say $\sigma$ is lexicographically less than $\sigma'$ if for all $i=1,\dots,x$, $\sigma(i)\leq \sigma'(i)$.
The next result shows that Algorithm~\ref{AST:m} always finds the optimal ranking with the least lexicographic value (referred to as the $\cL$-optimal ranking) for attention span $x$, denoted as $\sigma^x$.
\begin{proposition}\label{prop:dp}
For a customer with fixed attention span $x$, Algorithm~\ref{AST:m} returns the $\cL$-optimal ranking.
\end{proposition}

The property of $\cL$-optimal rankings plays an important role in the analysis, which can show a \emph{nested structure} of the optimal ranking.
\begin{restatable}{definition}{defnested}
\label{def:nested}
 Two rankings $\sigma$ and $\sigma'$ are said to have a \emph{nested structure}, or, $\sigma\subset \sigma'$, if for all $i \in \sigma$ there exists an $i' \in \sigma'$ such that $\sigma(i)= \sigma'(i')$ and for all $i,\, i'=1,\dots,|\sigma|-1$, if $\sigma(i)<\sigma(i')$, then $\sigma'(i)<\sigma'(i')$.
In other words, $\sigma\subset\sigma'$ if the assortment of $\sigma$ is a subset of that of $\sigma'$ and the products are ranked in the same order.
\end{restatable}

The next proposition demonstrates a nested structure for the $\cL$-optimal rankings as the customer's fixed attention span increases. 
Moreover, with the nested structure, we are able to design a greedy based algorithm that can find $\sigma^x$ for all $x \in [M]$ in just $O(nM)$ computations, significantly faster than Algorithm~\ref{AST:m}.
\begin{proposition}\label{prop:nested}
The $\cL$-optimal rankings have a nested structure:
\begin{equation}
    \sigma^1 \subset \sigma^2 \subset \dots \subset \sigma^M.
\end{equation}
\end{proposition}

Proposition~\ref{prop:nested} states that the optimal ranking for customers of attention span $x+1$ can be obtained by inserting or appending one product to the optimal ranking for attention span $x$.
For example, suppose there are 5 products and the $\cL$-optimal ranking for $x=3$ is $\sigma^3=\{1,3,4\}$.
Then the $\cL$-optimal ranking for $x=4$ might be $\sigma^4=\{1,2,3,4\}$ or $\sigma^4=\{1,3,4,5\}$.

Proposition~\ref{prop:nested} naturally leads to a simple algorithm to sequentially compute $\sigma^x$ for all $x\in [M]$.
In particular, Algorithm~\ref{alg:astopt2} provides an iterative approach to compute the $\cL$-optimal assortment.
Note that we only need to find the optimal assortment, and the ranking is automatically determined by Lemma~\ref{lem:fix-m-decrease-r}.
\begin{algorithm}[t]\caption{Assortment Optimization for Fixed Attention Spans (AssortOpt)}\label{alg:astopt2}
\begin{algorithmic}
\REQUIRE{$r_j$, $\lambda_j$ for $j\in [n]$}
\STATE { $\sigma^0 \gets \emptyset $}
\FOR{$x = 1, \dots, M$ }
\STATE Update $\sigma^x \gets \argmax_{\sigma=\sigma^{x-1}\cup \{j\} } R(\sigma,x)$
\ENDFOR
\RETURN $\{\sigma^x \}_{x=1}^M$
\end{algorithmic}
\end{algorithm}

Based on Proposition~\ref{prop:nested}, we are able to show the following structural property, which serves as a building block for the approximation algorithm.
In particular, we can show that the marginal revenue from increasing the attention span is diminishing:
\begin{theorem}\label{thm:decrease_margin}
The optimal revenues for fixed attention spans satisfy
    $$
R(\sigma^{x+1},x+1) - R(\sigma^{x},x) \leq R(\sigma^x,x)- R(\sigma^{x-1},x-1).
$$
\end{theorem}
Proposition~\ref{prop:nested} and Theorem~\ref{thm:decrease_margin} characterize the optimal ranking and expected revenues when customers have fixed attention spans.
Next, we use the properties to develop an approximation algorithm for a representative customer with random attention spans.

\subsection{Random Attention Spans: The Best-x Algorithm and $1/e$ Approximation Ratio}\label{sec:sub-approx}
For a representative customer, the attention span is random. As a result, the optimal ranking doesn't necessarily have the fixed ordering property in Lemma~\ref{lem:fix-m-decrease-r} (see Example~\ref{exp:random-span}) and dynamic programming cannot be applied.
However, the optimal rankings for fixed attention spans lend us the following intuition:
if we choose a proper $x$ and use the optimal ranking $\sigma^x$ for customers with fixed attention span $x$, then how does it perform when the attention span $X$ is actually random?
To analyze the performance of such an algorithm,
note that for a given ranking, the expected revenue is non-decreasing for consumers with longer attention spans.
Therefore, if we use the optimal ranking $\sigma^x$ developed for customers with fixed attention span $x$,
then for customers with attention span $X\ge x$, the expected revenue is at least $R(\sigma^x, x)$.
Moreover, the fraction of customers with attention spans no less than $x$ is given by $G_x= \Pr(X\ge x)$.
Therefore, a lower bound for the expected revenue when applying the ranking $\sigma^{x}$ for customers with random attention spans is $R(\sigma^x, x) G_x$.
Because of this observation, we proceed to designing an approximation algorithm to maximize this lower bound:
\begin{equation}\label{eq:app-alg}
    \argmax_{\sigma^x:1\le x\le M} R(\sigma^x, x) G_x.
\end{equation}
The algorithm, referred to as the Best-x Algorithm, is summarized in Algorithm~\ref{alg:approx}.
\begin{algorithm}[t]\caption{The Best-x Algorithm}\label{alg:approx}
\begin{algorithmic}
\REQUIRE{ $r_j$ and $\lambda_j$ for $j\in[n]$, $G_x$ for $x\in[M]$}
  \STATE{  Compute $\{R(\sigma^x,x) \}_{x=1}^M= \text{AssortOpt}$}
  \RETURN{ $\sigma = \argmax\limits_{\sigma^x:1\le x\le M} R(\sigma^x,x) G_x$}
\end{algorithmic}
\end{algorithm}

To analyze the performance of the Best-x Algorithm, we impose the following assumption on the distribution of consumers' attention spans:
\begin{assumption}\label{assump:IFR}
The distribution of the attention span $X$ has increasing failure rate (IFR).
That is,
$$
\frac{g_x}{G_x} \leq \frac{g_{x+1}}{G_{x+1}} \iff G_{x+1} G_{x-1} \leq G_x^2,\,\,\,\text{ for } x= 2,3,\cdots, M-1.
$$
\end{assumption}
Note that many common distributions used in practice have IFR, including the exponential distribution, the geometric distribution, the normal distribution, the uniform distribution and the negative binomial distribution \citep{rinne2014hazard}.
Therefore, Assumption~\ref{assump:IFR} does not significantly limit the generality of our results.
Moreover, empirical evidences \citep{FengJ07} show that consumers' attention to an item decreases exponentially with its distance to the top, which is consistent with the IFR~assumption.

We aim to show the ranking returned from the Best-x algorithm can guarantee an \emph{approximation ratio} of the optimal revenue.
That is, if
$$\frac{\max_x \EE(R(\sigma^x,X))}{\max_{\sigma} \EE(R(\sigma,X))} \geq \alpha,$$
for some $\alpha\in [0,1]$ under all possible inputs --- conditional purchase probability $\lambda$, product revenue $r$ and attention span distribution $G$ ---
then we may claim that the Best-x Algorithm achieves approximation ratio $\alpha$.

However, directly comparing with the optimal expected revenue is difficult as we do not know the value of $\max_{\sigma} \EE(R(\sigma,X))$.
Therefore, we first provide an intuitive upper bound for the optimal revenue based on a \emph{clairvoyant}.
We then provide an approximation ratio relative to the upper bound.

\paragraph{Clairvoyant upper bound for the expected revenue.}
Recall the definition of $\sigma^x$ in Section~\ref{sec:fix-span}: the optimal ranking for the customer with fixed attention span $x$.
Note that for $X=x$, the revenue of any ranking $\sigma$ is dominated by $\sigma^x$, i.e., $R(\sigma,x)\le R(\sigma^x,x)$.
Therefore, we have for any $\sigma$:
\begin{align*}
& \EE[ R(\sigma,X)] =  \sum_{x=1}^M g_x R(\sigma, x) 
\\
&  \leq \sum_{x=1}^M g_x R(\sigma^x, x)= \EE[R(\sigma^X,X)].
\end{align*}
Taking the maximum over $\sigma$ on the left-hand side, it leads to the following upper bound:
\begin{proposition} \label{prop:upper-bound}
    The optimal expected revenue \eqref{eq:random-span-revenue} is upper bounded by the expected clairvoyant revenue, where we can personalize the recommendation for each customer with realized attention span, i.e.,
    \begin{equation}\label{eq:UB}
\sum_{x=1}^M g_x R(\sigma^x,x).
\end{equation}
\end{proposition}
Note that the upper bound can be interpreted as follows: if the retailer is a clairvoyant, i.e., it can access the realized attention span $X=x$ and provide a customized ranking for customers with attention span $x$, then the optimal revenue is given in \eqref{eq:UB}.
Apparently, the expected revenue obtained from any ranking is upper bounded by that obtained by the clairvoyant.

Next we show that our Best-x algorithm can guarantee a $1/e$ approximation ratio relative to the upper bound \eqref{eq:UB} under the mild IFR Assumption~\ref{assump:IFR}.
For any fixed $x$, because $\EE[R(\sigma^x,X)| X\geq x]\ge R(\sigma^x, x)$, we have
\begin{align*}
    \EE[ R(\sigma^x,X)] &= \EE[R(\sigma^x,X)| X< x]\Pr(X< x) \nonumber 
    \\
    & \quad  + \EE[R(\sigma^x,X)| X\geq x]\Pr(X\geq x)\\
                      &\geq R(\sigma^x, x) G_x.
\end{align*}
Note that the output of the Best-x Algorithm maximizes $R(\sigma^x,x)G_x$.
Therefore, in order to find the approximation ratio of the Best-x Algorithm, it suffices to provide a lower bound for
\begin{equation}\label{eq:ratio}
    \frac{\max_k\, R(\sigma^k, k) G_k}{\sum_{x=1}^M g_x R(\sigma^x,x)},
\end{equation}
where the numerator is the lower bound of the expected revenue from Best-x and the denominator is an upper bound for the optimal expected revenue.

We prove the performance guarantee of Best-x by examining the worst-case structure of $\max_k \, \EE(R(\sigma^k,X))$ over all distributions of $X$ that satisfy Assumption~\ref{assump:IFR} and all functions of $R(\sigma^x,x)$ that satisfy Theorem~\ref{thm:decrease_margin}.
For convenience, we normalize the upper bound~(\ref{eq:UB}) to one and consider the following min-max problem:
\begin{align}
\label{prob:worstcase}
& \min_{R,G}\max_{k} \quad R(\sigma^k,k)G_k \nonumber \\
\mbox{s.t. } &1= G_1 \geq G_2 \geq \cdots \geq G_M \geq G_{M+1}=0 \nonumber   \\
                  & g_x=G_{x} - G_{x+1},\quad x =1,\cdots, M \nonumber  \\
                  &G_{x+1}G_{x-1} \leq G_x^2, \forall x =2,\cdots, M-1 \nonumber  \\
& R(\sigma^x,x)\le R(\sigma^{x+1},x+1), \forall x=1,\dots,M-1 r  \\
& R(\sigma^{x-1},x-1)+R(\sigma^{x+1},x+1)\nonumber    \\ 
& \quad \quad \quad \le 2R(\sigma^{x},x) , \forall x=2,\dots,M-1 \nonumber \\
& \sum_{x=1}^M g_x R(\sigma^x,x) = 1, \nonumber  \\
&R(\sigma^x,x) \geq 0, \quad\forall x=1,\dots,M. \nonumber 
\end{align}
The first three constraints capture the fact that $G_x$ is the tail probability of a random variable with IFR.
The fourth and fifth constraints follow from Theorem~\ref{thm:decrease_margin}.
The sixth constraint normalizes the upper bound to one.

Next, we prove that the optimal value of the min-max problem~\eqref{prob:worstcase} is $1/e$.
\begin{restatable}{theorem}{worstcase}
\label{thm:approx}
    The minmax problem \eqref{prob:worstcase} has an optimal value $1/e$,
    i.e.,
    $$
   \min_{R,G}\, \frac{\max_k \, R(\sigma^k, k) G_k}{\sum_{x=1}^M g_x R(\sigma^x,x)}= \frac{1}{e}
    $$
    and hence the approximation ratio of the Best-x Algorithm is $1/e$. 
    The $1/e$ bound is achieved when $R(\sigma^x,x)$ is linear in $x$ and the attention span $X$ has a geometric distribution with success probability approaching 0.
\end{restatable}
To prove the theorem, we find that if $R(\sigma^x,x)$ does not have a linear structure, or the attention span does not have a geometric distribution, we can always perturb their values such that $\max_k \, R(\sigma^k, k) G_k$ does not change but the clairvoyant revenue $\sum_{x=1}^M g_x R(\sigma^x,x)$ strictly increases. Once we establish the worst-case structure, the proof of the bound follows naturally.

Theorem \ref{thm:approx} is in sharp contrast to \cite{brubach2021follow}, where they show that any algorithm can be arbitrarily bad relative to the clairvoyant upper bound.
By adding the arguably mild IFR assumption, we can guarantee a $1/e$ performance ratio relative to the clairvoyant revenue.

As we can see, when considering random attention spans with IFR, the approximation algorithm we have developed proves to be quite efficient. In particular, when we direct our attention to problem formulation (\ref{prob:worstcase}), we find that the approximation ratio of $1/e$ is not only tight but also achievable. This scenario occurs when the revenue function $R(\sigma^x,x)$ increases linearly with $x$ and when the attention span has a geometric distribution with $M \rightarrow +\infty$ and the success probability tending to zero. However, it is important to note that the condition $R(\sigma^x,x)\le R(\sigma^{x+1},x+1), \, \forall x=1,\dots,M-1$ and $R(\sigma^{x-1},x-1) + R(\sigma^{x+1},x+1) \leq 2R(\sigma^{x},x)$ for all $x = 2, \ldots, M-1$ from formulation (\ref{prob:worstcase}) is merely a necessary condition for the clairvoyant revenue. Meanwhile, the term $R(\sigma^k,k)G_k$ represents a lower bound of the revenue achievable by the Best-x algorithm. This discrepancy suggests that there might be a loss in the tightness of real performance bound due to the relaxation of the problem formulation.

Fortunately, we can show that no algorithm can achieve more than $1/2$ of the clairvoyant revenue. Therefore, the optimality gap is capped at $1/2 - 1/e \approx 0.13$. We acknowledge that our current analytical methods fall short of completely closing this gap and leave it as a subject for future research.

\begin{proposition}\label{prop:tightness2}
        There exist $M$, $\{r_i,\lambda_i\}_{i=1}^M$, and a distribution of $X$ satisfying Assumption~\ref{assump:IFR} such that no algorithm can achieve more than $1/2$ of the clairvoyant upper bound:
        \begin{equation*}
            \max_{\sigma}\EE[R(\sigma,X)]\le \frac{1}{2}\sum_{x=1}^M g_x R(\sigma^x,x).
        \end{equation*}
\end{proposition}

In the proof, we construct a special instance in which we prove that no algorithms can achieve an expected revenue higher than $1/2$ of the clairvoyant upper bound.

\begin{remark}[Rank the remaining products.]\label{rmk:filling}
After finding the best x in Algorithm~\ref{alg:approx}, the algorithm doesn't fill in the remaining $M-x$ positions automatically.
    In fact, the theoretical results in Theorem~\ref{thm:approx} hold even with the remaining $M-x$ positions unfilled.
Empirically, however, filling the remaining positions with products always increases the expected revenue.
In practice, after ranking $x$ products for the Best-x algorithm, we can fill the positions up to $M$ greedily, i.e., iteratively inserting one of the remaining products into the current ranking that yields the highest marginal increase of the expected revenue.
In Section~\ref{sec:numeric-approx},  we show that empirically, Best-x algorithm with greedy filling always achieves more than 86\% of the clairvoyant upper bound, significantly outperforming a number of heuristic benchmarks.
\end{remark}

\begin{remark}[Comparison with \cite{kempe2008cascade}]\label{rmk:comparison_kempe} Compared to the model in \cite{kempe2008cascade},
we have made four technical contributions. (1) While \cite{kempe2008cascade} prove a performance ratio relative to the optimal expected revenue, we compared with a more powerful benchmark, i.e., the revenue of a clairvoyant who perfectly knows the attention span of each arrival; (2) Under Assumption~\ref{assump:IFR}, we improve the $(1/4 - \epsilon)$-approximation ratio of \cite{kempe2008cascade} to $1/e$. This improvement is noteworthy as it is achieved against the aforementioned more stringent benchmark.
Moreover, our algorithm is easy to interpret and implement while \cite{kempe2008cascade} consider a dynamic program with continuous state space and the error margin $\epsilon$ appears in the discretization. 
We also show that no algorithm can achieve better than $1/2$ of the clairvoyant revenue; (3) The computational cost for our Best-x algorithm is $O(Mn)$ while  \cite{kempe2008cascade} provide a fully-polynomial-time approximation scheme (FPTAS) with larger computational complexity; and (4) we provide a nested structure for the optimal rankings $\{ \sigma^x\}_{x=1}^M$ under the fixed attention span, which provides insights to the design of approximation algorithms and heuristics for future work. 
\end{remark}

\subsection{Special Cases with Solvable Optimal Ranking}\label{sec:special-cases}
Although we have established the performance bound of the Best-x algorithm, it is developed for the worst-case scenario.
For some special cases, the \emph{optimal} ranking under the random attention spans can be found in polynomial time.
We provide two such cases in this section.

\textbf{Case one: prefixing rankings.} Proposition~\ref{prop:nested} shows the nested structure in the optimal ranking. That is, the optimal ranking $\sigma^{x+1}$ is attained either by inserting a new product at the top or the midst of $\sigma^x$, or by appending a product to the end of of $\sigma^x$. 
It is important to note that in the former scenario, the top $x$ recommendations for customer $x+1$ do not constitute the optimal set for customer $x$. However, if it is the latter scenario, and this appending mechanism is consistent for all $x=1,\ldots,M-1$, then it is evident that the first $x$ recommendations within $\sigma^M $ form an optimal ranking for each customer $x=1,\ldots,M-1$. This particular optimal ranking configuration is denoted as a \emph{prefix structure}, which imposes a stricter criterion than the \emph{nested structure}.

\begin{restatable}{definition}{defprefix}
 \label{def:prefix}
    An optimal ranking for customers with attention span $x$ is a \emph{prefix} to that for customers with attention span $x+1$, if $\sigma^{x+1}$ can be attained by appending one product to the bottom of $\sigma^x$.   
\end{restatable}
This prefix condition inherently guarantees the optimality of rankings for varying attention spans. The retailer can employ $\sigma^{M}$, the optimal ranking for the maximum attention span $M$, computed via Algorithm~\ref{alg:astopt2}. For customers with attention spans shorter than $M$, they are effectively presented with the top $x$ products of $\sigma^M$, which are arranged identically to $\sigma^x$. Consequently, $\sigma^M$ serves as the optimal ranking irrespective of the distribution of attention spans. Essentially, the prefix structure is akin to assigning a consistent product ranking for all customers, with each customer viewing the same ordering up to the limit of their attention span.

We next provide the sufficient and necessary condition for prefixing rankings.
Suppose $i_k$ is the product that has the $k$th highest $\lambda \cdot r$ among all products for $k=1,\dots,M$.
That is, $\lambda_{i_1} r_{i_1} \ge \lambda_{i_2} r_{i_2} \ge \dots \ge \lambda_{i_M} r_{i_M}\ge \lambda_{j} r_{j}$ for $j\notin \{i_1, ..., i_M\}$.

\begin{proposition}\label{prop:prefix}
The optimal ranking $\sigma^k$ is a prefix of $\sigma^{k+1}$ for all $1\leq k \leq M-1$ if and only if
$i_1 < i_2 < \cdots < i_M$.
\end{proposition}

Recall that we index products in the order of decreasing prices according to \eqref{eq:product-indexing}.
Therefore, the condition in Proposition~\ref{prop:prefix} states that the order of the product prices is consistent with the expected revenues that take into account their attractiveness.
In other words, the products need to display clear ordering for the retailer: more expensive products also generate higher expected revenues.
If there are products sold at high prices, but their purchase probabilities are low and drag down the expected revenues, then prefixing rankings are not optimal.

\textbf{Case two: geometric distribution.}
When the attention span is random, we do not have an iterative formula like \eqref{eq:rec3} for the expected total revenue.
As a result, we cannot gradually add products to existing rankings to construct longer rankings and resort to the dynamic programming.
This motivates us to investigate special attention span distributions that can preserve the iterative structure in \eqref{eq:rec3}.
We show that when the attention spans have a (possibly truncated) geometric distribution,
i.e.,  the tail probability $G_k = \PP(X \geq k)$ can be written as $G_k = \alpha^{k-1}$ for $1\leq k \leq M$,
we can develop dynamic programming to find the optimal ranking.

Under the truncated geometric distribution, the optimal expected revenue \eqref{eq:opt_random} can be expressed as
\begin{align*}
& \EE[R(\sigma, X)] 
\nonumber \\
& =  \sum_{x=1}^{M} \prod_{i=1}^{x-1} (1-\lambda_{\sigma(i)} ) \cdot \lambda_{\sigma(x)} r_{\sigma(x)}  \alpha^{x-1}
\\
& = \alpha (1- \lambda_{\sigma(1)})  \Big (  \sum_{x=2}^{M} \prod_{i=1}^{x-1} (1-\lambda_{\sigma(i)} )  \lambda_{\sigma(x)} r_{\sigma(x)}  \alpha^{x-2}\Big )  \nonumber 
\\
& \quad + \lambda_{\sigma(1)} r_{\sigma(1)} 
\\
& =  \lambda_{\sigma(1)} r_{\sigma(1)} + \alpha (1- \lambda_{\sigma(1)})  \Big (\EE[ R(\sigma[2,M], X ) ] \Big ),
\end{align*}
where $\sigma[2,M]$ denotes the sub-ranking of $\sigma$ from position two to $M$. Note that $\EE [R(\sigma[2,M], X )]$ is expected revenue conditional on the event that the customer doesn't purchase the first product.
By the structure of the geometric distribution,  it is the same as the expected revenue from displaying $\sigma[2,M]$ from position one to position $M-1$ for customers whose attention spans are geometrically distributed and truncated at $M-1$.
The above recursive formula gives us the basis for dynamic programming---fixing the top product, the rest products should be displayed to maximize $\EE [R(\sigma[2,M], X )]$.

However, we still need a similar result to Lemma~\ref{lem:fix-m-decrease-r} that dictates the ordering of the products in the optimal ranking.
Otherwise, the state space of dynamic programming, which is the set of products in the sub-rankings,
is combinatorial and growing exponentially.
This is given in the next proposition:
\begin{proposition}\label{prop:geo_discount}
Suppose $G_k = \alpha^{k-1}$ for $1 \leq k \leq M$ with some $\alpha< 1$.
The optimal ranking $\sigma^*$ satisfies
$$    \frac{r_{\sigma^*(k)} \lambda_{\sigma^*(k)}}{ 1- \alpha ( 1- \lambda_{\sigma^*(k)} ) }  >     \frac{r_{\sigma^*(k+1)} \lambda_{\sigma^*(k+1)}}{ 1-  \alpha ( 1- \lambda_{\sigma^*(k+1)} )  }, \text{ for } 1\leq k \leq M-1. $$
\end{proposition}

By Proposition~\ref{prop:geo_discount}, we can relabel the products according to the descending order of $r\lambda/(1-\alpha(1-\lambda))$,
e.g., product one has the largest value $r_1\lambda_1/(1-\alpha(1-\lambda_1))$.
In the optimal ranking, the products must obey this order, which greatly limits the search space for dynamic programming and enables us to apply the same philosophy of Algorithm~\ref{AST:m} to design Algorithm~\ref{alg:dp-geometric}.

\begin{algorithm}[t]\caption{ Optimal Ranking Under Geometrically Distributed Random Attention Span}\label{alg:dp-geometric}
\begin{algorithmic}
\REQUIRE{$r_j$, $\lambda_j$ for $j\in [n]$ where products are indexed in decreasing order of $\frac{r\lambda}{1-\alpha(1-\lambda)}$; $\alpha$, $M$}
\STATE $H^0_{j} = 0$ for all $j = 1,2,\dots,n$; $Z^k_{n+1} =0$ for all $k=0,1,2,\dots,x$; $\tilde \sigma^k_{n+1} =\emptyset$ for all $k=0,1,2,\dots,M$
\FOR{$k = 1, \ldots, M$ }
   \FOR{$j = n, \ldots, 1$ }
   \STATE \begin{equation} 
   Z^k_j\gets \max \Big \{ r_{j} \lambda_j +  \alpha (1-\lambda_j) Z_{j+1}^{k-1}  , \, Z^k_{j+1} \Big \}
   \end{equation}
    \IF {$ \Big( r_{j} \lambda_j + \alpha (1-\lambda_j) Z_{j+1}^{k-1} \Big) \geq Z^k_{j+1}$}
   \STATE $\quad \tilde \sigma^k_j\gets \tilde \sigma^{k-1}_{j+1} \cup \{j\}$
    \ELSE
    \STATE  $\quad \tilde \sigma^k_j\gets \tilde \sigma^k_{j+1}$
    \ENDIF
   \ENDFOR
\ENDFOR
\RETURN $Z^{M}_1$ and $\tilde \sigma^M_1$
\end{algorithmic}
\end{algorithm}

\section{Personalized Ranking via Online Learning} \label{sec:learning}
In Section \ref{sec:app-alg}, we develop the Best-x Algorithm to compute a ranking that preserves a guaranteed $1/e$ revenue of the optimal.
Despite its computational efficiency, the algorithm encounters several practical challenges:
\begin{itemize}
    \item The algorithm requires the prior knowledge of the conditional purchase probabilities $\lambda_j$'s and the distribution $G$ of the attention span, which are typically unknown to the retailer in most real world scenarios, especially for new entrants to the industry.
    \item Customers may have vastly distinct tastes and preferences toward the products.
        That is to say, the conditional purchase probability of the same product may vary for different customers. Fortunately, customers usually arrive with a characterizing feature that can reveal her preference, and it is desirable to design personalized product rankings accordingly. However, the relationship between the feature and the preference is usually unknown.
    \item In online retailing, the firm is usually able to display a huge catalog of products even for the same search keyword or category.
        Some products can be very similar and only differ in a few dimensions such as color and size.
        In this case, one would expect the conditional purchase probabilities of the products may also be correlated with the product features.
\end{itemize}
In this section, we develop a framework based on online learning to address these practical considerations.
We assume the customers have the same attention span distribution, but distinct conditional purchase probabilities based on their own features.
The products may also be differentiated by product-related features.
We develop a learning algorithm that actively learns the span distribution and the conditional purchase probabilities,
and simultaneously approximates the personalized $1/e$-optimal ranking for each customer.
\subsection{Preliminaries}\label{sec:learning-pre}
The online-retailer is expecting the arrival of $T$ future customers.
Customer $t$ has a random attention span with distribution $G^{\ast}$.
Note that we use the superscript $*$ to denote the actual value, to differentiate with the estimation.
Moreover, her probability of purchasing item $j$ depends on both the customer feature $\by_t \in \R^{d_c}$ and the product feature $\bs_j \in \R^{d_p}$. Moreover, we assume there exists an unknown matrix $\Theta^* \in \R^{d_p \times d_c}$ encoding the interaction between the product and customer that yields a conditional purchase probability $\by_t^\top  \Theta^* \bs_j $. In particular, we have
\begin{equation}\label{eq:feature-choice-prob}
    \lambda_{t,j} = \bs_j^\top  \Theta^* \by_t \in(0,1).
\end{equation}
\textbf{Feature vectorization:} For customer $t$ with feature $\by_t$, let $\bx_{t,j} \triangleq vec(\bs_j \by_t^\top) \in \R^{d_p d_c}$ and $\btheta^* \triangleq vec(\Theta^* ) \in \R^{d_p d_c} $ be the vectorization of $\bs_j \by_t^\top$ and $\Theta^*$.
The term $\bs_j^\top  \Theta^* \by_t $ can be expressed as $\bs_j^\top  \Theta^* \by_t = \bx_{t,j}^\top \btheta^*$.
Throughout the rest of the paper, we use $\bx_{t,j}$ to represent the vectorized feature and $d = d_p d_c$ to represent their compounded dimension for ease of notation.
We assume that the $\ell_2$-norm of the features $\bx_{t,j}$ and the parameter $\theta^*$ are bounded:
\begin{assumption}\label{assump:feature_norm}
For all customer $t$ and product $j$, the vectorized feature is bounded by $\| \bx_{t,j} \|_2 \leq 1$. The parameter $\btheta^*$ satisfies $\| \btheta^*\|_2 \leq D$ for some constant $D>0$.
\end{assumption}
This assumption is mild and can always be satisfied by proper rescaling of the features.

\noindent
\textbf{Online learning:} We denote $r_{\max} \triangleq \max_{j \in [n]}r_j$ as the maximal revenue among all products.
To maximize the revenue, after observing the feature of a consumer $\by_t$, and thus $\bx_{t,j}$ for all $j\in [n]$, the firm would calculate the conditional purchase probabilities according to \eqref{eq:feature-choice-prob} and optimize the personalized ranking using the algorithm in Section~\ref{sec:app-alg}.
However, the firm usually does not know $\btheta^*$ or $G_x^{\ast}$ for $x=1,\dots,M$ initially.
Therefore, we set up an online learning framework to address the problem.

Suppose the firm is expecting $T$ customers arriving in sequence.
For customer $t$, the firm may display a ranking $\sigma_t$.
The feature $\by_t$ determines the purchase probability $\lambda_{t,j} = \bs_j^{\top} \Theta^* \by_t$, and the expected revenue conditional on $\by_t$ is denoted as $\EE[R(\sigma_t, X; \lambda_t,G^*)]$.
The goal of the firm is to maximize the total expected revenue gained from the $T$ customers.

The key component of online learning is that the decision of $\sigma_t$ depends on the information extracted from the interactions with customers prior to $t$, but not on the unknown $\btheta^*$ or $G^{\ast}$.
This is referred to as the information structure.
For a past customer $s<t$ facing ranking $\sigma_s$, there are two possible outcomes observed by the firm.
She may purchase a product in $\sigma_s$, say, the product in the third position.
She may also leave without purchasing anything.
In the latter case, we assume that the firm observes the product position after which the consumer leaves.
For example, she leaves after viewing four products.
This is a typical setting in online retailing as the product view in a mobile phone can be precisely tracked.

To encode the information structure, we use $(\Psi_t,\Upsilon_t)$ to represent the observed behavior of customer $t$.
More precisely, $\Psi_t\in \left\{0,1\right\}$ and $\Psi_t=1$ if and only if customer $t$ purchases a product.
In this case, $\Upsilon_t\in [M]$ is the position of the product that is purchased.
Otherwise, if $\Psi_t=0$, then $\Upsilon_t\in [M]$ is the number of products customer $t$ views before leaving.
With this set of notations, the ranking $\sigma_t$ determined by the firm may depend on $\mathcal F_{t-1}\triangleq \sigma(\Psi_{1},\Upsilon_1,\dots,\Psi_{t-1},\Upsilon_{t-1})$, i.e., all the past observations, and $\by_t$, or equivalently, $\lambda_{t,j}$ for $j\in[n]$.

\noindent
\textbf{Performance metric:}
We use the cumulative regret to evaluate the performance of the proposed algorithm,
which is one of the most common metrics in online learning.
For a given customer $t$ and its feature $\by_t$, an \emph{oracle} who knows the model parameters
would calculate the conditional purchase probabilities $\lambda_t$ according to \eqref{eq:feature-choice-prob}
and choose the optimal ranking for the customer. For the ease of presentation, we denote
$$
\mathcal{R}(\sigma;\lambda,G) = \EE[R(\sigma,X;\lambda,G)]
$$
throughout this section as the expected revenue from displaying $\sigma$ to a customer with conditional purchase probability $\lambda$ and attention span distribution $G$.
For customer $t$, the oracle would optimize the expected revenue
\begin{align}\label{eq:def_revenue_online}
  &  \max_{|\sigma|=M}\cR(\sigma;\lambda_t,G^{\ast}) \nonumber 
   \\
   & =   \max_{|\sigma|=M} \sum_{x=1}^{M} \prod_{i=1}^{x-1} (1-\lambda_{t,\sigma(i)} ) \cdot \lambda_{t,\sigma(x)} r_{\sigma(x)}  G_{x}^{\ast} .
\end{align}
On the other hand, the expected revenue for the firm using $\sigma_t$ is $\cR(\sigma_t;\lambda_t, G^{\ast})$.

Usually, the cumulative regret is defined to be the performance gap between the oracle and the firm that uses an online learning algorithm.
In our case, because of the computation cost in finding the optimal ranking for the oracle, we consider the Best-x Algorithm to be the target to learn.
More precisely, for a sequence of customers $\{\by_t\}_{t=1}^T$,
we can define
\begin{align*}
   & \text{Reg}(T,y) = \sum_{t=1}^T\text{Reg}_t 
    \\
 &    \triangleq\sum_{t=1}^T \left(  \max_{|\sigma|=M}\cR(\sigma;\lambda_t, G^{\ast})-   e \cdot \cR(\sigma_t;\lambda_t, G^{\ast})\right).
\end{align*}
The target of the firm is not to learn the optimal ranking under the true parameters, but the Best-x algorithm which is guaranteed to have $1/e$ approximation ratio by Theorem~\ref{thm:approx}.
In other words, if the firm can effectively implement Best-x after learning the values of $\btheta^*$ and $G_x^{\ast}$ as $t$ increases, then the regret per period is diminishing, i.e., $\limsup_{T\to \infty}\text{Reg} (T)/T\le 0$.
We point out that the proposed learning algorithm can be slightly modified to achieve the same rate of regret when the target is another offline algorithm or even the optimal ranking (if the computational cost is not a concern) other than the Best-x algorithm. In the latter case, we don't have the scaling factor $e$ in the regret calculation.

\noindent
\textbf{Challenges in the algorithmic design and analysis:}
Although online learning has been studied extensively and various standard frameworks have been proposed,
the personalized ranking problem has several unique challenges.
First, in contrast to the classic multi-armed bandit problem, the observations in each round are censored and highly non-regular.
In particular, the customer may leave after viewing the first product and provides no information for the attractiveness of products displayed below the first one.
Or the customer may purchase the second product, providing a right-truncated censored sample for her attention span.
It is unclear how to rank the products to explore effectively and learn the distribution of the attention span as well as $\btheta^*$ in the presence of censoring.
For example, for large $M$, there are hardly any consumers viewing products ranked below, say, the $30$th position.
The scarcity of samples makes it impossible to accurately estimate $G_x^{\ast}$ for $x\ge 30$.
We show that our algorithm is not sensitive to the parameters associated with a high degree of censoring.
Intuitively, this is because $G_x^{\ast}$ for large $x$ has little impact on the expected revenue of any ranking.

The second difficulty due to censoring lies in the estimation of $G_x^{\ast}$. {Recall that when a customer purchases a product in position $x$, we do not observe her attention span other than the fact that it must be greater than or equal to $x$.
On the other hand, if one only uses the customers who leave empty-handed to estimate $G^\ast$, whose attention span is indeed observed, then the data is not fully utilized and the estimation will be biased, because the empty-handed customers tend to have shorter attention spans.
The key step in our approach is to convert the estimand from the CDF $G_x^\ast$ to the failure rate $h_x^{\ast}\triangleq (G_x^{\ast}-G_{x+1}^{\ast})/G_x^{\ast}$.
For the failure rate, even the customers who make a purchase can be used and 
we construct unbiased estimators that serve as a building block for the learning algorithm.
}

The third difficulty lies in the simultaneous learning of attention spans and feature-based conditional purchase probabilities.
Unlike existing works in cascade bandits that assume customers leave the platform after viewing exactly $M$ products,
the random attention span in our model introduces another source of randomness, which drastically complicates the structure of the observed information and the analysis.
We develop efficient learning algorithm that incorporates both conditional purchase probabilities and attention spans into a unified framework, and achieve good performance in both theoretical and practical perspectives.
\begin{remark}[Vectorized features]
    In practice, the vectorized outer product $vec(\bs_j \by_t^\top)$ may be high-dimensional, which leads to unstable estimation for $\Theta^*$ or $\btheta^*$.
    A popular remedy is to use the concatenated features, $(\bs_j, \by_t)$, instead of the outer product.
    It tremendously reduces the dimension while ignoring the potential interaction effects of consumers and products.
    Based on the data availability, either formulation can be preferable.
    In this study, we focus on the outer product, which is more technically challenging because of the singularity of the outer product (not having full rank).
    Moreover, the algorithm can be easily adapted to the concatenated formulation.
\end{remark}
\begin{remark}[The benefit of using features]
    The benefit of collecting and leveraging the consumer feature to the retailer is clear: it allows the firm to design personalized product display, which better matches consumers to their preferred products and thus extracts more revenue.
    The benefit of using product features, as opposed to treating products independently, is less straightforward.
    In fact, if there are not many products, then it may be more efficient to learn $\lambda_{t,j}=\by_t^\top\btheta^*_j$ independently for each product, where the coefficient $\btheta^*_j$ encodes how product $j$ attracts consumers of certain features.
    The use of product features is most helpful when there are a large number of similar products which differ in a few dimensions.
    For example, in online retailing, the number of products $n$ usually exceeds 1,000, and they can usually be compactly represented by product features whose dimension is typically less than 20.
    Nevertheless, our algorithm works for both settings.
\end{remark}
\subsection{The RankUCB Algorithm}
\begin{figure}
\vskip 0.5cm
    \centering
    \includegraphics[width=1 \linewidth]{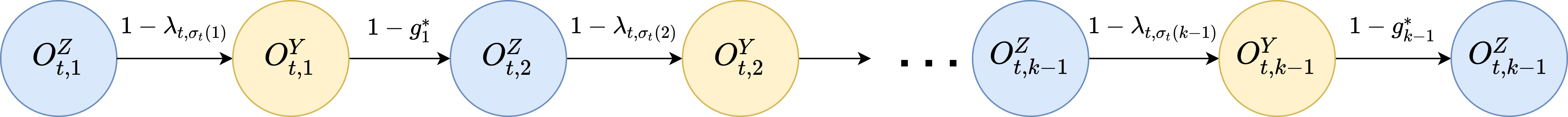}
    \caption{A graph illustrating the sequence of how $Z_{t,k}$ and $Y_{t,k}$ are observed.}
    \label{fig:observe}
\end{figure}

Next we introduce the algorithm, which is referred to as \emph{RankUCB}.
To introduce the algorithm, we first note that the observations $(\Psi_t,\Upsilon_t)$ from customer $t$ cannot be readily used.
In order to present the algorithm more compactly, we re-encode them as below.
We define a set of random variables $Y_{t,k}$, $Z_{t,k}$, $O^Y_{t,k}$ and $O^Z_{t,k}$ for $k=1,\dots,M$, contingent on $(\Psi_t,\Upsilon_t)$.
More precisely, if $\Psi_t=0$, then for all $k$
\begin{align*}
    Y_{t,k}=\ind_{k= \Upsilon_t}, \; Z_{t,k}= 0, \; O^Y_{t,k}= \ind_{k\leq \Upsilon_t},\; O^Z_{t,k} = \ind_{k\leq \Upsilon_t};
\end{align*}
if $\Psi = 1$, then
\begin{align*}
    Y_{t,k}=0, \; Z_{t,k}= \ind_{k = \Upsilon_t}, \; O^Y_{t,k}= \ind_{k\leq \Upsilon_t-1},\; O^Z_{t,k} = \ind_{k\leq \Upsilon_t}.
\end{align*}
To interpret, consider $Y_{t,k}$ to be a Bernoulli random variable with mean $h_k$, i.e., the failure rate at $X=k$, to indicate whether customer $t$ leaves after inspecting position $k$; and $Z_{t,k}$ to be a Bernoulli random variable with mean $ \bx_{t,\sigma_t(k)}^\top \btheta^*$, to indicate whether customer $t$ finds the product at position $k$ satisfying.
For example, if customer $t$ purchases the product ranked at the $i$-th position, then $Z_{t,k}$'s are zero for $k< i$ and $Z_{t,i}=1$.
More importantly, $O_{t,k}^Y$ and $O_{t,k}^Z$ indicate whether $Y_{t,k}$ and $Z_{t,k}$ are observed (not censored), respectively.
For example, if customer $t$ purchases the product ranked at position $i$, then we only observe that her attention span is no less than $i$, and thus $O_{t,k}^Y=1$ if and only if $k\le i-1$.
Note that the censoring has a nested structure: if $O_{t,k}^Y=1$, then $O_{t,i}^Y=1$ for $i\le k$.
The censoring mechanism is illustrated in Figure~\ref{fig:observe}.

\noindent
\textbf{Estimating $\btheta^*$:}
After observing the behavior of $t$ customers, we use
\begin{align}\label{eq:theta_t}
\hat \btheta_t =  \V_t^{-1} B_t
\end{align}
to estimate $\btheta^*$,
where $\V_t = \sum_{s=1}^t \sum_{k=1}^M O_{s,k}^Z \cdot \bx_{s,\sigma_s(k)} \bx_{s,\sigma_s(k)}^\top + \gamma \mathbf{I}_d$, $\mathbf I_d$ is the $d\times d$ identity matrix, and $B_t = \sum_{s=1}^t \sum_{k=1}^M O_{s,k}^Z  \cdot  \bx_{s,\sigma_s(k)}Z_{s,k}$. Here $\hat \btheta_t $ is the unique solution to $\min_{\btheta} \big \{\sum_{s=1}^t \sum_{k=1}^M O^Z_{s,k}(\bx_{s,\sigma_s(k)}^\top \btheta - Z_{s,k})^2 + \gamma \|\btheta \|^2 /2\big \}$, which is a typical $\ell_2$-regularized least square estimator.

\noindent
\textbf{Estimating the failure rate $h^{\ast}$: }
Recall that the failure rate $h_k^{\ast}$ is defined as $(G_{k}^{\ast}-G_{k+1}^{\ast})/G_k^{\ast}$, for $k=1,2,\dots,M-1$, and $h_M^{\ast}=1$.
We estimate the failure rate by
\begin{align*}
\hat h_{t,k}&  = \argmin_{h} \sum_{s=1}^tO^Y_{s,k} \cdot (h - Y_{s,k})^2  
\\
&   =\sum_{s=1}^t  O_{s,k}^Y \cdot \frac{Y_{s,k}}{N_{t,k}},
\end{align*}
where $N_{t,k} \triangleq \sum_{s=1}^t O_{s,k}^Y$ is the number of observed $Y_{s,k}$'s up to round $t$.
In other words, we use the frequency that a customer moves on to inspect the product ranked at $k+1$ from that at $k$.

The theoretical guarantees of the estimators are provided in the form of a confidence region below.
We denote $\| \bx \|_A = \sqrt{\bx^\top A \bx }$ for a positive definite matrix $A \in \RR^{d\times d}$.
\begin{lemma}\label{lemma:favorable_event}
For any $t\geq 1$, with probability at least $1-M/(t+1)^2$, the following event
occurs:
$
 \xi_{t} \triangleq \big \{  \| \btheta^* - \hat \btheta_{t}
 \|_{\V_t}  \leq \rho_{t},\, \,\, | h_k^* - \hat h_{t,k} | \leq  \sqrt{\ln(t+1)/N_{t,k}} , \,\,\forall k=1,2,\cdots, M-1 \big \},
$
where $\rho_t = \sqrt{d\log \big( 1 + tM/(\gamma d) \big )  + 4\log(t+1) }     + D\gamma^{1/2}$.
\end{lemma}
\noindent
\textbf{Optimistic estimator:}
The design of our algorithm follows the principle of ``optimism in the face of uncertainty,''
which is shared by all UCB-type algorithms.
In particular, when customer $t$ arrives, based on the confidence region provided in Lemma~\ref{lemma:favorable_event}, the firm calculates the \emph{optimistic estimators} in the confidence region:
\begin{align*}
h_{t,k}^L &\triangleq \min_{h_k \in  [0,1] }\Big \{ h_k: |h_k - \hat h_{t-1,k}| \leq \sqrt{\frac{ \ln (t)}{N_{t-1,k}}}  \Big\} 
\\
&   =  \text{Proj}_{[0,1]}\Big ( \hat h_{t-1,k} - \sqrt{ \frac{\ln (t)}{N_{t-1,k}}} \Big ) , \\
u_{t,j} &\triangleq \max_{\btheta: 0\leq \bx_{t,j}^\top \btheta \leq 1 } \Big \{\bx_{t,j}^\top \btheta: \|\btheta - \hat \btheta_{t-1}\|_{  \V_{t-1}}\leq \rho_{t-1} \Big\} 
\\
& = \text{Proj}_{[0,1]}\Big ( \bx_{t,j}^\top \hat \btheta_{t-1} + \rho_{t-1}\|\bx_{t,j}\|_{\V_{t-1}^{-1}} \Big ).
\end{align*}
In other words, we estimate the hazard rate $h_{t,k}^L$ to be the minimum in the confidence region; such choice guarantees that customers would view as many products as possible and thus generates the most optimistic revenue.
Moreover, $\prod_{k=1}^{x-1}h_{t,k}^L$ gives an optimistic estimator for $G_x^{\ast}$, denoted as $G_{t,x}^U$.
The quantity $u_{t,j}$ provides an optimistic estimator for the conditional purchase probability of product $j$, given the confidence region of $\btheta^*$ and the feature of customer $t$.

We present the details in Algorithm~\ref{alg:UCB}.
When customer $t$ arrives, the firm first calculates the optimistic estimators $u_{t,j}$ and $G_{k}^U$, as shown in Steps~\ref{step:u-est} and~\ref{step:h-est}.
Then, the optimistic estimators are plugged into Best-x to calculate the optimal ranking for customer $t$ (Step~\ref{step:approx-alg-ucb}).
After the observation of customer $t$ is collected, the firm updates the confidence region, which in turn is used for the next customer.


\begin{algorithm}[t]\caption{RankUCB}\label{alg:UCB}
    \begin{algorithmic}[1]
        \REQUIRE $1/e$-oracle {Best-x}, product feature $\bs_j$ and profit $r_j$ for all $j$, regularization parameter $\gamma \geq 1$, parameters $\rho_t$ for all $t$.
        \STATE{\bfseries Initialization:}{ $h_{0,k}^L = 0$ for each span $k=1,2,\cdots, M-1$, $\mathbf{V_0}= \gamma \mathbf{I} \in \R^{d\times d}$, $B_0 = 0 \in \RR^d$;}
        \FOR{$t = 1,2, \cdots, T$ }
        \STATE Observe $\by_t$
        \STATE{ Set $\bx_{t,j} = vec(\by_t \bs_j^\top)$ for each item $j $ and $\hat \btheta_{t-1} = \V_{t-1}^{-1} B_{t-1}$ }
        \STATE{\label{step:u-est} Update $u_{t,j} =  \text{Proj}_{[0,1]}\Big (\bx_{t,j}^\top \hat \btheta{t-1} + \rho_{t-1} \| \bx_{t,j}\|_{\V_{t-1}^{-1}} \Big ) $ for each item $j$}
        \STATE{ \label{step:h-est}Update $h_{t,k}^L = \text{Proj}_{[0,1]} \big ( \hat h_{t-1,k} - \sqrt{ \ln (t)/N_{t-1,k}} \big )$, $G_{t,x}^U = \prod_{s=1}^{x-1} (1-h_{t,s}^L)$ for each span $x=1,2,\cdots, M-1$ }
        \STATE{\label{step:approx-alg-ucb} Choose $\sigma_t = \text{Best-x}(r,u_{t},G_{t}^U)$ as the display for round $t$. Observe $\left\{(Y_{t,k}, Z_{t,k},O_{t,k}^Y,O_{t,k}^Z)\right\}_{k=1}^M$ }
        \STATE{ Update $\V_t = \V_{t-1} + \sum_{ i=1}^M O_{t,i}^Z \cdot  \bx_{t,\sigma_t(i)}\bx_{t,\sigma_t(i)}^\top  $, $B_t =B_{t-1}+ \sum_{i=1}^M  \bx_{t,\sigma_t(i)}\cdot O_{t,i}^Z \cdot  Z_{t,i}$}
        \STATE{Update $N_{t,k} = \sum_{s=1}^t O_{s,k}^Y$, set $\hat h_{t,k} = \sum_{s=1}^t O_{s,k}^Y \cdot Y_{s,k}/N_{t,k}$ for $k=1,2,\cdots, M$}
        \ENDFOR
\end{algorithmic}
\end{algorithm}

Next we analyze the regret of Algorithm~\ref{alg:UCB}, which is formally presented in the next theorem.
\begin{theorem}\label{thm:regret}
    Suppose Assumptions \ref{assump:IFR} and \ref{assump:feature_norm} hold for the sequence of customers $\{\by_t\}_{t=1}^T$. The cumulative regret of Algorithm \ref{alg:UCB} with $\gamma \geq 1$ can be bounded by
    \begin{align*}
     &    \text{Reg}(T,y) \\
       &  \leq  2 e \rho_{T-1} M r_{\max} \cdot   \sqrt{2 T  d \log \Big (1+ \frac{T M}{ \gamma d} \Big)}  \\
       & \quad + 4eM  r_{\max} \sqrt{T\ln T}  +  \frac{M\pi^2r_{\max}}{6},
    \end{align*}
    where $\rho_{T-1} = \frac{1}{2}\sqrt{d\log \big( 1 + (T-1)M/(\gamma d) \big )  + 4\log(T) }     + D\gamma^{1/2}$ and $r_{\max} = \max_{j \in [n]}r_j$.
\end{theorem}
We first remark on the dependence on various model parameters.
The regret grows at $\tilde{\cO}(\sqrt{T})$, which is the typical optimal rate in online learning problems.
The linear dependence on $r_{\max}$ is also necessary, which is the maximum of the single-period reward.
The regret is linear in $M$, the number of positions to display.
This is similar to the cascade bandit literature \citep{zong2016cascading}.
In terms of the dimension of the contextual information $d$, the dependence is $\tilde \cO(d)$, which is the same as the contextual bandit literature \citep{zong2016cascading}.
The linear dependence on $D$, the bound on $\|\btheta^*\|_2$, is also understandable, as the reward in each period scales linearly in $\|\btheta^*\|_2$.
Therefore, the regret matches the best-achievable rate in the literature.

We briefly remark on how we address the challenges raised in Section~\ref{sec:learning-pre}.
To utilize censored observations, we construct novel estimators $Y$ and $Z$ that can fully capture the information contained in the observation and are yet easy to manipulate to obtain unbiased estimators for the target quantities in spite of the censoring.
For the difficulty in estimating $G$, we transform it to the failure rate which can be efficiently estimated, whose estimation error is also easy to control.
To simultaneously learn the conditional purchase probabilities and the attention span distribution, we explicitly interpret the observing mechanism of choice and continuation action of customers (see Figure \ref{fig:observe}), and express the induced regret in terms of estimation errors of feature-based conditional purchase probabilities and failure rates.
These techniques allow us to show the regret bound which is of the same order as other similar problems with simpler settings.
\section{Numerical Experiments}\label{sec:experiment}
In this section, we conduct numerical experiments to examine the performance of algorithms in practice.
In particular, we first demonstrate the performance of Best-x
when the conditional purchase probabilities and the distribution of the attention span are known.
Then we show the performance of Algorithm~\ref{alg:UCB} when the parameters are unknown and need to be learned.
\subsection{The Best-x Algorithm}\label{sec:numeric-approx}
In our experiment, we consider 1000 products, $M=20$ products to display, and two possible distributions of the attention span (uniform and geometric), both satisfying Assumption~\ref{assump:IFR}. 
We also test a third setting where the attention span does not follow the IFR.
\begin{itemize}
\item Setting one: $G^{(1)} = (1,0.95,0.9,\cdots, 0.1,0.05)$.
\item Setting two: $G^{(2)} = (1,0.9,0.9^2,0.9^3,\cdots,0.9^{19})$.
\item Setting three: $G_k^{(3)} = \prod_{i=1}^{k-1}(1-h_i)$ where $h_i = 0.1 - 0.05i/M$.
\end{itemize}
For each setting, we randomly generate the conditional purchase probabilities and the prices $\{(\lambda_{j},r_j)\}_{j=1}^{1000}$ independently for 1000 instances.
More specifically, for each instance we generate independent prices from $U[0,10]$ and conditional purchase probability from $U[0,0.5]$, 1000 samples each, where $U[a,b]$ stands for a uniform random sample between $a$ and $b$.
We then sort prices and conditional purchase probabilities in the opposite order and assign the values to products such that $r_j \geq r_{j+1}$ and $\lambda_j \leq \lambda_{j+1}$ for $j=1,\cdots, 999$.
This is to create a realistic (popular products are more expensive) and ``hard'' scenario for algorithmic solutions,
which highlights the price and conditional purchase probability trade-off.

\begin{figure}[t]
\centering
\begin{subfigure}{0.4\textwidth}
\centering
\includegraphics[width=\textwidth]{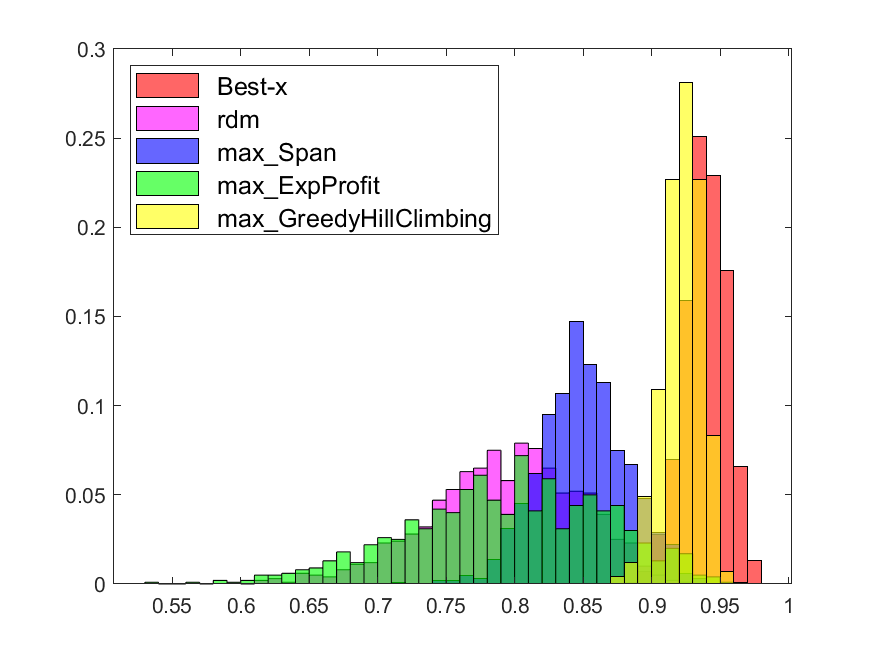}
\caption{$G^{(1)}$}
\end{subfigure}  
\begin{subfigure}{0.4\textwidth}
\centering
\includegraphics[width=\textwidth]{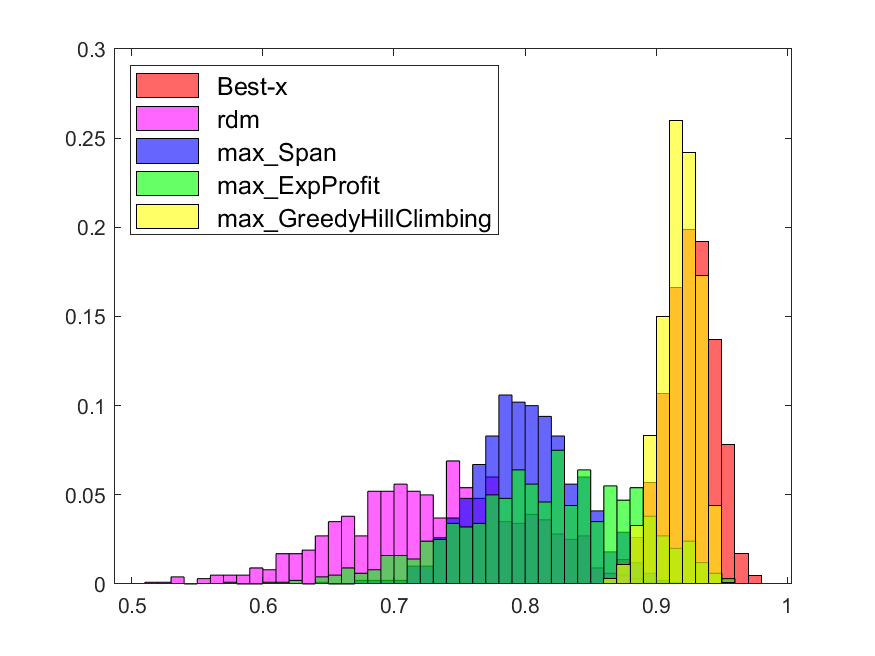}
\caption{$G^{(2)}$}           
\end{subfigure}
\begin{subfigure}{0.33\textwidth}
\centering
\includegraphics[width=\textwidth]{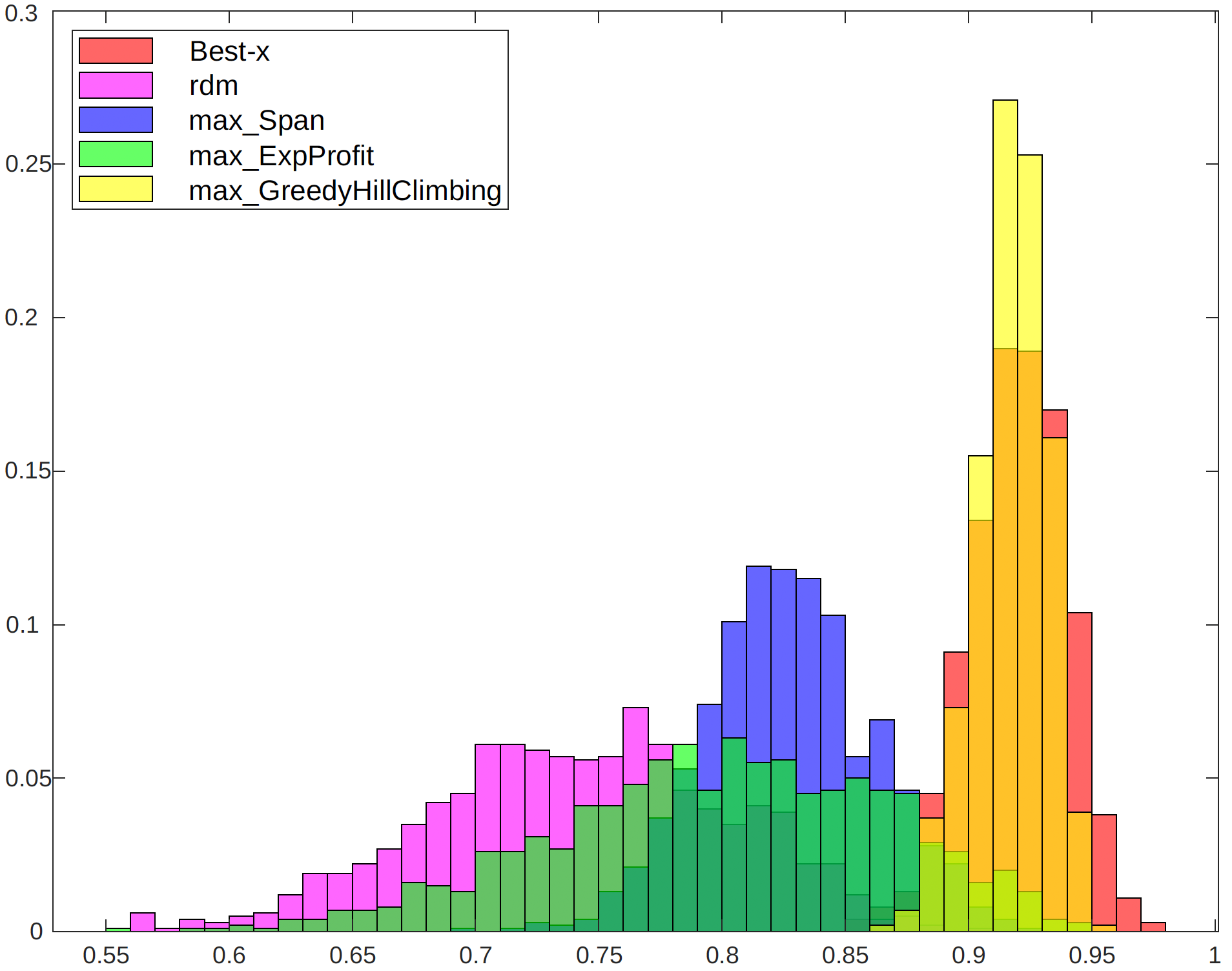}
\caption{$G^{(3)}$}
\end{subfigure}

\vskip 0.5cm
\caption{The histogram of 1000 instances for the performance of Best-x and other benchmarks relative to the clairvoyant upper bound under $G^{(1)} = (1,0.95,0.9,\cdots, 0.1,0.05)$, $G^{(2)} = (1,0.9,0.9^2,\cdots, 0.9^{19})$, and $G_k^{(3)} = \prod_{i=1}^{k-1}(1-h_i)$ where $h_i = 0.1 - 0.05i/M$.}
\label{fig:offline_01}
\end{figure}

The benchmark is the expected revenue relative to the upper bound provided in Proposition~\ref{prop:upper-bound}, as the optimal ranking is computationally difficult to solve.
We compare the performance of Best-x with four heuristic algorithms:
(1) \texttt{{rdm}}, randomly selecting 20 products, 
(2) \texttt{{max\_Span}}, using the optimal ranking when the attention span is fixed at $20$, 
(3) \texttt{{max\_ExpProfit}}, displaying the 20 products with the highest expected profits (price times the purchase probability), 
and (4)~\texttt{max\_GreedyHillClimbing}, adding 20 products greedily that maximize the marginal increase of the expected revenue. More precisely, for any ranking $\hat \sigma_k$ containing $k$ products, 
\texttt{max\_GreedyHillClimbing} constructs $\hat \sigma_{k+1}$ by inserting a product from the remaining pool at any position of $\hat \sigma_k$ that yields the largest marginal increase of the expected revenue, without changing the relative order of products within $\hat \sigma_k$. 
The performance is measured as the ratio to the upper bound derived in Proposition~\ref{prop:upper-bound}.\footnote{We have also tested two simple heuristics: 
\texttt{{max\_Profit}},	displaying the 20 most profitable products (with the highest prices) and
\texttt{{max\_Choice}}, displaying the 20 products with the highest purchase probabilities.
Both heuristics perform badly compared to others and we thus do not include them in the results.} 
The histogram of 1000 instances is illustrated in Figure~\ref{fig:offline_01} and their summary statistics in Table~\ref{tab:offline_01}.
We can see that ranking 20 products randomly performs worst among all strategies. 
Meanwhile, ranking the products based on their expected profits (\texttt{max\_ExpProfit}),
using the optimal ranking for customers with attention span 20 (\texttt{max\_Span}),
or the greedy heuristic
are reasonable strategies: they usually generate more than 80\% of the clairvoyant revenue.
The Best-x algorithm consistently outperforms all other heuristics.
It exceeds 86\% of the clairvoyant revenue in all the instances in both settings.

\begin{table}
\centering 
\begin{tabular}{ c c c c c c  c} 
 \toprule
  & Mean & Worst & 25\%  & 50\% & 75\% & Best \\ [0.5ex] 
 \midrule 
 \multicolumn{7}{c}{Uniform Distribution $G^{(1)}$}\\
 \midrule
\texttt{Best-x} & 0.9391  & 0.8878&  0.9292 & 0.9396& 0.9502 &0.9777 \\ 
 \texttt{rdm} & 0.7927 & 0.5974 &  0.7588 &  0.7963 & 0.8309 & 0.9279  \\
 \texttt{max\_Span} & 0.8497 & 0.7193  & 0.8288 &  0.8492 & 0.8699 &  0.9394  \\
 \texttt{max\_ExpProfit} & 0.7938 & 0.5395 & 0.7459 &  0.7995 & 0.8491 &  0.9531 \\
 \texttt{max\_GreeyHillClimbing} & 0.9225 & 0.8703 & 0.9143 & 0.9235 & 0.9322 &  0.9602  \\  
 \midrule 
 \multicolumn{7}{c}{Geometric Distribution $G^{(2)}$}\\
 \midrule
\texttt{Best-x} & 0.9255  & 0.8637 & 0.9125 &  0.9271 &  0.9392 & 0.9762  \\ 
 \texttt{rdm} & 0.7317 & 0.5182 &  0.6850 &  0.7352 &   0.7796 & 0.9027   \\
 \texttt{max\_Span} & 0.8008   & 0.6467 &  0.7754 &  0.8002 & 0.8256 &  0.9138  \\
 \texttt{max\_ExpProfit} & 0.8157 & 0.5758 &   0.7724 &  0.8205 &  0.8670 &  0.9572  \\
 \texttt{max\_GreeyHillClimbing} & 0.9175 & 0.8667 &  0.9090 &  0.9189 &  0.9282 &   0.9569  \\ 
 \midrule 
 \multicolumn{7}{c}{Decreasing Failure Rate $G^{(3)}$}\\
 \midrule
\texttt{Best-x} & 0.9167 & 0.8518  &   0.9034   &  0.9178& 0.9308  &  0.9681  \\ 
 \texttt{rdm} &  0.7357   & 0.5526  &   0.6939 &  0.7377 & 0.7782  & 0.8979  \\
 \texttt{max\_Span} & 0.8516   & 0.7158 &   0.82732032   &  0.8523 &   0.8786 &  0.9445\\
 \texttt{max\_ExpProfit} & 0.7988 & 0.5587 &  0.7543 &  0.8039 &  0.8499 &  0.9479   \\
 \texttt{max\_GreeyHillClimbing} & 0.9131   & 0.8597   & 0.9042  &   0.9144  & 0.9235  &  0.9533  \\ 
 \bottomrule
\end{tabular}
\caption{Summary statistics for the experiments in Figure~\ref{fig:offline_01}}
\label{tab:offline_01}
\end{table}

\subsection{RankUCB}\label{sec:numerical_rankUCB}
Next, we investigate the performance of Algorithm~\ref{alg:UCB} RankUCB.
In this experiment, we adopt a similar setting to that in Section~\ref{sec:numeric-approx}.
In particular, we consider 1000 products and $M=20$.
We implement RankUCB for 10 independent simulations, each with 10000 customers arriving sequentially.
For each customer, her attention span has a geometric distribution $G= (1,0.95,0.95^2,\cdots, 0.95^{19})$ from setting one in the last section, unknown to the firm.
We perform 10 simulation instances.
At the beginning of each simulation instance, we randomly generate the prices of the products uniformly from $[0,10]$
and their features $\bs_j \sim \cN(0.25,\mathbf{I}_{d_p\times d_p}) \in \RR^{d_p}$  with $d_p = 10$.
The features of the product are then normalized to $\| \bs_j\|_2 = 1$.
In each round, we randomly generate the customer feature $\by_t \sim \cN(1,0.1 \times \mathbf{I}_{d_p\times d_p}) \in \RR^{d_c}$ with $d_c= 5$ and also normalize it to $\| \by_t\|_2 = 1$.
We generate $ \btheta^* \sim \cN(0.25, \mathbb{I}_{50\times 50})$ and then normalize so that $\|\btheta^*\|_2 = 0.906$.
The same $\btheta^*$ remains the same for the 10 independent simulations. We also normalize the compound feature $\by_t \otimes  \bs_j$ when $\| \by_t \otimes  \bs_j \|_F$ is greater than 1, so that $\|\by_t \otimes  \bs_j \|_F \leq 1$.
In this setup, for any product $j$, it is guaranteed that the conditional purchase probability $|\lambda_{t,j}|  \le 0.906$.

\begin{figure}
\vskip 0.5cm
    \centering
    \includegraphics[width=0.5 \linewidth]{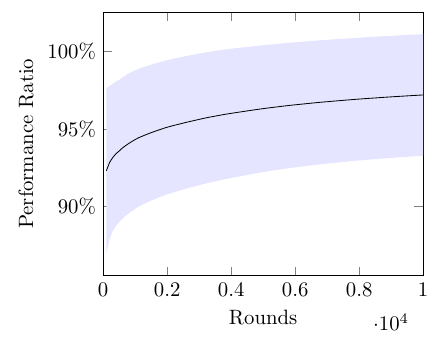}
    \caption{Performance ratio of RankUCB over 10000 rounds.}
    \label{fig:online_ratio}
\end{figure}

Because of the dimension of the problem, the instances tend to have high variances due to the random draw of the feature vectors.
To show the performance, we use a different measure than the regret.
In particular, in each round we calculate the \emph{expected} revenue of the ranking suggested by RankUCB conditional on the feature $\bx_{t,j}$ and thus the conditional purchase probabilities.
Then we calculate the ratio of the conditionally expected revenue of the Best-x Algorithm (Algorithm~\ref{alg:approx}) when all the information is known.
The expected revenue instead of the realized revenue helps smooth the performance.
Figure~\ref{fig:online_ratio} illustrates the ratio over 10000 rounds as well as the standard error of 10 simulations.
As we can observe, the expected reward of RankUCB increases to around 90\% in 10000 rounds.
This demonstrates its practical efficiency even under high dimensions $d=50$.

\begin{figure}
\vskip 0.5cm
    \centering
    \includegraphics[width=0.48 \linewidth]{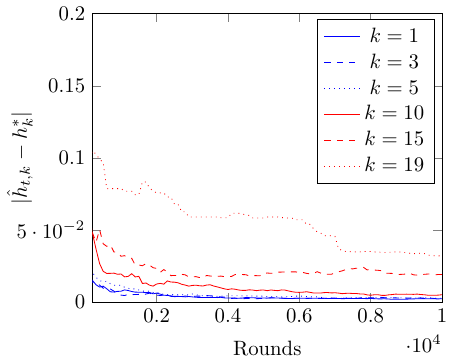}
        \includegraphics[width=0.48 \linewidth]{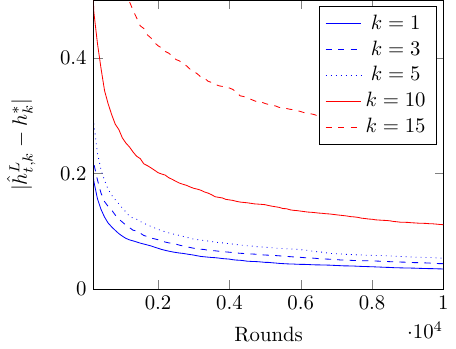}
        \caption{Estimated failure rate in the learning process.}
    \label{fig:online_failure}
\end{figure}

We also investigate learning of the failure rate in the experiment.
In particular, we show the point estimator $\hat h_{t,k}$ and the optimistic estimator $\hat h_{t,k}^L$ for the failure rate $h_k^*$ at $k\in \left\{1,3,5,10,15,19\right\}$.
We plot the absolute errors relative to the actual value $h_k^{\ast}\equiv 0.05$.
The results are illustrated in Figure~\ref{fig:online_failure}.

In Figure \ref{fig:online_failure}, we observe that the estimators for the smaller $k$ converge faster, because there is less censoring for products ranked
at the front.
The optimistic estimators converge much slower than the point estimators, because of the slow decay of the confidence region.
In particular, $\hat h_{t,1}^L$, $\hat h_{t,3}^L$ and $\hat h_{t,3}^L$ are very accurate after 5000 rounds, while $\hat h_{t,10}^L$, $\hat h_{t,15}^L$ and $\hat h_{t,19}^L$ are still too optimistic after 10000 rounds.
Nevertheless, as $T$ increases, the convergence of the estimates is reflected in the experiment.

\section{Conclusion}\label{sec:conclusion}
In this paper, we study the revenue maximization problem of an online retailer, when customers browse the ranked products in order.
The model extends the well-known cascade model in two directions: customers may have random attention spans, and the products have different prices.
Based on the structure of optimal ranking under fixed attention span, we develop an approximation algorithm with $1/e$-approximation ratio.
When the conditional purchase probabilities of the products (which may be based on customer and product features) and the distribution of customers' attention spans are unknown ,
we provide a learning algorithm that can effectively learn the parameters and achieve near-optimal $\tilde \cO(\sqrt{T})$ regret.
Our study addresses the two major challenges of the product ranking model when the firm is interested in revenue instead of click maximization.

\theendnotes

\section*{Code and Data} The code for numerical experiments are uploaded at GitHub (\textbf{https://github.com/chenny888/product-ranking-and-learning}).

\begin{appendices}

\section*{Appendix}

\section{Proofs in Section \ref{sec:app-alg}}\label{sec:proofs-main}
\textit{Proof of Lemma~\ref{lem:fix-m-decrease-r}:}
For the simplicity of notations, let us assume that $S=\left\{1,\dots,x\right\}$.
Suppose in the ranking $\sigma$ there exists $i\in\left\{1,\dots,x-1\right\}$ such that $r_{\sigma(i)}< r_{\sigma(i+1)}$.
We will argue that we can strictly improve the expected revenue by swapping the two products without changing the positions of other products.
Consider the new ranking $\sigma'$ with
$\sigma'(i)=\sigma(i+1)$, $\sigma'(i+1)=\sigma(i)$, and $\sigma'(k)=\sigma(k)$ for all other $k$.
Let $\pi_{\sigma(i)}(\sigma) \triangleq  \prod_{s=1}^{i-1} \big(1-\lambda_{\sigma(s)}\big)$ be the probability that the customer views product $\sigma(i)$.
By definition, it is easy to see $\pi_{\sigma(k)}(\sigma)  = \pi_{\sigma'(k)}(\sigma')$ for $k=1,\dots, i, i+2,\dots,x$.
Therefore, the expected revenues generated from products in position $k$ are equal under $\sigma$ and $\sigma'$ for $k=1,\dots,i-1,i+2,\dots,x$, recalling formula \eqref{eq:fix-span-revenue}.

In order to show $R(\sigma',x)>R(\sigma,x)$, it suffices to compare the revenues generated from the products in position $i$ and $i+1$ under the two rankings:
\begin{equation*}
    \pi_{\sigma(i)}(\sigma)  \lambda_{\sigma(i)} r_{\sigma(i)} +\pi_{\sigma(i+1)}   (\sigma)  \lambda_{\sigma(i+1)} r_{\sigma(i+1)}<\pi_{\sigma'(i)}(\sigma')  \lambda_{\sigma'(i)} r_{\sigma'(i)} +\pi_{\sigma'(i+1)}   (\sigma')   \lambda_{\sigma'(i+1)} r_{\sigma'(i+1)}.
\end{equation*}
This is indeed the case because
 \begin{align*}
& \pi_{\sigma(i)}(\sigma) \cdot r_{\sigma(i)} \lambda_{\sigma(i)} +\pi_{\sigma(i+1)}   (\sigma)  \cdot r_{\sigma(i+1)}\lambda_{\sigma(i+1)}\\
=&\pi_{\sigma(i)}(\sigma)  \Big (  r_{\sigma(i)} \cdot \lambda_{\sigma(i)}+(1-\lambda_{\sigma(i)}\big) \cdot r_{\sigma(i+1)} \cdot \lambda_{\sigma(i+1)} \Big ) \\
=&\pi_{\sigma(i)}(\sigma) \Big ( r_{\sigma(i+1)}  \cdot \lambda_{\sigma(i+1)}+r_{\sigma(i)}  (1-\lambda_{\sigma(i+1)}) \cdot \lambda_{\sigma(i)} + (r_{\sigma(i)}-r_{\sigma(i+1)}) \lambda_{\sigma(i)} \lambda_{\sigma(i+1)} \Big ) \\
<& \pi_{\sigma(i)}(\sigma) \Big ( r_{\sigma(i+1)} \cdot \lambda_{\sigma(i+1)}  +  r_{\sigma(i)} (1-\lambda_{\sigma(i+1)}\big) \cdot \lambda_{\sigma(i)}  \Big ) \\
=&\pi_{\sigma'(i)}(\sigma') \cdot r_{\sigma'(i)} \lambda_{\sigma'(i)} +\pi_{\sigma'(i+1)}   (\sigma')  \cdot r_{\sigma'(i+1)}\lambda_{\sigma'(i+1)}.
\end{align*}
The inequality is because $r_{\sigma(i)}-r_{\sigma(i+1)}<0$ and $1>\lambda>0$.
Therefore, $\sigma$ cannot be optimal and we have completed the proof.

\QED

\textit{Proof of Proposition~\ref{prop:dp}:}
First of all, we show $\sigma_j^k$ is the optimal display of selecting $k$ products from $\{j,j+1,\cdots,n\}$ based on the recursive relationship in dynamical programming, with corresponding revenue $H_j^k$.

For $x = 1$, we can see $H_j = \max_{s \geq j} r_s \lambda_s $, which exactly represents the maximum reward obtained by displaying one product from $\{j,j+1,\cdots, n\}$.
Meanwhile, for any arbitrary $x \leq n$, we can see $\sigma_{n-x+1}^x = \{n-x+1,n-x+2,\cdots, n\}$, which displays $x$ products the last $x$ ones. Thus, $H_{n-x+1}^x$ is also optimal for $x\leq n$. We use these two scenarios as the base case, and prove the claim by induction. In particular, we show that if both $\sigma_{j+1}^x$ and $\sigma_{j}^{x-1}$ are optimal displays, then $\sigma_{j}^x$ is also optimal.

Recall the recursive relationship \eqref{eq:dp_recursion} that
$$
H_{j}^x = \max \Big(r_j \lambda_j + (1-\lambda_j) H_{j+1}^{x-1}, H_{j+1}^x \Big).
$$
Suppose $\sigma_j^x$ is not optimal, then there must exist another display $\tilde \sigma_j^x$ with corresponding expected reward $\tilde H_j^x > H_j^x$. We consider two scenarios: (i) Suppose $j \in \tilde \sigma_j^x$, let $\tilde \sigma_{j}^x [2,x]$ be the display from the second product of $\sigma_j^x$ containing $(x-1)$ products with reward $R(\tilde \sigma_{j}^x [2,x], x-1)$, then we have
$$
\tilde H_{j}^x = r_j \lambda_j + (1-\lambda_j) R(\tilde \sigma_{j}^x [2,x], x-1) > H_{j}^x \geq  r_j \lambda_j + (1-\lambda_j) H_{j+1}^{x-1},
$$
which implies $R(\tilde \sigma_{j}^x [2,x], x-1)  > H_{j+1}^{x-1}$ and $\sigma_{j}^x [2,x]$ yields higher reward than $\sigma_{j+1}^{x-1}$. This contradicts the optimality of $\sigma_{j+1}^{x-1}$.
(ii) Suppose $j \notin \tilde \sigma_j^x$, we have $\tilde \sigma_j^x = \sigma_{j+1}^x$, which also contradicts the assumption that $\tilde H_j^x > H_j^x \geq H_{j+1}^x$. Combining both cases, we conclude that $\sigma_j^x$ is also optimal and completes the induction proof.

Next, it suffices to show that $\sigma_j^k$ preserves the least lexicographic value to ensure $\sigma_j^k$ is $\cL$-optimal.
We use induction to prove that for any $1 \leq k \leq n$, $\sigma_j^k$ is the $\cL$-optimal display of selecting $k$ products from $\{j,j+1,\cdots,n\}$.

For $k =1$, let $\sigma_j^1 =\{s\}$ be the solution returned by Algorithm \ref{AST:m} for $1\leq j \leq s \leq n$, based on the recursive relationship \eqref{eq:dp_recursion}, we can see
$ \lambda_u r_u < \lambda_s r_s$ for all $j \leq u \leq s-1$, and $\lambda_v r_v \leq \lambda_s r_s$ for all $s+1 \leq v \leq n$. Hence, $\sigma_j^1$ is $\cL$-optimal for any $1\leq  j \leq n$.

Next, suppose that argument is true for $k$. That is,
$\sigma_j^k$ is $\cL$-optimal for all $1\leq j \leq n-k+1$, we show that $\sigma_j^{k+1}$ is also $\cL$-optimal for $1\leq j \leq n-k$.

Suppose $\sigma_j^{k+1}$ is \emph{not} $\cL$-optimal, then there exists a size-$(k+1)$ display $\tilde \sigma_j^{k+1}$ among products $\{j,j+1,\cdots, n\}$ that preserves same expected revenue $\tilde H_j^{k+1} = H_j^{k+1}$ (as $\sigma_j^{k+1}$ is optimal) but smaller lexicographic value than $\sigma_j^{k+1}$. Let $s_{1} = \min\{s, s\in \sigma_j^{k+1}\}$ and $\tilde s_1 = \min\{\tilde s, \tilde s\in \tilde \sigma_j^{k+1}\}$ be the item placed on the first position of  $\sigma_j^{k+1}$ and $\tilde \sigma_j^{k+1}$ respectively. There are three scenarios to consider: (i) Suppose $\tilde s_1 < s_1$, based on the dynamic programming relationship \eqref{eq:dp_recursion}, we can see
\begin{equation*}\label{eq:loptimal}
\tilde H_{j}^{k+1} = \lambda_{\tilde s_1} r_{\tilde s_1} +(1-\lambda_{\tilde s_1})H_{{\tilde s_1}+1}^{k} < \lambda_{s_1} r_{s_1} +(1-\lambda_{s_1})H_{s_1+1}^{k} = H_j^{k+1},
\end{equation*}
which contradicts the  $\cL$-optimality of $\tilde \sigma_j^{k+1}$.
(ii) Suppose $\tilde s_1 > s_1$, the lexicographic value of $\tilde \sigma_j^{k+1}$ is indeed strictly higher than $ \sigma_j^{k+1}$, which also contradicts the $\cL$-optimality of $\tilde \sigma_j^{k+1}$.
(iii) Suppose $\tilde s_1 = s_1$ that the lexicographic value of the items placed on the first position of $\sigma_j^{k+1}$ and $\tilde \sigma_j^{k+1}$ are the same. Let $\tilde \sigma_{s_1+1}^k$ be the display from 2nd to the last position of $\tilde \sigma_j^{k+1}$, we can see that $\tilde \sigma_{s_1+1}^k \neq \sigma_{s_1+1}^k$, otherwise we have $\tilde \sigma_j^{k+1}: = \{s_1\} \cup \tilde \sigma_{s_1+1}^{k} = \{s_1\} \cup  \sigma_{s_1+1}^{k} = :\sigma_j^{k+1}$.
As $ \sigma_{s_1+1}^k$ is $\cL$-optimal from the induction argument,  the lexicographic value of $\sigma_{s_1 + 1}^{k}$  is strictly less than that of $\tilde \sigma_{s_1+1}^{k}$, which also contradicts the assumption that $\tilde \sigma_j^{k+1}$ is $\cL$-optimal. Combining all three scenarios above, we conclude that $\sigma_j^{k+1}$ is also $\cL$-optimal, which completes the induction proof.
\QED

\textbf{Proof of Proposition~\ref{prop:nested}:}

In order to lay out the main idea of the proof, we first introduce some notations.
Recall $H_j^x$ is the optimal revenue for customers with attention span $x$, when the products are chosen from $\{j,j+1,\dots,n\}$.
Note that $j$ and $x$ satisfy $j+x\le n+1$ for the problem to be non-trivial. Otherwise, we just display all products.
Similarly, let $\sigma_j^x$ denote the corresponding $\cL$-optimal ranking.
By Lemma~\ref{lem:fix-m-decrease-r}, the products in $\sigma_j^x$ are ordered by decreasing prices.
We first prove the base case for the induction:

\textbf{Claim:} $\sigma_{n-2}^{1}\subset \sigma_{n-2}^2 \subset \sigma_{n-2}^{3}$.

We state the proof for the claim below.
Since the products are chosen from $\{n-2, n-1, n\}$, the ranking $\sigma_{n-2}^{3}$ includes all products and thus $\sigma_{n-2}^2 \subset \sigma_{n-2}^{3}$.
To show $\sigma_{n-2}^{1}\subset \sigma_{n-2}^2$, first consider the case $\sigma_{n-2}^1 = \{n-2\}$.
It implies that $r_{n-2}\lambda_{n-2} \geq \max\{r_{n-1}\lambda_{n-1},r_n\lambda_n\}$ because it is optimal to display product $n-2$ when customers have attention span fixed at one.
If $n-2\notin \sigma_{n-2}^2$, then we have $\sigma_{n-2}^2=\{n-1, n\}$.
By Lemma~\ref{lem:fix-m-decrease-r} and~\eqref{eq:product-indexing}, the optimal ranking has the same order as the indices of the products in it.
Therefore, with a slight abuse of notation, we simply use a set to denote the ranking, which is non-ambiguous.
If we replace product $n$ in the second position of $\sigma_{n-2}^2$ by product $n-2$, the expected revenue conditional on the event that the customer doesn't choose product $n-1$ in the first position is $r_{n-2}\lambda_{n-2}$, no less than $r_n\lambda_n$ which is the conditional expected revenue of product $n$.
Therefore, $\sigma_{n-2}^2=\{n-1, n\}$ cannot be $\cL$-optimal and we conclude that $n-2\in \sigma_{n-2}^2$.
This implies $\sigma_{n-2}^{1}\subset \sigma_{n-2}^2$.
Similar argument can be applied to the cases $\sigma_{n-2}^1 =\{n-1\}$ or $\{n\}$.
Therefore, we have completed the proof for the claim.

Now that we have proved the base case, the next lemma serves as the induction step.

Consider the $\cL$-optimal assortment $\sigma_j^x$ of choosing $x$ products from $\{j,j+1,\cdots,n\}$, as there are $n-j+1$ products available among $\{j,j+1,\cdots, n\}$, we can see $1\leq x \leq n-j+1$. The following lemma states that if the $\cL$-optimal displays among $\{j,j+1,\cdots, n\}$ preserve a nested structure, then the $\cL$-optimal displays among $\{j-1,j,j+1,\cdots, n\}$ must follow such a nested structure as well.

\begin{lemma}\label{lemma:nested}
Given $j \in \{2,3,\cdots, n-2\}$, for any $\cL$-optimal assortment $\sigma_j^x$ of choosing $x$ products from $\{j,j+1,\cdots,n\}$ with $1\leq x\leq n-j+1$,
if they are nested such that $\sigma_j^1\subset \sigma_j^2 \subset \cdots \sigma_j^{n-j+1}$
, then the $\cL$-optimal displays among products $\{j-1,j,\cdots, n\}$ are also nested, i.e.,
$\sigma_{j-1}^1 \subset \sigma_{j-1}^2 \subset \cdots \subset \sigma_{j-1}^{n-j+2}$.
\end{lemma}
With Lemma~\ref{lemma:nested}, we are able to show Proposition~\ref{prop:nested} recursively.

Before proving Lemma~\ref{lemma:nested}, we first provide a lemma characterizing the relationship between $H_j^{m}$ and $H_j^{m+1}$, which is frequently applied throughout our analysis.
\begin{lemma}\label{lemma:rev_increase}
For any $j=1,2,\cdots,n-1$ and $x=1,2,\cdots,n-j$, we have $H_j^x \leq H_j^{x+1}$.
\end{lemma}

\textit{Proof of Lemma \ref{lemma:rev_increase}:} Recall that $\sigma_j^x$ and $\sigma_j^{x+1}$ are the optimal $\cL$-optimal ranking for $x$ and $x+1$ products respectively. Let $P_0 = \prod_{s = 1}^x  (1-\lambda_{\sigma_j^x(s)})$ be the probability that no products are purchased within $\sigma_j^x$.  Without loss of generality, for any product $k \geq j, k \notin \sigma_j^{x}$, by adding this product at the last position of $\sigma_j^x$, we can construct a ranking $\tilde \sigma_j^{x+1}$ such that $\tilde \sigma_j^{x+1}(s) = \sigma_j^x(s)$ for $s=1,2,\cdots,m$ and $\tilde \sigma_j^{x+1}(x+1) = k$.
It is easy to see that
$$
R(\tilde \sigma_j^{x+1},x+1) = R(\sigma_j^x,x) + P_0 \cdot \lambda_k r_k \geq R(\sigma_j^x,x).
$$
Due to the fact $|\tilde \sigma_j^{x+1}| = x+1$ and the optimality of $\sigma_j^{x+1}$, we also have $H_j^{x+1} = R(\sigma_j^{x+1} ,x+1) \geq R(\tilde \sigma_j^{x+1},x+1)$. Combining them together, we conclude that
$$
H_j^{x+1} \geq R(\tilde \sigma_j^{x+1},x+1) \geq R(\sigma_j^x,m) = H_j^x,
$$
which completes the proof.
\QED

\begin{lemma}\label{lemma:decreased_margin}
Given $j\in\{1,\dots, n-2\}$, if $\sigma_j^1 \subset \sigma_j^2 \subset \dots \subset \sigma_j^{n-j+1}$, then we have
$$
H_j^{x+2} - H_j^{x+1} \leq H_j^{x+1} - H_j^{x},
$$
for all $x = 1,2,\ldots, n-j-1$.
\end{lemma}
\textit{Proof of Lemma \ref{lemma:decreased_margin}:}
Due to the condition of the lemma, we have $\sigma_j^x \subset \sigma_{j}^{x+1} \subset \sigma_j^{x+2} $.
Let $u$ and $v$ be the two items added to the optimal ranking for attention span $x+1$ and $x+2$ involving products $\{j,j+1,\ldots, n\}$. That is,
$$
\sigma_{j}^{x+1} = \sigma_j^{x} \cup \{u\}\text{ and } \sigma_{j}^{x+2} = \sigma_j^{x+1} \cup \{v\}.
$$
Again, we treat a ranking as a set without ambiguity because of Lemma~\ref{lem:fix-m-decrease-r}.
Suppose on the contrary the result does not hold, that is,
\begin{equation}\label{eq:reverse_ineq}
H_j^{x+2} - H_j^{x+1} > H_j^{x+1} - H_j^{x}.
\end{equation}
We consider the two scenarios $u < v$ and $u > v$ separately.
Also recall that $H_j^k = R(\sigma_j^k, k)$ for $k=x,x+1,x+2$.

\begin{itemize}
    \item
Suppose $u < v$ and let $x_v$ be the number of items displayed in $\sigma_j^{x+2}$ after item $v$.
By Lemma~\ref{lem:fix-m-decrease-r}, the products displayed after item $v$ are chosen from $\{v+1,v+2,\ldots, n\}$.
Moreover, they must generate the optimal revenue conditional on the event that the customer does not purchase any product after viewing product $v$.
Therefore, the sub-ranking after $v$ is $\sigma_{v+1}^{x_v}$ and the conditional expected revenue is $H_{v+1}^{x_v}$.

Let $P_1$ be the probability that a customer views product $v$ in the ranking $\sigma_j^{x+2}$.
Suppose product $v$ is inserted in the $i_v$-th position of $\sigma_j^{x+2}$, i.e., $\sigma_j^{x+2}(i_v) = v$.
Due to the fact that $\sigma_j^{x+2} = \sigma_j^{x+1} \cup \{v\}$ and $v$ is inserted after $u$ (because $u<v$ and Lemma~\ref{lem:fix-m-decrease-r}), the probability of viewing the $i_v$-th item in $\sigma_{j}^{x+1}$ is also $P_1$.
The expected revenues generated before the $i_v$-th product are identical in $\sigma_{j}^{x+1}$ and $\sigma_{j}^{x+2}$.
Thus, we have
\begin{equation}\label{eq:hm+2-hm+1}
H_j^{x+2}- H_j^{x+1}= P_1 \cdot \Big ( \lambda_v r_v  +(1-\lambda_v)H_{v+1}^{x_v} - H_{v+1}^{x_v}\Big ) = P_1 \cdot \Big ( \lambda_v r_v  -\lambda_v H_{v+1}^{x_v}\Big ).
\end{equation}
Moreover, combining \eqref{eq:hm+2-hm+1} with Lemma~\ref{lemma:rev_increase}, we can see that $r_v \geq H_{v+1}^{x_v}$ because $P_1>0$ and $\lambda_v>0$.

Let $\tilde \sigma_j^{x+1} = \sigma_j^x \cup \{v\}$.
Note that $\tilde \sigma_j^{x+1}$ includes $x+1$ products and it is not the optimal ranking.
But we arrange the product in the ascending order of their indices as Lemma~\ref{lem:fix-m-decrease-r}.
Let $\tilde P_1$ be the probability of viewing product $v$ in the ranking $\tilde \sigma_j^{x+1}$ for a customer of attention span $x+1$.
Comparing $\tilde \sigma_j^{x+1}$ and $\sigma_j^{x+2}$, the product $u$ is inserted to $\tilde \sigma_j^{x+1}$ to turn it into  $\sigma_i^{x+2}$.
Because $u < v$, we have $P_1 = (1-\lambda_u) \tilde P_1\le \tilde P_1$.
Note that the expected revenue conditional on the event that a customer doesn't purchase after viewing product $v$ in $\tilde \sigma_j^{x+1}$ is still $H_{v+1}^{x_v}$ as the sub-ranking after $v$ remains the same as $\sigma_j^{x+2}$.

Now consider the expected revenue of $\tilde \sigma_j^{x+1}$ and $\sigma_j^x$, which removes product $v$ in the former ranking.
We have
\begin{align*}
& R(\sigma_i^{x} \cup \{v\},x+1) - H_i^x = \tilde P_1 \Big ( \lambda_v r_v +(1-\lambda_v)H_{v+1}^{x_v} -H_{v+1}^{x_v}\Big )= \tilde P_1 \Big ( \lambda_v r_v  -\lambda_v H_{v+1}^{x_v}\Big ) \\
 \geq & P_1 \Big ( \lambda_v r_v  -\lambda_v H_{v+1}^{x_v}\Big ) = H_{j}^{x+2}- H_j^{x+1} > H_{j}^{x+1}- H_j^{x},
\end{align*}
where the first inequality follows from $ P_1 \le \tilde P_1  $ and $r_v  - H_{v+1}^{x_v} \geq 0$, the last equality follows from \eqref{eq:hm+2-hm+1}, and the last inequality follows from \eqref{eq:reverse_ineq}.
Therefore, we have $R(\sigma_j^x \cup \{v\},x+1) > H_j^{x+1})$, which contracts the $\cL$-optimality of $\sigma_j^{x+1}$.
\item
Next consider the case $v < u$.
Suppose there are $x_v$ items placed after $v$ in $\sigma_j^{x+2}$.
Similarly, denote the sub-ranking as $\sigma_{v+1}^{x_v}$ and the expected reward as $H_{v+1}^{x_v}$.
Define $P_0$ as the probability of viewing item $v$ under $\sigma_j^{x+2}$.
Similar to \eqref{eq:hm+2-hm+1}, we have
$$
H_j^{x+2} - H_j^{x+1}= P_0 (\lambda_v r_v - \lambda_v H_{v+1}^{x_v}).
$$
Consider the ranking $\tilde \sigma_j^{x+1} = \sigma_j^x \cup \{v\}$.
The sub-ranking after $v$ in $\tilde \sigma_j^{x+1}$ is $\sigma_{v+1}^{x_v}\setminus\{u\}$, or equivalently $\sigma_{v+1}^{x_v -1}$.
By Lemma \ref{lemma:rev_increase}, we have $H_{v+1}^{x_v -1} \leq H_{v+1}^{x_v}$.
Therefore, similar to \eqref{eq:hm+2-hm+1}, we have
\begin{equation*}
\begin{split}
& R(\sigma_j^x \cup \{v\},x+1) - H_j^x\\
= & P_0 (\lambda_v r_v - \lambda_v H_{v+1}^{x_v -1} ) \geq P_0 (\lambda_v r_v - \lambda_v H_{v+1}^{x_v} ) = H_j^{x+2} - H_j^{x+1}> H_j^{x+1} - H_j^{x},\end{split}
\end{equation*}
which implies that $R(\sigma_j^x \cup \{ v\},x+1) > H_j^{x+1}$ and contradicts the $\cL$-optimality of $\sigma_j^{x+1}$.
\end{itemize}
Combining both cases, we conclude $H_j^{x+2} - H_j^{x+1} \leq H_j^{x+1} - H_j^{x} $, which completes the proof.
\QED

\textit{Proof of Lemma \ref{lemma:nested}:}
First, consider $\sigma_{j-1}^1$ which has a single product.
Suppose $\sigma_{j-1}^1 = \{u\}$ where $u\ge  j-1$. We have $r_{u} \lambda_{u} \geq \max_{k \geq j-1} \{r_k \lambda_k\}$ and
$r_{u} \lambda_{u} > \max_{k=j-1,\dots,u-1} \{r_k \lambda_k\}$
because of the $\cL$-optimality of $\sigma_{j-1}^1$.
Therefore, if $u \notin \sigma_{j-1}^2$, then we can replace the last product in $\sigma_{j-1}^2$ by $u$ to increase the expected revenue of $H_{j-1}^2$ or make it lexicographically less.
As a result, we must have $u \in \sigma_{j-1}^2$ and $\sigma_{j-1}^1 \subset \sigma_{j-1}^2$.

Next, we prove $\sigma_{j-1}^x \subset \sigma_{j-1}^{x+1}$ for $2 \leq x \leq n-j$.
We first consider the case $j-1\notin \sigma_{j-1}^x$.
In this case, we have $\sigma_{j-1}^x = \sigma_j^x$ by their definitions.
If $j-1\notin\sigma_{j-1}^{x+1}$, then we have $\sigma_{j-1}^{x+1} = \sigma_j^{x+1}$ and thus $\sigma_{j-1}^x \subset \sigma_{j-1}^{x+1}$ by the induction hypothesis.
If $j-1\in\sigma_{j-1}^{x+1}$, then by the $\cL$-optimality, we must have that product $j-1$ is displayed in the first position of $\sigma_{j-1}^{x+1}$.
The sub-ranking from the second position onward in $\sigma_{j-1}^{x+1}$ must be equivalent to $\sigma_{j-1}^x$ to maximize the expected revenue.
Therefore, we have $\sigma_{j-1}^x \subset \sigma_{j-1}^{x+1}$.

Now suppose $j-1\in \sigma_{j-1}^x$.
Since $\sigma_{j-1}^x$ is $\cL$-optimal, $j-1$ must be ranked the first and thus
\begin{equation*}
H_j^x\le H_{j-1}^x = \lambda_{j-1} r_{j-1} + (1-\lambda_{j-1}) H_j^{x-1} .
\end{equation*}
If ${j-1} \in \sigma_{j-1}^{x+1}$, then $\sigma_{j-1}^x \subset \sigma_{j-1}^{x+1}$ can be proved by the induction hypothesis on the sub-rankings of $\sigma_{j-1}^x$ and $\sigma_{j-1}^{x+1}$ from the second position onward.
Suppose $j-1 \notin \sigma_{j-1}^{x+1}$.
Based on the induction hypothesis that $\sigma_j^{x-1} \subset \sigma_j^x \subset \sigma_j^{x+1}$, there exist items $u,v \notin \sigma_j^{x-1}$ such that
\begin{align}\label{eq:m-1+s+j-m+1}
\sigma_j^x = \sigma_j^{x-1} \cup \{u\}, \,\,\, \sigma_j^{x+1} = \sigma_j^x \cup \{ v\}.
\end{align}
Since we have assumed that $j-1 \notin \sigma_{j-1}^{x+1}$,
we must have $\sigma_{j-1}^{x+1}=\sigma_{j}^{x+1}$.
Suppose that in $\sigma_{j-1}^{x+1}$ or $\sigma_{j}^{x+1}$ there are $x_u$ and $x_{v}$ items placed after items $u$ and $v$ respectively.
Recall that $H_{u+1}^{x_u}$ and $H_{v+1}^{x_{v}}$ are the expected revenue obtained from items after $u$ and $v$.
Since $\sigma_{j-1}^{x+1}$ is $\cL$-optimal, the expected revenue strictly decreases if one replaces either $u$ or $v$ in $\sigma_{j-1}^{x+1}$ by item $j-1$.
Conditional on the event that the customer views product $u$ (or $j-1$), we have
\begin{equation*}
\begin{split}
\lambda_{j-1} r_{j-1} + (1-\lambda_{j-1}) H_{u+1}^{x_u} <& \lambda_u r_u+ (1-\lambda_u) H_{u+1}^{x_u}, \\
\lambda_{j-1} r_{j-1} + (1-\lambda_{j-1}) H_{v+1}^{x_{v}} < & \lambda_{v} r_{v}+ (1-\lambda_{v}) H_{v+1}^{x_{v}}.
\end{split}
\end{equation*}
Note that $r_{j-1} \geq r_u \geq H_{u+1}^{x_u}$ (the first inequality is because of the indexing of the products; the second inequality is because all products in $\sigma_{u+1}^{x_u}$ have revenues lower than $r_u$) and $r_{j-1} \geq r_{v} \geq H_{v+1}^{x_{v}}$.
Therefore, we have $\lambda_{j-1} < \lambda_u$ and $\lambda_{j-1} < \lambda_{v}$, which implies
\begin{equation}\label{eq:lambda_order}
1-\lambda_{j-1} > 1-\lambda_u, \text{ and }1-\lambda_{j-1} > 1-\lambda_{v}.
\end{equation}

Let $r^*:= r_{j-1}\lambda_{j-1} + (1-\lambda_{j-1}) H_j^{x}$ be the expected revenue obtained by inserting item ${j-1}$ in the first position of ranking $\sigma_{j}^{x}$.
Since $j-1\notin\sigma_{j-1}^{x+1}$, we have $r^* < H_j^{x+1}$. We will show that $r^* \geq H_j^{x+1}$ in the rest of analysis to yield a contradiction which proves that $j-1\in\sigma_{j-1}^{x+1}$.

Recall that $H_{j-1}^x = r_{j-1}\lambda_{j-1} + (1-\lambda_{j-1}) H_j^{x-1}$.
By definition, we have
\begin{equation}\label{eq:r_ast}
r^* = r_{j-1}\lambda_{j-1} + (1-\lambda_{j-1}) H_j^{x} = H_{j-1}^x + (1-\lambda_{j-1}) (H_j^x - H_j^{x-1}).
\end{equation}
As $\sigma_j^{x-1}\cup \{v\}$ is not $\cL$-optimal for $\sigma_j^x$, we also have
\begin{align*}
R(\sigma_j^{x-1}\cup \{v\}, x) \leq R(\sigma_j^{x}, x) = H_j^x.
\end{align*}
We consider two cases: $u<v$ and $u > v$.
\begin{enumerate}
\item[(i)] Suppose $u <v$.
Note that the difference $H_{j}^{x+1} - H_{j}^x$ is only caused by the products displayed after the inserted position of product $v$ (see \eqref{eq:m-1+s+j-m+1}).
Therefore, we have
\begin{equation}\label{eq:case_1}
\begin{split}
H_{j}^{x+1} - H_{j}^x  = & \Pr( \text{viewing product } v \text{ in }\sigma_j^{x+1} ) \cdot (\lambda_{v} r_{v} +(1-\lambda_{v})H_{v+1}^{x_v}- H_{v+1}^{x_{v}})\\
= & \Pr( \text{viewing product } v \text{ in }\sigma_i^{x+1}\setminus\{u\})  (1-\lambda_u)   \lambda_{v}  (r_{v} - H_{{v}+1}^{x_{v}}) \\
= &  (1-\lambda_u) \cdot \big (R(\sigma_j^{x-1}\cup \{v\},x)  - H_j^{x-1} \big ) \\
\leq & (1-\lambda_u) (H_j^x - H_j^{x-1}),
\end{split}
\end{equation}
In the first equality, we condition on the event that a customer  views product $v$ in $\sigma_j^{x+1}$.
In the second equality, because $u<v$, product $u$ is displayed before $v$ in $\sigma_{j}^{x+1}$ by Lemma~\ref{lem:fix-m-decrease-r}.
In the third equality, after excluding product $u$, we are comparing the revenues of $\sigma_j^{x-1}$ and $\sigma_j^{x-1}\cup \{v\}$.
The inequality due to the suboptimality of $\sigma_{j}^{x-1}\cup \{j\}$ in \eqref{eq:m-1+s+j-m+1}.
By \eqref{eq:r_ast}, we have
\begin{align*}
r^* = & H_{j-1}^x + (1-\lambda_{j-1}) (H_j^x - H_j^{x-1}) > H_{j-1}^x + (1-\lambda_u) (H_j^x - H_j^{x-1}) \\
\geq & H_j^x + (H_j^{x+1} - H_j^x) = H_j^{x+1},
\end{align*}
where the first inequality follows from \eqref{eq:lambda_order}, and the second inequality follows from \eqref{eq:case_1} and $H_{j-1}^x \geq H_j^x$. Therefore, we have $r^* > H_j^{x+1}$ if $u<v$.
\item[(ii)]  Suppose $v < u$.
Similar to \eqref{eq:case_1}, we have
\begin{equation}\label{eq:case_2}
\begin{split}
& R(\sigma_j^{x-1} \cup \{v,u\},x+1) - R(\sigma_j^{x-1} \cup \{v\},x)
=   H_j^{x+1} - R(\sigma_j^{x-1} \cup \{v\},x) \\
= & \Pr(\text{viewing product } u \text{ in } \sigma_j^{x+1}) \cdot \lambda_u (r_u - H_{u+1}^{x_u})\\
= & (1-\lambda_{v}) \Pr(\text{viewing product } u \text{ in } \sigma_j^{x+1}\setminus\{v\}) \cdot  \lambda_u (r_u - H_{u+1}^{x_u})
\\
= & (1-\lambda_{v}) \cdot \Big (R(\sigma_j^{x-1} \cup \{u\},x) - H_j^{x-1} \Big) \\
= &  (1-\lambda_{v})(H_j^x - H_j^{x-1}).
\end{split}
\end{equation}
Again, by \eqref{eq:r_ast}, we have
\begin{align*}
r^* = & H_{j-1}^x + (1-\lambda_{j-1}) (H_j^x - H_j^{x-1}) \\
> &  H_{j-1}^x + (1-\lambda_{v}) (H_j^x - H_j^{x-1}) \\
= & H_{j-1}^x + H_j^{x+1} - R(\sigma_j^{x-1} \cup \{v\},x) \ge H_j^{x+1},
\end{align*}
where the first inequality follows from $1-\lambda_{j-1} > 1-\lambda_v$ in \eqref{eq:lambda_order}, the second equality is by \eqref{eq:case_2}, and the third inequality follows from $R(\sigma_j^{x-1} \cup \{v\},x)  \leq H_{j-1}^x$ (suboptimality of $\sigma_j^{x-1} \cup \{v\}$ for $\sigma_{j-1}^x$). We conclude that $r^* > H_j^{x+1}$ also holds if $v < u$.
\end{enumerate}
Combining both cases, we have $r^* > H_j^{x+1}$, which yields a contradiction. Therefore, we conclude that $j-1 \in \sigma_{j-1}^{x+1}$.
By the previous analysis, it implies that $\sigma_{j-1}^x\subset \sigma_{j-1}^{x+1}$ and thus $\sigma_{j-1}^{1}  \subset \sigma_{j-1}^{2} \subset \cdots \subset \sigma_{j-1}^{n-j+1}$.
To see $\sigma_{j-1}^{n-j+1} \subset \sigma_{j-1}^{n-j+2}$, note that $\sigma_{j-1}^{n-j+2} = \{j-1,j,\cdots,n\}$ is the full collection of products and the inclusion holds automatically.
This completes the proof.
\QED

\proof{Proof of Proposition~\ref{prop:upper-bound}:}
The proof is given in the text before the proposition.
\QED
\endproof

\textbf{Proof of Theorem \ref{thm:approx}:} To show Theorem \ref{thm:approx}, we first prove Lemma~\ref{lemma:max_y}, Lemma~\ref{lm:geom} and Lemma~\ref{lm:Delta} that show the optimal structures of the $R(\cdot)$ revenue function and the attention span distribution of $X$ in the worst case.
We assume $R(1)>0$ because otherwise $R(k)\equiv 0$ and it is a trivial case to verify.

\begin{lemma} \label{lemma:max_y}
Let $f(k) \triangleq G_k R(k)$ for $k \in \{1,2,\ldots,M\}$. 
We can show $f(k)$ is a unimodal function, and there exist at most two optimal solutions to $\max_k \, f(k)$. 
\end{lemma}
\proof{Proof:}
Note that because $f(1)=G_1R(1)>0$, the optimal value of $f(k)$ is positive.
It is also clear that when $f(k)=G_k=0$, we have $f(k+1)=\dots=f(M)=0$.

To show the result, it suffices to show that for any $k$ such that $0<G_{k} R(k) \le G_{k-1}R(k-1)$, then $G_{k+1}R(k+1) < G_k R(k)$.
Because $R(k)$ is a concave increasing function in $k$, if we let $\Delta = R(k) - R(k-1) \geq 0$, we always have $R(k+1) - R(k) \leq \Delta$. Then having
$$
G_k R(k)= G_{k-1} R({k-1}) \frac{R({k-1}) + \Delta}{R({k-1})} (1-h_{k-1}) \leq G_{k-1} R({k-1})
$$
implies that 
$\frac{R({k-1}) + \Delta}{R({k-1})} (1-h_{k-1}) \leq 1$. 
If it is strictly less than one, then we have
\begin{align*}
    G_{k+1} R({k+1}) &= G_{k} R({k} )\frac{R({k+1} )}{R({k})} (1-h_{k}) \\
    &\leq G_{k} R({k}) \frac{R({k}) + \Delta }{R({k})} (1-h_{k-1})\\
    &\leq G_{k} R({k}) \frac{R({k-1}) + \Delta }{R({k-1})} (1-h_{k-1}) \\
    &<  G_{k} R({k}),
\end{align*}
where the first inequality uses the facts that $R({k+1}) = R({k}) + (R({k+1}) - R(k)) \leq R(k) + \Delta$ and $h_{k} \geq h_{k-1}$, the second inequality uses the fact that $R(k) \ge R({k-1})$. 
On the other hand, suppose $\frac{R({k-1}) + \Delta}{R({k-1})} (1-h_{k-1}) = 1$. 
Because of IFR, we have $\frac{R({k-1}) + \Delta}{R({k-1})}>1$ and $h_{k-1}>0$.
As a result, we have $\Delta>0$, which implies $R(k)>R(k-1)$ and $\frac{R({k}) + \Delta }{R({k})}<\frac{R({k-1}) + \Delta }{R({k-1})}$.
Following the same reasoning, the second inequality above becomes strict and this
completes the proof of Lemma~\ref{lemma:max_y}.
\QED

\begin{lemma}\label{lm:geom}
For any $R(x): x \in \N_+ \to \RR_+$, consider $\min_{X} \max_{y }\frac{R(y) G_y}{\EE[R(X)]}$, the minimizer $X^*$ must have a geometric distribution. 
\end{lemma}
\proof{Proof:} Recall that $X$ fulfills the IFR condition that $h(1) \leq h(2) \leq \cdots$. Suppose $X^*$ does not have a geometric distribution, which means there exist at least two integers $i$ and $i+1$ such that $h_{i} < h_{i+1}$. Let $y^*$ be the largest index that maximizes $R(y)G_y$ (According to Lemma~\ref{lemma:max_y}, when there are two optimal solutions, they must be $y^*-1$ and $y^*$). We consider two cases:
\begin{enumerate}
    \item $i+1 \leq y^*-1$. We write $R(y^*)G_{y*} =R(y^*) \Pi_{x=1}^{y^*-1} (1-h_x)$. 
    Suppose we increase $h_i$ by $\delta >0 $ such that $h_i + \delta \leq h_{i+1}$. By Lemma~\ref{lemma:max_y}, we can see that such an increasing of $h_i$ would not change the optimality of $y^*$ (or $y^{*}-1$ when there are two optimizers) when $\delta$ is sufficiently small. 
 Therefore, the term $R(y^*) G_{y^*}$ decreases to $R(y^*) G_{y^*}(\delta) = R(y^*) (1-h_x - \delta) \Pi_{x=1, x \neq i}^{y^*-1} (1-h_x)  $. Meanwhile, recall that 
    $g_x  = h_x \Pi_{k=1}^{x-1} (1-h_k) $, we have 
    \begin{equation*}
        \begin{split}
        & \EE[R(X) ] = \sum_{x=1}^\infty R(x) g_x 
        \\
        & =  \sum_{x=1}^{i-1} R(x) g_x + \Pi_{k=1}^{i-1}(1-h_k)R(i)h_i  + \Pi_{k=1}^{i-1}(1-h_k) (1-h_i  ) \sum_{x=i+1}^\infty R(x)    \Pi_{k= i+1}^{x-1} (1-h_k) h_{x}     
        \\
        & = C_0 + C_1 R(i)h_i  + C_1 (1-h_i) C_2 
        \end{split}
    \end{equation*}
    where $C_0 = \sum_{x=1}^{i-1} R(x) g_x$, $C_1 = \Pi_{k=1}^{i-1}(1-h_k)R(i)$, and $C_2 = \sum_{x=i+1}^\infty R(x)    \Pi_{k= i+1}^{x-1} (1-h_k) h_{x}$ are constants. 
By increasing $h_i $ to $h_i + \delta$, the above term becomes 
\begin{equation*}
  \EE[R(X); \delta ] =   C_0 + C_1 R(i)(h_i + \delta)   + C_1 (1-h_i - \delta) C_2. 
\end{equation*}
Now we consider the objective $R(y^*) G_{y^*} /\EE[R(X)]$, which now becomes 
\begin{equation*}
    \begin{split}
       &  \frac{R(y^*) G_{y^*}(\delta)}{\EE[R(X) ;\delta]} = \frac{ R(y^*) \Pi_{x=1}^{y^*-1} (1-h_x ) \cdot  (1-h_i -\delta)/(1-h_i) }{C_0 + C_1 R(i)(h_i + \delta)   + C_1 (1-h_i - \delta) C_2  } 
        \\
        & =       \frac{ R(y^*) \Pi_{x=1}^{y^*-1} (1-h_x ) /(1-h_i) }{C_0 /(1-h_i - \delta) + C_1 R(i)(h_i + \delta) /(1-h_i - \delta)  + C_1  C_2  } 
        \\ 
       &  <  \frac{ R(y^*) \Pi_{x=1}^{y^*-1} (1-h_x ) /(1-h_i) }{C_0 /(1-h_i ) + C_1 R(i)h_i /(1-h_i )  + C_1  C_2  }  = \frac{R(y^*) G_{y^*}}{\EE[R(X) ]} 
    \end{split}
\end{equation*}
because $1-h_i - \delta < 1-h_i$. Therefore, we strictly decrease the objective value, violating the optimality of $X$. 

    \item $i +1 \geq y^*$. In this case, because $R(y^*) G_{y^*} = R(y^*) \Pi_{x\leq y^*-1} (1-h_x)$, slightly decreasing the value of $h_{i+1}$ would not affect the optimal value $R(y^*) G_{y^*}$. However, we can see that $g_{i+1} = \Pi_{x=1}^i (1-h_x) h_{i+1}$ strictly decreases to $\Pi_{x=1}^i (1-h_x ) ( h_{i+1} -\delta) $ but $g_{i+2} =  \Pi_{x=1}^{i+1} (1-h_x) h_{i+2}$ strictly increases to $ \Pi_{x=1}^{i} (1-h_x) (1-h_{i+1} + \delta) h_{i+2}$. Similarly, we can also see that all items $g_{i+2},g_{i+3},\cdots$ strictly increase. Because $R(x)$ is an increasing function and $\sum_{x=i+1}^\infty g_x$ remains unchanged, it is easy to see that $\EE[R(X)]$ strictly increases. Therefore, the objective strictly decreases, violating its optimality. This also indicates that the optimal number of slots should be unlimited, i.e., $M\rightarrow +\infty.$
\end{enumerate}
Combining the two cases, we can see that $X^*$ must have an untruncated geometric distribution. 
This concludes the proof of Lemma~\ref{lm:geom}.
\QED  

\begin{lemma}\label{lm:Delta}
    For any $X$, the worst case $R(\cdot)$ function that solves $\min_{R,X} \max_{y }\frac{R(y) G_y}{\EE[R(X)]}$  is obtained when $R(x)$ takes the form 
    $
    R(x) = R(1)  + \Delta (x -1)
    $
    for some $\Delta$ such that $0 \leq \Delta \leq R(1)$.
\end{lemma}
\proof{Proof:}
Again, let $y^*$ be the largest index of the maximizers for $R(k)G_k$, i.e., $y^*=\max\{ \argmax_k R(k)G_k \}$. According to Lemma~\ref{lemma:max_y} there are at most two optimizers of $R(k)G_k$, and given one is $y^*$, the only other possible optimizer is $y^*-1$.

Now let us denote $\Delta \triangleq R(y^*)-R(y^*-1)$. We want to show the revenue function $R(x)$ in the worst case, i.e., the solution for $\min_{R,X} \max_{y }\frac{R(y) G_y}{\EE[R(X)]}$, has the following structure:
\begin{enumerate}
    \item[i)] $\forall x>y^*$, $R(x)=R(y^*)+(x-y^*)\Delta$;
    \item[ii)] $\forall x<y^*$, $R(x)=R(y^*)-(y^*-x)\Delta$.
\end{enumerate}
In other words, $R(x)-R(x-1)=\Delta$, $\forall x=2,\ldots,M$, or equivalently, $R(x)=R(1)+(x-1)\Delta, \forall x=1,\ldots,M$ and by the concavity of $R(\cdot)$, we have $R(1) \geq \Delta$. As a consequence, Lemma~\ref{lm:Delta} will hold.

Next we prove claim i) and ii) of the optimality structure of $R(x)$.
For claim i), suppose it is not the case. Denote $x$ as the smallest index that violates the condition, so $R(x)\neq R(y^*)+(x-y^*)\Delta$. By the concavity of $R(\cdot)$, it must be $R(x)< R(y^*)+(x-y^*)\Delta$. In this case, we can simultaneously increase the value of $R(x),R(x+1),\ldots,R(M)$ by a small enough quantity, but without changing the optimality of $y^*$ or violating any constraint. This is because by Lemma~\ref{lemma:max_y}, $R(x)G_x<R(y^*)G_{y^*}$ for all $x>y^*$. However, the denominator in $\frac{R(y) G_y}{\EE[R(X)]}$ strictly increases, contradicting the optimality of $R(\cdot)$.

For claim ii), following the same logic, suppose it is not the case. Denote $x$ as the largest index that violates the condition, so $R(x)\neq R(y^*)-(y^*-x)\Delta$. By the decreasing margin property of $R(\cdot)$, it must be $R(x)< R(y^*)-(y^*-x)\Delta$. In this case, we can simultaneously increase the value of $R(1),R(2),\ldots,R(x)$ by a small enough quantity, but without changing the optimality of $y^*$ (or $y^*$ and $y^*-1$) or violating any constraint. This is because by Lemma~\ref{lemma:max_y}, $R(x)G_x<R(y^*)G_{y^*}$ for all $x<y^*-1$. However, the denominator in $\frac{R(y) G_y}{\EE[R(X)]}$ strictly increases, contradicting the optimality of $R(\cdot)$.

    Combining these two cases, we conclude that at the optimal solution, $R(x) = R(1) + \Delta (x-1)$ for some $\Delta$ such that $0 \leq \Delta \leq R(1)$.
    This concludes the proof of Lemma~\ref{lm:Delta}.
\QED

Now we proceed to combine the lemmas above and prove Theorem~\ref{thm:approx}.
By Lemma~\ref{lm:geom} and \ref{lm:Delta}, we can show the problem $\min_{R,X} \max_{y }\frac{R(y) G_y}{\EE[R(X)]}$ will be reduced to 
\begin{align} \label{eq:red}
    \min_{R(1),\Delta, p} \max_{y }\, & \frac{\big(R(1)+(y-1)\Delta\big)(1-p)^{y-1}}{R(1)+(\EE[X]-1)\Delta}\\
    \mbox{s.t.}\quad & 0 \leq \Delta \leq R(1), \nonumber \\
    & \EE[X]=1/p. \nonumber
\end{align}
This is because the numerator
\begin{align*}
    R(y)G(y)&=R(y) (1-p)^{y-1}  &\mbox{(by Lemma~\ref{lm:geom})}\\
    &=\big(R(1)+(y-1)\Delta\big)(1-p)^{y-1}, & \mbox{(by Lemma~\ref{lm:Delta})}
\end{align*}
and the denominator
\begin{align*}
    \EE[R(X)]&=\sum_{x=1}^{\infty} R(x)p(1-p)^{x-1} &\mbox{(by Lemma~\ref{lm:geom})}\\
    &=\sum_{x=1}^{\infty} \big(R(1)+(x-1)\Delta\big)p(1-p)^{x-1} & \mbox{(by Lemma~\ref{lm:Delta})}\\
    &=R(1)+(\EE[X]-1)\Delta.
\end{align*}
We now show problem~(\ref{eq:red}) has a lower bound $1/e$ and the bound is attainable when $R(x)=x\Delta,\, \forall x=1,\cdots, M$, the number of slots $M\rightarrow \infty$ and the attention span $X$ has a geometric distribution with probability $p \rightarrow 0$.

Lemma~\ref{lm:geom} has provided that the worst-case distribution of $X$ is geometric distribution. 
Meanwhile, Lemma~\ref{lm:Delta} indicates that the optimal value of $R(x)=R(1)+(x-1)\Delta$. Therefore, we should prove that in the worst case, $R(1)=\Delta$ and the success probability $p$ in the geometric distribution satisfies $p \rightarrow 0$.

First, let us show that $R(1)=\Delta$ in the worst case. Note the objective function in (\ref{eq:red}) is a decreasing function of $R(1)$ if $y>\EE[X]$ and an increasing function of $R(1)$ otherwise. 
We analyze how a potential choice of $y$'s value would affect the entire optimization result, based on the two cases: i) $y>\E[X]$ and ii) $y \leq \EE[X]$.

For i), given we choose a $y>\E[X]$, the objective of (\ref{eq:red}) will decrease in $R(1)$. As a result, the worst case will set $R(1)$ at $+\infty$ to minimize the objective. In this case, the entire problem reduces to
$$\min_p\, \max_{y>\EE[X]}\, (1-p)^{y-1}.$$
Because $(1-p)^{y-1}$ monotonically decreases in $y$, the optimal $y$ must be set at $\floor{\EE[X]+1}=\floor{\frac{1}{p}+1}$, where $\floor{a}\triangleq\max \{i \in N, \, i\leq a\}$.

However, if we look at the problem $\min_p \, (1-p)^{\floor{\frac{1}{p}+1}-1}$, it will converge to 0 when $p\rightarrow 1$, i.e., $\lim_{p \rightarrow 1} (1-p)^{\floor{\frac{1}{p}+1}-1}=0$, indicating $y>\E[X]$ may not be a good choice. 

For ii), we choose $y \leq \EE[X]$, so the objective is an increasing function in $R(1)$. Since we require $R(1) \geq \Delta$, the worst case $R(1)$ will be set at $\Delta$ to minimize the objective. Therefore, the whole problem boils down to
\begin{align}\label{eq:red2}
    \min_{p}\, \max_{y \leq \EE[X]=1/p}\, \frac{y(1-p)^{y-1}}{\EE[X]}=yp(1-p)^{y-1}.
\end{align}
Given the probability $p$, the inner problem $\max_{y \leq 1/p}\, yp(1-p)^{y-1}$ can be easily solved. However, instead of picking the optimal $y$, let us set $y=\floor{\frac{1}{p}}$. Obviously the constraint $y \leq \EE[X]$ is satisfied. Now we focus on the problem
\begin{equation}\label{eq:red3}
    \min_p \, p \floor{\frac{1}{p}} (1-p)^{\floor{\frac{1}{p}}-1}.
\end{equation}
Note that the $\floor{\frac{1}{p}}$ causes non-smoothness to the objective function. However, when $\floor{\frac{1}{p}} \leq \frac{1}{p} < \floor{\frac{1}{p}}+1$, the objective (\ref{eq:red3}) is smooth and monotonically increases in $p$. So to minimize the objective, $p$ should be reduced to the lowest value in the interval so that $\frac{1}{p}$ is arbitrarily close to $\floor{\frac{1}{p}}+1$. Let us replace the $\floor{\frac{1}{p}}$ in (\ref{eq:red3}) by $\frac{1}{p}-1$. Therefore, the problem finally reduces to $$\min_p \, (1-p)^{\frac{1}{p}-1}.$$
Here we discard the constraint that $\frac{1}{p}$ should be arbitrarily close to $\floor{\frac{1}{p}}+1$. Note this will relax the problem and make the bound even worse. However, later we will see this indeed returns a tight bound even with the relaxation.

Looking back to the reduced problem, it is easy to check that $ (1-p)^{\frac{1}{p}-1}$ monotonically increases in $p$ so the optimal $p$ should converge to 0, while $$\lim_{ p \rightarrow 0} \, (1-p)^{\frac{1}{p}-1}=\frac{1}{e}.$$
Therefore, we show that by setting $y=\floor{\frac{1}{\EE[X]}}$ in the worst case, we can achieve performance guarantee $\frac{1}{e}$.

Finally, we show the tightness of the $\frac{1}{e}$ bound in the earlier derivation.
Note that we made two relaxations: first, in the choice of $y$, when solving the inner optimization problem of \eqref{eq:red2}, instead of picking the optimal $y$, we just set it at a potentially sub-optimal value $\floor{\frac{1}{\EE[X]}}$. 
Second, for a tight bound, $\frac{1}{p}$ should be arbitrarily close to $\floor{\frac{1}{p}}+1$ when solving \eqref{eq:red3}. 

To demonstrate tightness of the $\frac{1}{e}$ bound, we only need one sample set of input for the $R(\cdot)$ function and the attention span distribution and show that the best achievable performance under this set of input is $1/e$. Now, let us set $R(x)=x\Delta,\, \forall x=1,2,\ldots$ and $X$ follows $\text{Geom}(p)$ with $p \rightarrow 0$. Therefore, the bound $1/e$ is tight once we can prove $$\max_y \, p y (1-p)^{y-1} \leq 1/e,$$
when $p\rightarrow 0$. Note that the optimal $y$ that maximizes $p y (1-p)^{y-1}$ under a given $p$ is $y^*(p)=-\frac{1}{\log(1-p)}.$ It is easy to show that 
$$\lim_{p \rightarrow 0} \, p y^*(p) (1-p)^{y^*(p)-1}=\frac{1}{e}.$$
Thus complete the proof of Theorem~\ref{thm:approx}.
\QED 
\endproof

\proof{Proof of Proposition~\ref{prop:tightness2}:}
We consider a sufficiently large $M$.
Moreover, assume consumer attention spans have a geometric distribution with success probability equal to $1-\alpha$. That is, $\PP(X \geq k) = \alpha^{k-1}$ for $1 \leq k \leq M$.
It is clear that $X$ satisfies Assumption~\ref{assump:IFR}.
Product one has $r_1=1$ and $\lambda_1=1$; and for the rest of the products $r_j=r>1$, $\lambda_j=\lambda$, and $r_j \lambda_j=1-\alpha$ for all $j$.
We also have $1 = \lambda_1 r_1 = \frac{\lambda_j r_j}{1 - \alpha } >  \frac{\lambda_j r_j}{1 - \alpha (1-\lambda_j)}$ for $j\ge 2$.

Suppose the optimal ranking contains product one.
By Proposition~\ref{prop:geo_discount} presented in Section 4.3, product one must be placed on the first position.
    Because $\lambda_1 = 1$, any random customer would not continue to view products beyond that, so that the expected revenue for this ranking is 1.
    Suppose on the contrary that the optimal ranking does not contain product 1, because other products are identical, an optimal ranking is to place product $j$ on position $j-1$, which yields an expected revenue
\begin{equation*}
\begin{split}
 & r_{2} \lambda_{2}  + (1-\lambda_2) \lambda_3 r_3 G_2 + \cdots  + \prod_{j=2}^{M}(1 - \lambda_j) \cdot \lambda_{M+1} r_{M+1} G_M
 \\
 &= (1-\alpha) + (1-\lambda_2)(1-\alpha) \alpha + \cdots +  \prod_{j=2}^{M}(1 - \lambda_j) \cdot (1-\alpha) \alpha^{M-1}
 \\
 & \leq(1-\alpha) + (1-\alpha) \alpha + \cdots +   (1-\alpha) \alpha^{M-1}  = (1-\alpha) \frac{1 - \alpha^{M}}{1-\alpha} < 1.
 \end{split}
\end{equation*}
Clearly, such ranking is not optimal and the optimal ranking must contain product 1. We conclude that the optimal expected revenue $\max_{\sigma}\EE[R(\sigma,X)]=1$.

In terms of the clairvoyant upper bound,
    the clairvoyant retailer would set the following ranking for customers with attention span $x$:
    display product one in position $x$ and display other (identical) products in position one to $x-1$.
    In this case, the optimal revenue for customers with attention span $x=1$ is 1.
    Similarly, the optimal revenue for customers with attention span $x=2$ is $2-\alpha-\lambda$, by displaying product two in position one and product one in position two.
    Observe that the revenue is monotonically decreasing with respect to $\lambda$, the purchase probability of product two, so the revenue is maximized when $\lambda \rightarrow 0$.
    Then the expected revenue for customer with $x=2$ is arbitrarily close to $ 2-\alpha$.
    Using the same argument, we can argue that for customer with attention span $x$, the maximal expected revenue is $x-(x-1)\alpha$, which can be attained when $\lambda \rightarrow 0$.
    As a result, noting that the fraction of customers with attention span $x$ is $\alpha^{x-1}(1-\alpha)$, the clairvoyant upper bound is
\begin{align*}
\sum_{x=1}  \big(x-(x-1)\alpha\big) \alpha^{x-1}(1-\alpha) = 1+\alpha,
\end{align*}
which is maximized when $\alpha \rightarrow 1$.
Therefore, the clairvoyant upper bound can be arbitrarily close to 2.
Since the optimal revenue for random attention spans is 1, no algorithm can achieve a performance ratio better than $1/2$ with respect to the clairvoyant upper bound.
\QED
\endproof

\proof{Proof of Proposition~\ref{prop:prefix}:}
We first show the only if part.
Suppose $\sigma^M$ is the optimal ranking under attention span $M$.
Clearly $\sigma^k $ contains exactly the first $k$ items within $\sigma^M$. In particular, we have $\sigma^1 = \sigma^M(1)$.
In this case, to maximize the expected revenue of the single product, we must have $\sigma^M(1) = \max_{j} \lambda_j r_j$.

Then we consider $\sigma^2$, whose first item has been determined as $\sigma^M(1)$.
Conditional on a customer with $x=2$ not buying the first item,
clearly the second item $j$ should maximize the conditional expected revenue $\lambda_j r_j$ among the remaining products.
Moreover, by Lemma~\ref{lem:fix-m-decrease-r}, we must have $\sigma^2(2) = \max_{j > \sigma^1(1)} \lambda_j r_j $.
That implies $i_1<i_2$.
By iteratively applying this process, we conclude that we must have $i_1 < i_2 < \dots < i_M$.

Next we show the if part.
We use mathematical induction in $M$ to show the claim that if $i_1 < i_2 < \dots < i_M$, then $\sigma^k$ is a prefix of $\sigma^{k+1}$ for $1\le k\le M-1$.
For $M=1$, the claim is degenerate and always true.
Suppose the claim holds for $M\le n$.
For $M=n+1$, by the inductive hypothesis, because $i_1<\dots<i_{n}$, we have $\sigma^{k}$ is a prefix of $\sigma^{k+1}$ for $k=1,\dots,n-1$
and $\sigma^k=\{i_1,\dots,i_k\}$ for $k=1,\dots,n$.
To show $\sigma^n$ is a prefix of $\sigma^{n+1}$, consider the first product in $\sigma^{n+1}$.
If $\sigma^{n+1}(1)\coloneqq j< i_1$, then for the optimal ranking from position $2$ to $n+1$, we can consider a customer with attention span $n$.
Since $i_1<\dots<i_{n}$, by the inductive hypothesis, this subranking is exactly $\sigma^n$.
However, note that $\lambda_j r_j<\lambda_{i_{n+1}} r_{i_{n+1}}$ by the definition of $i_{n+1}$.
Moreover, $\lambda_{j} <\lambda_{i_{n+1}}$ because $r_{j}>r_{i_{n+1}}$ since $ j<i_1<i_{n+1}$ and by condition~\eqref{eq:product-indexing}.
It increases the revenue by replacing $j$ by $i_{n+1}$ due to the last but one equation of \eqref{eq:rec3}.
This implies that $\sigma^{n+1}(1)\ge i_1$.
By Proposition~\ref{prop:nested}, $i_1\in \sigma^n\subset \sigma^{n+1}$.
By Lemma~\ref{lem:fix-m-decrease-r} and the fact that $\sigma^{n+1}(1)\ge i_1$, we must have $\sigma^{n+1}(1)=i_1$.
Now we consider the subranking from position 2 to $n+1$.
Consider a customer with attention span $n$, because $i_2<\dots<i_{n+1}$, by the inductive hypothesis, we can infer that
$\sigma^{n+1}(k)=i_k$ for $k=2,\dots,n+1$.
This completes the inductive step and the claim.
\QED
\endproof

\proof{Proof of Proposition~\ref{prop:geo_discount}:} Let $G_k = \alpha^{k-1}$ for some $\alpha \in [0,1)$ and let $\sigma$ be the optimal ranking consisting of $M$ products.
Assume there is a position $i < M$ such that $    \frac{r_{\sigma(i)} \lambda_{\sigma(i)}}{1-\alpha( 1- \lambda_{\sigma(i)} ) } <     \frac{r_{\sigma(i+1)} \lambda_{\sigma(i+1)}}{1- \alpha( 1- \lambda_{\sigma(i+1)} ) } $. Let $a = \sigma(i)$ and $a' = \sigma(i+1)$ be two products in positions $i$ and $i+1$, respectively. We consider an alternative ranking that swaps $a$ and $a'$ and leaves the positions of all other products unchanged. It is easy to see that the purchasing probabilities for all positions $j \notin \{i,i+1\}$ remain the same.

Let $C_i(\sigma) \triangleq  \prod_{s=1}^{i-1} \big(1-\lambda_{\sigma(s)}\big) $.
By swapping the positions of $a$ and $a'$ and using the facts that $G_i = \alpha^{i-1}$ and $G_{i+1} = \alpha^i$,
the total change in the expected reward is
\begin{align*}
    &  C_i(\sigma) \Big ( r_{a'} \lambda_{a'}  G_i + r_{a} (1-\lambda_{a'}) \lambda_{a} G_{i+1} -  r_{a} \lambda_{a}  G_i - r_{a'} (1-\lambda_{a}) \lambda_{a'} G_{i+1}
    \Big )
     \\
& =  C_i(\sigma)\Big(  r_{a'} \lambda_{a'}  ( G_i  - (1-\lambda_{a}) G_{i+1} )
- r_{a} \lambda_{a}  ( G_i - (1-\lambda_{a'}) G_{i+1} ) \Big ) 
\\
& =  \alpha^{i-1}C_i(\sigma)\Big(  r_{a'} \lambda_{a'}  ( 1 -  \alpha (1-\lambda_{a})  )
- r_{a} \lambda_{a}  ( 1-  \alpha (1-\lambda_{a'}) ) \Big ) 
\geq 0 ,
\end{align*}
where the last inequality uses the assumption that $r_{a'} \lambda_{a'}( 1 - \alpha (1- \lambda_{a}))\geq r_{a} \lambda_{a}( 1 - \alpha ( 1- \lambda_{a'}))$. Hence, swapping the position for $\sigma(i)$ and $ \sigma(i+1)$ would increase the total revenue, which contradicts the optimality of this solution. This completes the proof. 

\QED 
\endproof

\section{Proofs in Section~\ref{sec:learning}} \label{app:learning}
\textit{Proof of Lemma~\ref{lemma:favorable_event}:} This result follows standard techniques in linear cascade bandits, we present the details here for self-completeness.
First of all, we consider the event $\{| \bx^\top(\btheta^* - \hat \btheta_t)|  \leq \rho_t \|\bx\|_{\V_t^{-1}},\,\,\forall \|\bx\|\leq 1 \}$.
Let $ \cI_t^o$ be the set of viewed products in round $t$. In particular, we have $\cI_t^o = \{k | O_{t,k}^Z = 1\}$. For any $t=1,2,\cdots$ and $k=1,2,\cdots, |\cI_t^o|$, define
$$
\eta_{t,k} = Z_{t,k} - \bx_{t,\sigma_t(k)}'\btheta^*.
$$
As $\eta_{t,j} \in [- \bx_{t,\sigma_t(k)}'\btheta^*, 1- \bx_{t,\sigma_t(k)}'\btheta^*]$,
we can see that $\eta_{t,j}$'s are bounded and $\frac{1}{2}$-subgaussian conditional on $\mathcal F_{t-1}$ and $\by_t$ with $\EE[\eta_{t,k}|\mathcal F_{t-1},\by_t] = 0$.
Recall that $\V_t = \sum_{s=1}^t \sum_{k=1}^{|\cI_s^o|} \bx_{s,\sigma_t(k)}\bx_{s,\sigma_t(k)}^\top + \gamma \mathbf{I}$ and $\| \btheta^*\|_2 \leq D$, by Theorem 2 in \cite{abbasi2011improved}, for any $\delta \in (0,1)$, we have
$$
\|\hat \btheta_t -\btheta^*\|_{\V_t} \leq \sqrt{\frac{1}{2}\log \Big(\frac{\det(\V_t)^{1/2} \det(\V_0)^{-1/2})}{\delta}\Big )} + \gamma^{1/2}D, \,\,\,\forall t=0,1,\cdots
$$
with probability at least $1-\delta$. Using trace-determinant inequality (Lemma 10 in \cite{abbasi2011improved}) and the facts that $ \|\bx_{s,\sigma_s(k)}\|^2 \leq 1 $ and $|\cI_t^o| \leq M$, we have
$$
\det(\V_t) \leq \Big ( \frac{\trace(\V_t)}{d} \Big )^{d}= \Big ( \gamma + \frac{1}{d} \sum_{s=1}^t \sum_{k=1}^{|\cI_t^o|} \|\bx_{s,\sigma_s(k)}\|^2 \Big )^d \leq \Big ( \gamma+ tM/d \Big )^d,
$$
which further implies
$$
\frac{\det(\V_t)}{\det(\V_0)} \leq \big (\frac{\gamma + tM/d}{\gamma} \big)^d = \big (1+ \frac{tM}{\gamma d} \big )^d.
$$
Finally, letting $\delta = 1/t^2$,  we obtain that with probability at least $1-1/t^2$,
\begin{equation}\label{eq:x-bound}
   \|\hat \btheta_t -\btheta^*\|_{\V_t} \leq \frac{1}{2}  \sqrt{d\log \big( 1 + \frac{tM}{\gamma d} \big )  + 4\log(t) }     + D\gamma^{1/2} = \rho_t.
\end{equation}
Next, we consider the second part of the favorable event $\xi_t$. For any span $k$, as $Y_{t,k}$'s are a sequence of Bernoulli random variables with mean $h_{k}^*$, recall Hoeffding's inequality,
for any $\epsilon>0$, we have
$$
\PP ( |\hat h_{t,k} - h_k^*|\leq \epsilon  ) \geq 1-\exp(-2\epsilon^2N_{t,k}).
$$
Choosing $\epsilon = \sqrt{\ln(t)/N_{t,k}}$, we obtain
$$
\PP \Big( |\hat h_{t,k} - h_k^*|\leq \sqrt{\frac{\ln(t)}{N_{t,k}}} \Big) \geq 1-\exp(-2\ln(t)) = 1-1/t^2.
$$
Applying the union bound for $k=1,2,\cdots,M-1$, together with \eqref{eq:x-bound},
we have shown that $\PP(\xi_t) \geq 1-M/t^2$ and thus completes the proof.
\QED

Next we outline the main idea behind the proof of Theorem~\ref{thm:regret}.
Recall the definition of the favorable event $\xi_t$ in Lemma~\ref{lemma:favorable_event}.
As a standard technique in UCB-type algorithms,
we can decompose the regret in period $t$ by
$$
\text{Reg}_t = \EE[  \text{Reg}_t\I_{\xi_{t-1}}]+  \EE[ \text{Reg}_t \I_{\xi^c_{t-1}}]\leq \EE[ \text{Reg}_t  \I_{\xi_{t-1}}]+   \frac{ r_{\max}M}{t^2}.
$$
Here the regret from the unfavorable event $\xi_{t-1}^c$ can be bounded by the maximum revenue in a single period $r_{\max}$ and the probability $\PP(\xi_{t-1}^c)\le M/t^2$ by Lemma~\ref{lemma:favorable_event}.

To bound the regret from the favorable event, the standard approach is to show that when the parameters are optimistically estimated,
the corresponding reward (revenue in our case) provides an upper bound for the actual reward as long as the true parameters fall into the confidence region.
However, in our case, the optimistic estimators $G_{t,k}^U$ may not have IFR (see Assumption \ref{assump:IFR}) anymore.
It is not clear if the approximation algorithm in Step~\ref{step:approx-alg-ucb} of Algorithm~\ref{alg:UCB} would still yield at least $1/e$ of the optimal revenue.
Our next result addresses this concern.
\begin{lemma}\label{lemma:1/e-UCB}
    We have
    \begin{equation*}
        \I_{\xi_{t-1}}\EE\left[\cR(\sigma_t;u_t,  G_t^U)-\frac{1}{e}\cR(\sigma^*; \lambda_t,  G^{\ast})\bigg|\mathcal F_{t-1}, \by_t\right]\geq 0.
    \end{equation*}
\end{lemma}
The term $\cR(\sigma_t;u_t, G_t^U)$ stands for the (conditionally) expected revenue according to the optimistic estimators and $\cR(\sigma^* ; \lambda_t, G^{\ast})$ is the optimal expected revenue for customer $t$.
The latter is independent of $\mathcal F_{t-1}$ and depends on $\by_t$ through $\lambda$.

\begin{proof}{Proof of Lemma \ref{lemma:1/e-UCB}: }
    Suppose $u \geq \lambda$, we consider $G$ to be a decision variable and
    search for the optimal ranking $\tilde \sigma$ and $\tilde G$
    that satisfies Assumption~\ref{assump:IFR} and $\tilde{G} \leq G^U_t$ under choice probability $u$.
    That is,
\begin{equation}\label{prob:est}
\begin{split}
(\tilde \sigma,\tilde G) = & \argmax_{\sigma,G} \sum_{k=1}^{M} \prod_{s=1}^{k-1} (1-u_{\sigma(s)}) \cdot u_{\sigma(k)} r_{\sigma(k)} \cdot G_k\\
\mbox{s.t.}\, & G_k \leq  G_k^U\,\,\text{ for }k=1,2,\cdots, M,\\
& G_1,\cdots,G_n \text{ follow IFR}.
\end{split}
\end{equation}
Let  $\{ \sigma^x \}$'s be the solution return by AssortOpt with reward $r_j$'s and choice probability $u_x$'s, which is independent of the attention span distribution.
As $\tilde G$ follows IFR, we have
$$
\max_x R(\sigma^x,x; u) \tilde G_x \geq \frac{1}{e} \max_{\sigma} \sum_{x=1}^{M} \prod_{s=1}^{k-1} (1-u_s) \cdot u_x \cdot r_x \cdot \tilde G_x = \frac{1}{e}\cdot \cR(\tilde\sigma; u,   \tilde G).
$$
On the other hand, the actual distribution $G$ and the corresponding optimal ranking $\sigma^*$ are also feasible for \eqref{prob:est}. As $u \geq \lambda$,
it is thus easy to see that
$$
 \cR(\sigma^*; \lambda,   G )\leq   \cR(\sigma^*; u,   G ) \leq  \cR(\tilde \sigma; u,   \tilde G) .
$$
Combining them together, we have
\begin{align*}
     & \cR(\sigma^*;  \lambda,  G^{\ast})\leq  \cR(\tilde \sigma;  u,  \tilde{G}) \leq e \cdot \max_x R(\sigma^x,x; u) \cdot \tilde G_x\\
     \leq & e \cdot \max_x R(\sigma^x,x; u)  \cdot G_x^U   = e \cdot R(\sigma_t,|\sigma_t|; u) G_{|\sigma_t|}^U  \leq e \cdot  \cR(\sigma_t; u, G^U),
\end{align*}
where the third inequality is due to $\tilde G_k \leq G_k^U$ in \eqref{prob:est} and the fourth inequality comes from the fact that $\sigma_t = \argmax_{\sigma^x: 1\leq x \leq M} R(\sigma^x,x;u)G_x^U$. This completes the proof.
\end{proof}

Next we introduce a lemma that will be used in the proof for Theorem~\ref{thm:regret}.
\begin{lemma}\label{lemma:regret_error}
Suppose $u \geq \lambda$ and $h^L \leq h^*$. Let $G_k^U = \prod_{i=1}^{k-1}(1-h_i^L)$, we have
\begin{equation*}
\begin{split}
& \sum_{k=1}^{M}\Big ( \prod_{i=1}^{k-1} (1-u_i) \cdot u_k \cdot G_{k}^U   - \prod_{i=1}^{k-1} (1-\lambda_i) \cdot \lambda_k \cdot G_{k}^*\Big ) r_k  \\
\leq & \sum_{k=1}^M \EE[O_{t,k}^Y]\cdot |h_k^* - h_k^L| \cdot r_{\max} + \sum_{k=1}^M \EE[O_{t,k}^Z]\cdot |u_k - \lambda_k| \cdot r_{\max}.
\end{split}
\end{equation*}
\end{lemma}
\textit{Proof of Lemma \ref{lemma:regret_error}:}
Given a ranking $\sigma_t$, conditional purchase probability $\lambda$, and failure rate $h$, for any $k \geq 1$, we have
\begin{equation}\label{eq:yz_relationship}
   \PP(O_{t,k+1}^Z = 1; \lambda,h) = (1-h_k) \PP[O_{t,k}^Y = 1; \lambda,h]\text{ and }\PP[O_{t,k}^Y = 1; \lambda,h] = (1-\lambda_k) \PP[O_{t,k}^Z = 1; \lambda,h].
\end{equation}
We can see that
\begin{equation*}
\begin{split}
    & \prod_{i=1}^{k-1} (1-u_i) \cdot u_k \cdot G_{k}^U   - \prod_{i=1}^{k-1} (1-\lambda_i) \cdot \lambda_k \cdot G_{k}^* \\
    = & \prod_{i=1}^{k-1} (1-u_i)(1-h_i^L) \cdot u_k   - \prod_{i=1}^{k-1} (1-\lambda_i)(1-h_i^*) \cdot \lambda_k \\
   = & \PP[O_{t,k}^Y = 1; u,h^L]\cdot u_{k} - \PP[O_{t,k}^Z = 1; \lambda,h^*]\cdot \lambda_{k}.
   \end{split}
\end{equation*}
To bound the right-hand side, by \eqref{eq:yz_relationship}, we have
\begin{equation}\label{eq:yz_01}
\begin{split}
    &\PP[O_{t,k}^Z = 1; u,h^L]\cdot u_{k} - \PP[O_{t,k}^Z = 1; \lambda,h^*]\cdot \lambda_{k} \\
    =  & \big ( \PP[O_{t,k}^Z = 1; u,h^L] - \PP[O_{t,k}^Z = 1; \lambda,h^*] \big )\cdot u_{k} + \PP[O_{t,k}^Z = 1; \lambda,h^*] \cdot(u_k- \lambda_{k} ) \\
   = &  \big ( \PP[O_{t,k}^Z = 1; u,h^L] - \PP[O_{t,k}^Z = 1; \lambda,h^*] \big )\cdot u_{k}   +
    \EE[O_{t,k}^Z] \cdot (u_k- \lambda_{k}).
\end{split}
\end{equation}
Then consider the term $\PP[O_{t,k}^Z = 1; u,h^L] - \PP[O_{t,k}^Z = 1; \lambda,h^*] $. we have
\begin{equation}\label{eq:yz_02}
    \begin{split}
       & \PP[O_{t,k}^Z = 1; u,h^L] - \PP[O_{t,k}^Z = 1; \lambda,h^*] \\
       = & (1-h_{k-1}^L) \cdot \PP[O_{t,k-1}^Y = 1; u,h^L]  + (1-h_{k-1}) \cdot \PP[O_{t,k-1}^Y = 1; \lambda,h^*] \\
        = &
        (1-h_{k-1}^L) \cdot \big ( \PP[O_{t,k-1}^Y = 1; u,h^L]  - \PP[O_{t,k-1}^Y = 1; \lambda,h^*]  \big )
        + (h_{k-1}^*- h_{k-1}^L) \cdot \PP[O_{t,k-1}^Y = 1; \lambda,h^*] )  \\
        = &  (1-h_{k-1}^L) \cdot \big ( \PP[O_{t,k-1}^Y = 1; u,h^L]  - \PP[O_{t,k-1}^Y = 1; \lambda,h^*]  \big )
        + \EE[ O_{t,k-1}^Y]\cdot (h_{k-1}^* - h_{k-1}^L).
    \end{split}
\end{equation}
Furthermore, we have
\begin{equation*}
\begin{split}
    &  \PP[O_{t,k-1}^Y
    = 1; u,h^L]  - \PP[O_{t,k-1}^Y = 1; \lambda,h^*]  \\
    = &
    (1-u_{k-1})   \PP[O_{t,k-1}^Z = 1; u,h^L]   - (1-\lambda_{k-1})\PP[O_{t,k-1}^Z = 1; \lambda,h^*]  \\
     = & (1-u_{k-1})   \Big (\PP[O_{t,k-1}^Z = 1; u,h^L]    -\PP[O_{t,k-1}^Z = 1; \lambda,h^*]  \Big ) +  (\lambda_{k-1} - u_{k-1})\PP[O_{t,k-1}^Z = 1; \lambda,h^*]  \\
     \leq  & (1-u_{k-1})   \Big (\PP[O_{t,k-1}^Z = 1; u,h^L]   -\PP[O_{t,k-1}^Z = 1; \lambda,h^*] \Big )  ,
\end{split}
\end{equation*}
where the last inequality comes from the fact that $\lambda \leq u$.
Combining the preceding two inequalities together, we can see that
\begin{equation}
    \begin{split}
    &  \PP[O_{t,k}^Z = 1; u,h^L] - \PP[O_{t,k}^Z = 1; \lambda,h^*] \\
      \leq & (1-h_{k-1}^L) (1-u_{k-1})   \Big (\PP[O_{t,k-1}^Z = 1; u,h^L]   -\PP[O_{t,k-1}^Z = 1; \lambda,h^*] \Big ) .
    \end{split}
\end{equation}
Recursively applying this process, we obtain
\begin{equation}\label{eq:yz_03}
    \begin{split}
          \PP[O_{t,k}^Z = 1; u,h^L] - \PP[O_{t,k}^Z = 1; \lambda,h^*]
        \leq  \EE \Big [ \sum_{j=1}^{k-1}  O_{t,j}^Y  \cdot |h_j^* - h_j^L|  \cdot \big ( \prod_{s=j+1}^{k-1} (1-u_s)(1-h_s^L)  \big ) \Big ].
    \end{split}
\end{equation}
Plugging \eqref{eq:yz_03} into \eqref{eq:yz_01}, we have
\begin{equation}
    \begin{split}
        & \prod_{i=1}^{k-1} (1-u_i) \cdot u_k \cdot G_{t,k}^U   - \prod_{i=1}^{k-1} (1-\lambda_j) \cdot \lambda_k \cdot G_{k}^*  \\
        \leq & \EE \Big [ \sum_{i=1}^{k-1}  O_{t,i}^Y  \cdot |h_i^* - h_i^L|  \cdot \big ( \prod_{s=i+1}^{k-1} (1-u_s)(1-h_s^L) \cdot u_k \big ) + O_{t,k}^Z  \cdot |u_k - \lambda_k| \Big ].
    \end{split}
\end{equation}
Multiplying the corresponding profit $r_k$ and summing over $k=1,2,\cdots, M$, we can see that
\begin{equation*}
    \begin{split}
       &  \sum_{k=1}^{M}\Big ( \prod_{i=1}^{k-1} (1-u_i) \cdot u_k \cdot G_{k}^U   - \prod_{j=1}^{k-1} (1-\lambda_i) \cdot \lambda_k \cdot G_{k}^*\Big ) r_k \\
        \leq & \sum_{k=1}^{M} \EE \Big [ \sum_{i=1}^{k-1}  O_{t,i}^Y  \cdot |h_i^* - h_i^L|  \cdot \big ( \prod_{s=i+1}^{k-1} (1-u_s)(1-h_s^L) \cdot u_k \big ) + O_{t,k}^Z  \cdot |u_k - \lambda_k| \Big ]\cdot r_k \\
        = & \sum_{k=1}^{M-1} \EE[O_{t,k}^Y]\cdot |h_k^* - h_k^L|  \cdot \underbrace{ \Big [ \sum_{i=k+1}^M   \Big ( \prod_{s=k+1}^{i-1} (1-u_s)(1-h_s^L) \cdot u_i \cdot r_i \Big ) \Big ] }_{Q_k} + \sum_{k=1}^M \EE[O_{t,k}^Z]\cdot |u_k - \lambda_k| \cdot r_k.
    \end{split}
\end{equation*}
Also note that $O_{t,1}^Y = 1$ always holds, which implies that the agent is always able to observe customer's choice decision on the first item.
Finally, consider the term $Q_k$, we can see
\begin{equation*}
    Q_k \triangleq \sum_{i=k+1}^M   \Big ( \prod_{s=k+1}^{i-1} (1-u_s)(1-h_s^L) \cdot u_i \cdot r_i  \Big ) = \EE[\cR(\sigma_t; u,   G^U  )| O_{k+1}^Z = 1] ,
\end{equation*}
which is the conditional expected reward given that user views the $(k+1)$-th item under $u_t, G^u$. Thus, a natural bound obtained for $Q_k$ is
$$
Q_k \leq r_{\max}.
$$
Plug this into the preceding relationship, we have
\begin{equation*}
\begin{split}
& \sum_{k=1}^{M}\Big ( \prod_{i=1}^{k-1} (1-u_i) \cdot u_k \cdot G_{k}^U   - \prod_{i=1}^{k-1} (1-\lambda_i) \cdot \lambda_k \cdot G_{k}^*\Big ) r_k  \\
\leq & \sum_{k=1}^{M-1} \EE[O_{t,k}^Y]\cdot |h_k^* - h_k^L| \cdot r_{\max} + \sum_{k=1}^M \EE[O_{t,k}^Z]\cdot |u_k - \lambda_k| \cdot r_k,
\end{split}
\end{equation*}
which completes the proof.
\QED 

\textit{Proof of Theorem \ref{thm:regret}:}
Let $\cF_{t}$ be the $\sigma$-algebra of historical observations up to the end of round~$t$, $\by_t$ be the feature of user $t$, and $\sigma_t$ be the ranking displayed in round $t$.
The favorable event is defined in Lemma \ref{lemma:favorable_event}
$$\xi_t = \Big \{  \{\| \btheta^* - \hat \btheta_t \|_{\V_t} \leq \rho_t\} \cap \{  h_k^* \geq \hat h_{t,k} - \sqrt{\ln(t)/N_{t,k}} \}, \,\,\forall k=1,2,\cdots, M-1 \Big \}.$$
By Lemma \ref{lemma:favorable_event}, we have $\PP(\xi_t) \geq 1- M/t^2$.
Under $\xi_t$, we have $\lambda_{t,j} \leq u_{t,j}$ for each product $j$ and $h_{t,k}^L \leq h_k^*$ for all $k=1,2,\cdots, M$, which further implies $G_{t,k}^U \geq G_k^*$ for all $k=1,\cdots, M$.

Recall that $\text{Reg}_t$ is the scaled regret incurred in round $t$, as $\text{Reg}_t  \leq r_{\max}$, given the customer feature $\by_t$, we have
\begin{equation*}
\begin{split}
& \EE[\text{Reg}_t | \cF_{t-1},\by_t] = \EE[ \cR(\sigma^*; \lambda_t,G^{\ast}) |\by_t]- e \cdot  \EE[ \cR(\sigma_t; \lambda_t,G^{\ast}) | \cF_{t-1},\by_t ]\\
= &\EE[ \big ( \cR(\sigma^*; \lambda_t,G^{\ast}) - e \cdot  \EE[ \cR(\sigma_t; \lambda_t,G^{\ast}) \big )\cdot \I_{\xi_{t-1}} | \cF_{t-1},\by_t ]+ r_{\max} \cdot \PP( \xi_{t-1}^c)
\\
 \leq & \EE[ \big ( \cR(\sigma^*; \lambda_t,G^{\ast}) - e \cdot  \EE[ \cR(\sigma_t; \lambda_t,G^{\ast}) \big )\cdot \I_{\xi_{t-1}} | \cF_{t-1},\by_t ]  + \frac{Mr_{\max}}{t^2} \\
= & \EE[ \big ( \cR(\sigma^*; \lambda_t,G^{\ast}) - e\cdot \cR(\sigma_t; u_t,G_t^{U})    \big )\cdot \I_{\xi_{t-1}} | \cF_{t-1},\by_t ]  + \frac{Mr_{\max}}{t^2} \\
& + \EE[  e \cdot \cR(\sigma_t; u_t,G_t^{U})  - e \cdot  \EE[ \cR(\sigma_t; \lambda_t,G^{\ast}) \big )\cdot \I_{\xi_{t-1}} | \cF_{t-1},\by_t ].
\end{split}
\end{equation*}
Since $u_{t,j}\geq \lambda_{t,j}$ and $h_{t,k}^L \leq h_k^*$ under $\xi_{t-1}$, using Lemma \ref{lemma:1/e-UCB}, we can see that
\begin{align*}
\EE[  e\cdot \cR(\sigma_t; u_t,G_t^{U})    \cdot \I_{\xi_{t-1}} | \cF_{t-1},\by_t ]  \geq &  \EE[   e\cdot \max_{x}R(\sigma_t,x; u_t,G_t^{U})    \cdot \I_{\xi_{t-1}} | \cF_{t-1},\by_t ]   \\
\geq &  \EE[   \cR(\sigma_t; \lambda_t,G^*)    \cdot \I_{\xi_{t-1}} | \cF_{t-1},\by_t ] ,
\end{align*}
so that the preceding inequality could be further expressed as
\begin{equation}\label{eq:regret_01}
     \EE[\text{Reg}_t | \cF_{t-1},\by_t] \leq  \EE[  e \cdot \cR(\sigma_t; u_t,G_t^{U})  - e \cdot  \EE[ \cR(\sigma_t; \lambda_t,G^{\ast}) \big )\cdot \I_{\xi_{t-1}} | \cF_{t-1},\by_t ] + \frac{Mr_{\max}}{t^2}  .
\end{equation}
Moreover, with the help of Lemma \ref{lemma:regret_error}, we have
\begin{equation*}
\begin{split}
& \EE \Big[      \big ( \cR(\sigma_t; u_t,G_t^U)   -   \cR(\sigma_t; \lambda_t,G^{\ast})  \big ) \cdot \I_{\xi_{t-1}}  \Big | \cF_{t-1},\by_t \Big ]  \\
= & \sum_{x=1}^M  \EE \Big [ \Big ( \prod_{s=1}^{x-1} (1-u_{t,\sigma(s)}) \cdot u_{t,\sigma_t(x)} G_{t,x}^U- \prod_{s=1}^{x-1} (1-\lambda_{t,\sigma_t(s)}) \cdot \lambda_{t,\sigma_t(x)} \cdot G_x \Big ) \cdot \I_{\xi_{t-1}}  \Big | \cF_{t-1},\by_t \Big]\cdot  r_{\sigma_t(x)}\\
\leq &  r_{\max} \cdot \EE \Big [ \Big (\sum_{k=1}^M O_{t,k}^Y\cdot |h_k^* - h_k^L|  + \sum_{k=1}^M O_{t,k}^Z\cdot |u_k - \lambda_k| \Big ) \cdot \I_{\xi_{t-1}} \Big | \cF_{t-1},\by_t \Big ]
\\
\leq & 2 r_{\max} \cdot \EE \Big [ \Big ( \underbrace{\sum_{k=1}^M  O_{t,k}^Y \cdot  \sqrt{\frac{\log(t-1)}{N_{t-1,k}}} }_{L_{t,1}} + \underbrace{\sum_{k=1}^M  O_{t,k}^Z\cdot  \rho_{t-1} \|\bx_{t,\sigma_t(k)}\|_{\V_{t-1}^{-1}} }_{L_{t,2}} \Big ) \cdot \I_{\xi_{t-1}} \Big | \cF_{t-1},\by_t \Big ]
,
 \end{split}
 \end{equation*}
where the last inequality uses the fact that $|h_k^* - h_{t,k}^L| \leq 2 |h_k^* - \hat h_{t,k}|\leq 2 \sqrt{\frac{\ln(t-1)}{N_{t-1,k}}} $ and $|u_{t,\sigma_t(k)} - \lambda_{t,\sigma_t(k)}| \leq 2 |\bx_{t,\sigma_t(k)}^\top(\btheta^* - \hat  \btheta_{t-1})| \leq 2\rho_{t-1} \|\bx_{t,\sigma_t(k)}\|_{\V_{t-1}^{-1}}$ under the favorable event $\xi_{t-1}$.
Let
\begin{align*}
    L_{t,1} =  \sum_{k=1}^M  O_{t,k}^Y \cdot  \sqrt{\frac{\ln(t-1)}{N_{t-1,k}}} , \,\, L_{t,2} = \sum_{k=1}^M  O_{t,k}^Z\cdot  \rho_{t-1} \|\bx_{t,\sigma_t(k)}\|_{\V_{t-1}^{-1}} .
\end{align*}
Plugging them into \eqref{eq:regret_01} and summing the scaled-regret over $t=1,2,\cdots, T$, we have
\begin{equation*}
    \begin{split}
         \sum_{t=1}^T \EE[\text{Reg}_t ]
        \leq &  2e\cdot  r_{\max} \EE \Big[ \sum_{t=1}^T (L_{t,1}+L_{t,2}   )\I_{\xi_{t-1}} \Big] +  \sum_{t=1}^T \frac{2r_{\max}}{t^2}  \\
        \leq & 2e\cdot  r_{\max} \EE \Big[\sum_{t=1}^T (L_{t,1}+ L_{t,2}   )\I_{\xi_{t-1}} \Big] +  \frac{M\pi^2r_{\max}}{6},
    \end{split}
\end{equation*}
where the last inequality uses the fact that $\sum_{t=1}^\infty 1/t^2 = \pi^2/6 $. The rest of our analysis is to bound both $\sum_{t=1}^T \EE [L_{t,1}\cdot \I_{\xi_{t-1}}]$ and $\sum_{t=1}^T \EE [L_{t,2}\cdot \I_{\xi_{t-1}}]$. Firstly, consider $\sum_{t=1}^T \EE [L_{t,1}\cdot \I_{\xi_{t-1}}]$,
 we have
\begin{equation*}
    \begin{split}
       \sum_{t=1}^T L_{t,1}= & \sum_{t=1}^T \sum_{k=1}^M  O_{t,k}^Y \cdot  \sqrt{\frac{\ln(t-1)}{N_{t-1,k}}} \\
        \leq & \sum_{k=1}^M    \int_{s=1}^{N_{t-1,k} +1}\sqrt{\frac{\ln T}{s}}  ds \\
        \leq & 2M \sqrt{(N_{t-1,k}+1)\ln T} \leq   2M \sqrt{T\ln T}.
    \end{split}
\end{equation*}
Consequently, we have
\begin{equation}\label{eq:L1}
    \sum_{t=1}^T \EE [L_{t,1}\cdot \I_{\xi_{t-1}}] \leq \EE[ \sum_{t=1}^T L_{t,1} ] \leq 2M \sqrt{(N_{t-1,k}+1)\ln T} \leq   2M \sqrt{T\ln T},
\end{equation}
where the first inequality is due to the fact that $L_{t,1} \geq 0$.
Next consider $\sum_{t=1}^T \EE[ L_{t,2}\I_{\xi_{t-1}}]$, using similar analysis as Lemma 2 in \cite{wen2017online}, we have
\begin{equation*}\label{eq:L2}
    \begin{split}
       \sum_{t=1}^T L_{t,2}  = &  \sum_{t=1}^T \sum_{k=1}^M  O_{t,k}^Z \cdot  \rho_{t-1} \|\bx_{t,\sigma_t(k)}\|_{\V_{t-1}^{-1}}
        \leq \rho_{T-1} \sqrt{(\sum_{t=1}^T \sum_{k=1}^M  O_{t,k}^Z ) \cdot  2d M \log(1+ \frac{T M}{\gamma d})}.
    \end{split}
\end{equation*}
As a result, $\sum_{t=1}^T \EE [ L_{t,2}\cdot  \I_{\xi_{t-1}}]$ can be further bounded as
 \begin{equation}\label{eq:L2_02}
     \begin{split}
         \sum_{t=1}^T \EE [ L_{t,2} \cdot \I_{\xi_{t-1}}] \leq & \sum_{t=1}^T \EE [ L_{t,2}] \leq   \rho_{T-1} \cdot \EE \left  [ \sqrt{ (\sum_{t=1}^T \sum_{k=1}^M  O_{t,k}^Z ) \cdot 2d M \log(1+ \frac{T M}{\gamma d})} \right ] \\
         \leq & \rho_{T-1} \cdot   \sqrt{  \sum_{t=1}^T \EE (\sum_{k=1}^M  O_{t,k}^Z) \cdot  2d M \log \Big (1+ \frac{T M}{\gamma d} \Big )} \\
         \leq & \rho_{T-1} \cdot   \sqrt{ 2M^2 T  d \log\Big (1+ \frac{T M}{\gamma d} \Big )} .
     \end{split}
 \end{equation}
 Again, the first inequality is due to $L_{t,2} \geq 0$, and the third inequality uses the concavity of $g(x) = \sqrt{x}$ and Jensen's inequality.
 Combining Eqs. \eqref{eq:L1} and \eqref{eq:L2_02} together, we have
 \begin{equation*}
     \sum_{t=1}^T \EE[\text{Reg}_t ]
        \leq 4eM r_{\max} \sqrt{T\ln T}  +  2 r_{\max}\rho_{T-1} \cdot   \sqrt{ 2M^2 T  d \log\Big (1+ \frac{T M}{\gamma d} \Big )}+  \frac{M\pi^2r_{\max}}{6},
 \end{equation*}
 which completes the proof.
\QED

\section{Multiple Purchases and Product-dependent Leaving Probability}\label{sec:multiple-purchase}
In the base model in Section~\ref{sec:model}, there are two outcomes after a customer inspects a product inside her attention span:
she may purchase the product and leave the platform, or not purchase product and keep viewing.
The actions of purchasing and leaving are coupled through the probability $\lambda_k$ for product $k$.
In this section, we extend our model in Section~\ref{sec:model} by decoupling the two actions:
the customer may purchase a product and continue viewing the remaining products, which potentially leads to multiple purchases,
or leave the platform prematurely (not yet reaching the attention span) after viewing an unfavorable product.

Specifically, after viewing product $k$, the consumer purchases the product with probability $\lambda_k$, similar to the base model.
Given that the attention span has not been reached, the customer keeps viewing the products on the platform with probability $c_k$, which also depends on the quality of the product just viewed.
Note that we do not need to specify the joint distribution of the two events: they can be independent or highly correlated.
For example, in the base model, we have $\lambda_k=1-c_k$ and the events of purchasing and leaving are perfectly correlated.

In this model, given ranking $\sigma$ and a customer with attention span $x$, her probability of purchasing the $k$-th product ($k \leq x$) is
\begin{equation*}
\lambda_{\sigma(k)} \cdot \prod_{i=1}^{k-1} c_{\sigma(i)}.
\end{equation*}
Consequently, the expected revenue from a consumer with attention span $x$ is
\begin{equation}\label{eq:fix-span-revenue-general}
    R(\sigma,x) \triangleq \sum_{k=1}^{x\wedge M} \prod_{i=1}^{k-1} c_{\sigma(i)}  \cdot \lambda_{\sigma(k)} r_{\sigma(k)}.
\end{equation}
Comparing \eqref{eq:fix-span-revenue-general} with \eqref{eq:fix-span-revenue},
we can see that the model in this section is a more general setup.
For a customer with random attention spans, the seller takes the expected value of $x\sim X$ in \eqref{eq:fix-span-revenue-general} and obtain the total expected revenue
\begin{align*}
\EE[R(\sigma, X)]= &
\sum_{x=1}^M  g_x R(\sigma,x) = \sum_{x=1}^M  g_x\Big ( \sum_{k=1}^{x} \prod_{i=1}^{k-1} c_{\sigma(i)}  \cdot \lambda_{\sigma(k)} r_{\sigma(k)} \Big )\\
= & \sum_{x=1}^{M} \prod_{i=1}^{x-1} c_{\sigma(i)}  \cdot \lambda_{\sigma(x)} r_{\sigma(x)}  G_x.
\end{align*}
This is the objective to maximize for the seller.

For the ease of discussion,
we assume that we can index the products in the following order
\begin{equation}\label{eq:order_quit}
\frac{r_1 \lambda_1}{1-c_1} > \frac{r_2 \lambda_2}{1-c_2}  > \cdots > \frac{r_n \lambda_n}{1-c_n}.
\end{equation}
The strict inequality is merely a technical assumption to avoid ties.
The results can be extended using concepts such as $\cL$-optimal rankings in Section~\ref{sec:model}.
Similar to Lemma~\ref{lem:fix-m-decrease-r}, the next lemma states that the optimal product ranking for fixed attention spans should always follow the same order as the product indices.

\begin{lemma}\label{lem:fix-m-decrease-r-general}
    Fix attention span at $x$. Suppose the assortment $S$ is given and $\sigma^x$ maximizes $R(\sigma,x)$ among $\sigma\in P(S)$.
    We have that the products in $\sigma^{x}$ are displayed in the increasing order of their product indices, i.e., $\sigma^{x}(i) < \sigma^{x}(i+1)$ for $i=1,\dots,|S|-1$.
\end{lemma}

\textit{Proof of Lemma~\ref{lem:fix-m-decrease-r-general}:}
For the simplicity of notations, let us assume that $S=\left\{1,\dots,x\right\}$.
Suppose in the ranking $\sigma$ there exists $i\in\left\{1,\dots,x-1\right\}$ such that $\frac{r_{\sigma(i)} \lambda_{\sigma(i)}}{1-c_{\sigma(i)}} < \frac{r_{\sigma(i+1)} \lambda_{\sigma(i+1)} }{1-c_{\sigma(i+1)}}$.
We will argue that we can strictly improve the expected revenue by swapping the two products without changing the positions of other products.
Consider the new ranking $\sigma'$ with
$\sigma'(i)=\sigma(i+1)$, $\sigma'(i+1)=\sigma(i)$, and $\sigma'(k)=\sigma(k)$ for all other $k$.
Let $\pi_{\sigma(i)}(\sigma) \triangleq  \prod_{s=1}^{i-1} c_{\sigma(s)}$ be the probability that the customer views product $\sigma(i)$.
By definition, it is easy to see $\pi_{\sigma(k)}(\sigma)  = \pi_{\sigma'(k)}(\sigma')$ for $k=1,\dots, i, i+2,\dots,x$.
Therefore, the expected revenues generated from products in position $k$ are equal under $\sigma$ and $\sigma'$ for $k=1,\dots,i-1,i+2,\dots,x$, recalling formula \eqref{eq:fix-span-revenue-general}.

In order to show $R(\sigma',x)>R(\sigma,x)$, it suffices to compare the revenues generated from the products in position $i$ and $i+1$ under the two rankings:
\begin{equation*}
    \pi_{\sigma(i)}(\sigma)  \lambda_{\sigma(i)} r_{\sigma(i)} +\pi_{\sigma(i+1)}   (\sigma)  \lambda_{\sigma(i+1)} r_{\sigma(i+1)}<\pi_{\sigma'(i)}(\sigma')  \lambda_{\sigma'(i)} r_{\sigma'(i)} +\pi_{\sigma'(i+1)}   (\sigma')   \lambda_{\sigma'(i+1)} r_{\sigma'(i+1)}.
\end{equation*}
This is indeed the case because
 \begin{align*}
& \pi_{\sigma(i)}(\sigma) \cdot r_{\sigma(i)} \lambda_{\sigma(i)} +\pi_{\sigma(i+1)}   (\sigma)  \cdot r_{\sigma(i+1)}\lambda_{\sigma(i+1)}\\
 =&\pi_{\sigma(i)}(\sigma)  \Big (  r_{\sigma(i)} \cdot \lambda_{\sigma(i)}+ c_{\sigma(i)} \cdot r_{\sigma(i+1)} \cdot \lambda_{\sigma(i+1)} \Big ) \\
 =&\pi_{\sigma(i)}(\sigma) \Big ( r_{\sigma(i+1)}  \cdot \lambda_{\sigma(i+1)}+r_{\sigma(i)}  c_{\sigma(i+1)} \cdot \lambda_{\sigma(i)}
+ (1 - c_{\sigma(i+1)})r_{\sigma(i)} \lambda_{\sigma(i)}
-
(1-c_{\sigma(i)})r_{\sigma(i+1)} \lambda_{\sigma(i+1)}
\Big ) \\
 <& \pi_{\sigma(i)}(\sigma) \Big ( r_{\sigma(i+1)} \cdot \lambda_{\sigma(i+1)}  +  r_{\sigma(i)} c_{\sigma(i+1)} \cdot \lambda_{\sigma(i)}  \Big ) \\
 =&\pi_{\sigma'(i)}(\sigma') \cdot r_{\sigma'(i)} \lambda_{\sigma'(i)} +\pi_{\sigma'(i+1)}   (\sigma')  \cdot r_{\sigma'(i+1)}\lambda_{\sigma'(i+1)},
\end{align*}
where the inequality holds by $\frac{r_{\sigma(i)} \lambda_{\sigma(i)}}{1-c_{\sigma(i)}} < \frac{r_{\sigma(i+1)} \lambda_{\sigma(i+1)} }{1-c_{\sigma(i+1)}}$ and
\begin{equation*}
\begin{split}
& (1 - c_{\sigma(i+1)})r_{\sigma(i)} \lambda_{\sigma(i)}
-
(1-c_{\sigma(i)})r_{\sigma(i+1)} \lambda_{\sigma(i+1)}  \\
& = (1-c_{\sigma(i)}) (1 - c_{\sigma(i+1)}) \Big ( \frac{ r_{\sigma(i)} \lambda_{\sigma(i)} }{1-c_{\sigma(i)}}  - \frac{r_{\sigma(i+1)}.\lambda_{\sigma(i+1)} }{1 - c_{\sigma(i+1)}} \Big ) < 0.
\end{split}
\end{equation*}
Therefore, $\sigma$ cannot be optimal, and we have completed the proof.
\QED

\begin{algorithm}[t]\caption{Assortment Optimization for Multiple Purchase and Product-dependent Leaving Probability Given $X=x$}\label{AST_multiple:m}
\begin{algorithmic}
\REQUIRE{$r_j$, $\lambda_j$ for $j\in [n]$, $x$}
\STATE $H^0_{j} = 0$ for all $j = 1,2,\dots,n$; $H^k_{n+1} =0$ for all $k=0,1,2,\dots,x$; $\sigma^0_j =\emptyset$ for all $j = 1,2,\dots,n$; $\sigma^k_{n+1} =\emptyset$ for all $k=0,1,2,\dots,x$
\FOR{$k = 1, \ldots, x$ }
   \FOR{$j = n, \ldots, 1$ }
   \STATE \begin{equation} 
   H^k_j\gets \max \Big \{ r_{j} \lambda_{j} + c_j H_{j+1}^{k-1}, H_{j+1}^k \Big \}.
   \end{equation}
    \IF {$ \Big( r_{j} \lambda_{j} + c_j H_{j+1}^{k-1}  \Big) \geq H^k_{j+1}$}
   \STATE $\quad \sigma^k_j\gets\sigma^{k-1}_{j+1} \cup \{j\}$
    \ELSE
    \STATE  $\quad \sigma^k_j\gets\sigma^k_{j+1}$
    \ENDIF
   \ENDFOR
\ENDFOR
\RETURN $H^{x}_1$ and $\sigma^x_1$
\end{algorithmic}
\end{algorithm}

Next we extend the nested structure to the general model.
With a slight abuse of notation, we denote by $H_j^x$ the optimal revenue for customers with attention span $x$, when the products are chosen from $\{j,j+1,\dots,n\}$, and we denote by $\sigma_j^x$ the optimal ranking for $H_{j}^x$. 
Note that $j$ and $x$ satisfy $j+x\le n+1$ for the problem to be non-trivial.
Otherwise, we just display all products.
In particular, $H_k^{n-k+1} $ is the revenue when the display is chosen from $\sigma \in  P(\{ k,k+1,\cdots, n\})$.
In the next result, we show that the revenue generated by such a display is no more than $\frac{r_k \lambda_k }{1-c_k}$.
This result serves as a building block in for the nested structure of the optimal rankings under fixed attention spans.
\begin{lemma}\label{lem:reward_bound}
For all $ 1\leq k \leq n$, we have
$$
\frac{r_k \lambda_k }{1-c_k} \geq H_k^{n-k+1}.
$$
\end{lemma}

\textit{Proof of Lemma~\ref{lem:reward_bound}:}
We use backward induction for $k=n,n-1,\cdots 1$ to show the statement. Firstly, consider the case that $k=n$. Since $ 1-c_n \leq 1$, clearly, we have
$\frac{r_n \lambda_n}{1-c_n} \geq r_n \lambda_n  = H_n^1$. This implies that our claim is true for $k=n$. Next, suppose the argument is true for $k  = i \in \{2,3,\cdots,n\}$, that is, $\frac{r_i \lambda_i }{1-c_i} \geq H_i^{n-i+1} $. Then for $k=i-1$, we have
\begin{equation*}
\begin{split}
H_{i-1}^{n-i+2} & \le  r_{i-1} \lambda_{i-1} + c_{i-1} H_i^{n-i+1} \leq  r_{i-1} \lambda_{i-1} + c_{i-1} \frac{r_{i} \lambda_{i} }{1-c_{i}}   \\
& \leq r_{i-1} \lambda_{i-1} + c_{i-1} \frac{r_{i-1} \lambda_{i-1} }{1-c_{i-1}}  =  r_{i-1} \lambda_{i-1}  \big (1 +  \frac{c_{i-1}}{1-c_{i-1}} \big )  = \frac{r_{i-1} \lambda_{i-1}  }{1-c_{i-1}},
\end{split}
\end{equation*}
where the second inequality comes from the relationship \eqref{eq:order_quit}.
This implies that correctness of our claim for $k= i-1$ and completes the induction. Consequently, we conclude that $$
\frac{r_k \lambda_k }{1-c_k} \geq H_k^{n-k+1} \text{ for all }1\leq k \leq n.
$$
\QED

In order to prove the nested structure like Proposition~\ref{prop:nested},
We introduce a dynamical programming Algorithm \ref{AST_multiple:m} to derive the optimal ranking $\sigma_j^x$ under each fixed attention span $X = x$.
By Lemma~\ref{lem:fix-m-decrease-r-general}, the products in $\sigma_j^x$ are ranked in the ascending order.
We first prove the base case for the induction:

\textbf{Claim:} $\sigma_{n-2}^{1}\subset \sigma_{n-2}^2 \subset \sigma_{n-2}^{3}$.

We state the proof for the claim below.
Since the products are chosen from $\{n-2, n-1, n\}$, the ranking $\sigma_{n-2}^{3}$ includes all products and thus $\sigma_{n-2}^2 \subset \sigma_{n-2}^{3}$.
To show $\sigma_{n-2}^{1}\subset \sigma_{n-2}^2$, first consider the case $\sigma_{n-2}^1 = \{n-2\}$.
It implies that $r_{n-2}\lambda_{n-2} \geq \max\{r_{n-1}\lambda_{n-1},r_n\lambda_n\}$ because it is optimal to display product $n-2$ when customers have an attention span fixed at one.
If $n-2\notin \sigma_{n-2}^2$, then we have $\sigma_{n-2}^2=\{n-1, n\}$.
Note that with a slight abuse of notation, we simply use a set to denote the ranking.
This is because by Lemma~\ref{lem:fix-m-decrease-r-general} and~\eqref{eq:order_quit}, the optimal ranking has the same order as the indices of the products.
If we replace product $n$ in the second position of $\sigma_{n-2}^2$ by product $n-2$, the expected revenue conditional on the event that the customer doesn't leave after viewing product $n-1$ in the first position is $r_{n-2}\lambda_{n-2}$, no less than $r_n\lambda_n$ which is the conditional expected revenue of product $n$.
Therefore, $\sigma_{n-2}^2=\{n-1, n\}$ cannot be optimal and we conclude that $n-2\in \sigma_{n-2}^2$.
This implies $\sigma_{n-2}^{1}\subset \sigma_{n-2}^2$.
Similar argument can be applied to the cases $\sigma_{n-2}^1 =\{n-1\}$ or $\{n\}$.
Therefore, we have completed the proof for the claim.

Now that we have proved the base case, the next lemma serves as the induction step.
Consider the optimal ranking/assortment $\sigma_j^x$.
Since there are $n-j+1$ products available among $\{j,j+1,\cdots, n\}$, we only focus on $1\leq x \leq n-j+1$.
The following lemma states that if the optimal displays among $\{j,j+1,\cdots, n\}$ preserve a nested structure, then the optimal displays among $\{j-1,j,j+1,\cdots, n\}$ must follow such a nested structure as well.

\begin{lemma}\label{lemma:nested-general}
Given $j \in \{2,3,\cdots, n-2\}$, for any optimal assortment $\sigma_j^x$ of choosing $x$ products from $\{j,j+1,\cdots,n\}$ with $1\leq x\leq n-j+1$,
if they are nested such that $\sigma_j^1\subset \sigma_j^2 \subset \cdots \sigma_j^{n-j+1}$
, then the optimal displays among products $\{j-1,j,\cdots, n\}$ are also nested, i.e.,
$\sigma_{j-1}^1 \subset \sigma_{j-1}^2 \subset \cdots \subset \sigma_{j-1}^{n-j+2}$.
\end{lemma}

Once Lemma~\ref{lemma:nested-general} is proved,
we can apply the mathematical induction to show the nested structure, i.e., the generalization of Proposition~\ref{prop:nested}.
In order to prove Lemma~\ref{lemma:nested-general}
we first provide a lemma characterizing the relationship between $H_j^{x}$ and $H_j^{x+1}$, which is used in our subsequent analysis.
\begin{lemma}\label{lemma:rev_increase-general}
For any $j=1,2,\cdots,n-1$ and $x=1,2,\cdots,n-j$, we have $H_j^x \leq H_j^{x+1}$.
\end{lemma}
\textit{Proof of Lemma \ref{lemma:rev_increase-general}:}
    Recall that $\sigma_j^x$ and $\sigma_j^{x+1}$ are the optimal ranking for attention spans $x$ and $x+1$ respectively.
    Equivalently, $\sigma_j^x$ (respectively $\sigma_j^{x+1}$) includes $x$ (respectively $x+1$) products in the rankings.
    Without loss of generality, for any product $k \geq j, k \notin \sigma_j^{x}$, by adding this product at the last position of $\sigma_j^x$, we can construct a ranking $\tilde \sigma_j^{x+1}$ such that $\tilde \sigma_j^{x+1}(s) = \sigma_j^x(s)$ for $s=1,2,\cdots,m$ and $\tilde \sigma_j^{x+1}(x+1) = k$.
Let $P_0 = \prod_{s = 1}^x  c_{\sigma_j^x(s)}$ be the probability that the consumer would view the last product within $\tilde \sigma_j^{x+1}$.
It is easy to see that
$$
R(\tilde \sigma_j^{x+1},x+1) = R(\sigma_j^x,x) + P_0 \cdot \lambda_k r_k \geq R(\sigma_j^x,x).
$$
Due to the fact $|\tilde \sigma_j^{x+1}| = x+1$ and the optimality of $\sigma_j^{x+1}$, we also have $H_j^{x+1} = R(\sigma_j^{x+1} ,x+1) \geq R(\tilde \sigma_j^{x+1},x+1)$. Combining them together, we conclude that
$$
H_j^{x+1} \geq R(\tilde \sigma_j^{x+1},x+1) \geq R(\sigma_j^x,m) = H_j^x,
$$
which completes the proof.
\QED

Next we present the proof of Lemma~\ref{lemma:nested-general}.
\\
\textit{Proof of Lemma~\ref{lemma:nested-general}:}
First, consider $\sigma_{j-1}^1$ which has a single product.
Suppose $\sigma_{j-1}^1 = \{u\}$ where $u\ge  j-1$. We have $r_{u} \lambda_{u} \geq \max_{k \geq j-1} \{r_k \lambda_k\}$ and
$r_{u} \lambda_{u} > \max_{k=j-1,\dots,u-1} \{r_k \lambda_k\}$
because of the optimality of $\sigma_{j-1}^1$.
Therefore, if $u \notin \sigma_{j-1}^2$, then we can replace the last product in $\sigma_{j-1}^2$ by $u$ to increase the expected revenue of $H_{j-1}^2$ or make it lexicographically less.
As a result, we must have $u \in \sigma_{j-1}^2$ and $\sigma_{j-1}^1 \subset \sigma_{j-1}^2$.

Next, we prove $\sigma_{j-1}^x \subset \sigma_{j-1}^{x+1}$ for $2 \leq x \leq n-j$.
We first consider the case $j-1\notin \sigma_{j-1}^x$.
In this case, we have $\sigma_{j-1}^x = \sigma_j^x$ by their definitions.
If $j-1\notin\sigma_{j-1}^{x+1}$, then we have $\sigma_{j-1}^{x+1} = \sigma_j^{x+1}$ and thus $\sigma_{j-1}^x \subset \sigma_{j-1}^{x+1}$ by the induction hypothesis.
If $j-1\in\sigma_{j-1}^{x+1}$, then by Lemma~\ref{lem:fix-m-decrease-r-general}, we must have that product $j-1$ is displayed in the first position of $\sigma_{j-1}^{x+1}$.
The sub-ranking from the second position onward in $\sigma_{j-1}^{x+1}$ must be equivalent to $\sigma_{j-1}^x$ to maximize the expected revenue.
Therefore, we have $\sigma_{j-1}^x \subset \sigma_{j-1}^{x+1}$.

Now suppose $j-1\in \sigma_{j-1}^x$.
Since $\sigma_{j-1}^x$ is optimal, $j-1$ must be ranked the first and thus
\begin{equation*}
H_j^x\le H_{j-1}^x = \lambda_{j-1} r_{j-1} + c_{j-1} H_j^{x-1} .
\end{equation*}
If ${j-1} \in \sigma_{j-1}^{x+1}$, then $\sigma_{j-1}^x \subset \sigma_{j-1}^{x+1}$ can be proved by the induction hypothesis on the sub-rankings of $\sigma_{j-1}^x$ and $\sigma_{j-1}^{x+1}$ from the second position onward.
Suppose $j-1 \notin \sigma_{j-1}^{x+1}$.
Based on the induction hypothesis that $\sigma_j^{x-1} \subset \sigma_j^x \subset \sigma_j^{x+1}$, there exist items $u,v \notin \sigma_j^{x-1}$ such that
\begin{align}\label{eq:m-1+s+j-m+1_gen}
\sigma_j^x = \sigma_j^{x-1} \cup \{u\}, \,\,\, \sigma_j^{x+1} = \sigma_j^x \cup \{ v\}.
\end{align}
Since we have assumed that $j-1 \notin \sigma_{j-1}^{x+1}$,
we must have $\sigma_{j-1}^{x+1}=\sigma_{j}^{x+1}$ and both have products $u$ and $v$.
Suppose that in $\sigma_{j-1}^{x+1}$ or $\sigma_{j}^{x+1}$ there are $x_u$ and $x_{v}$ items placed after items $u$ and $v$, respectively.
Recall that $H_{u+1}^{x_u}$ and $H_{v+1}^{x_{v}}$ are the expected revenue obtained from items after $u$ and $v$, respectively.
Since $\sigma_{j-1}^{x+1}$ is optimal, the expected revenue strictly decreases if one replaces either $u$ or $v$ in $\sigma_{j-1}^{x+1}$ by item $j-1$.
Conditional on the event that the customer views product $u$ (or $j-1$), we have
\begin{equation*}
\begin{split}
\lambda_{j-1} r_{j-1} + c_{j-1} H_{u+1}^{x_u} & < \lambda_u r_u+ c_u  H_{u+1}^{x_u}, \\
\lambda_{j-1} r_{j-1} + c_{j-1} H_{v+1}^{x_{v}} & <  \lambda_{v} r_{v}+ c_v H_{v+1}^{x_{v}}.
\end{split}
\end{equation*}
We express the above inequalities as
\begin{equation*}
\begin{split}
(1 - c_{j-1}) \frac{ \lambda_{j-1} r_{j-1} }{1 - c_{j-1}} + c_{j-1} H_{u+1}^{x_u} &  < (1 - c_{u})  \frac{  \lambda_u r_u }{1- c_u}+ c_u  H_{u+1}^{x_u}, \\
(1 - c_{j-1})  \frac{ \lambda_{j-1} r_{j-1} }{1-c_{j-1}}+ c_{j-1} H_{v+1}^{x_{v}} &<  (1 - c_{v}) \frac{ \lambda_{v} r_{v} }{1-c_v}+ c_v H_{v+1}^{x_{v}}.
\end{split}
\end{equation*}
In this way, the left-hand side of the first inequality can be viewed as the convex combination of $\frac{r_{j-1} }{1 - c_{j-1}}$ and $H_{u+1}^{x_u}$, and the right-hand side of the first inequality is the convex combination of $\frac{  \lambda_u r_u }{1- c_u}$ and $H_{u+1}^{x_u}$.
Note that by Lemma \ref{lem:reward_bound}, we have {$\frac{\lambda_{j-1} r_{j-1} }{1-c_{j-1}}\geq \frac{ \lambda_u r_u }{1-c_u} \geq \frac{ \lambda_{u+1} r_{u+1} }{1-c_{u+1}}  \geq  H_{u+1}^{n-u} \geq H_{u+1}^{x_u}$ (the first and second inequalities are because of the indexing of the products)}, which further implies that
$1-c_{j-1} < 1-c_u$. 
Similarly, we also have $1-c_{j-1} < 1-c_v$.
To summarize, we conclude
\begin{equation}\label{eq:c_order}
c_{j-1} > c_u \text{ and }c_{j-1} > c_{v}.
\end{equation}
Let $r^*:= r_{j-1}\lambda_{j-1} +c_{j-1} H_j^{x}$ be the expected revenue obtained by inserting item ${j-1}$ in the first position of ranking $\sigma_{j}^{x}$.
Since $j-1\notin\sigma_{j-1}^{x+1}$, we have $r^* < H_j^{x+1}$. We will show that $r^* \geq H_j^{x+1}$ in the rest of analysis to yield a contradiction which proves that $j-1\in\sigma_{j-1}^{x+1}$.

Recall that $H_{j-1}^x = r_{j-1}\lambda_{j-1} + c_{j-1} H_j^{x-1}$.
By definition, we have
\begin{equation}\label{eq:r_ast_general}
r^* = r_{j-1}\lambda_{j-1} + c_{j-1}H_j^{x} = H_{j-1}^x + c_{j-1} (H_j^x - H_j^{x-1}).
\end{equation}
As $\sigma_j^{x-1}\cup \{v\}$ is not optimal for $\sigma_j^x$, we also have
\begin{align*}
R(\sigma_j^{x-1}\cup \{v\}, x) \leq R(\sigma_j^{x}, x) = H_j^x.
\end{align*}
We consider two cases: $u<v$ and $u > v$.
\begin{enumerate}
\item[(i)] Suppose $u <v$.
Note that the difference $H_{j}^{x+1} - H_{j}^x$ is only caused by the products displayed after the inserted position of product $v$ (see \eqref{eq:m-1+s+j-m+1_gen}).
Therefore, we have
\begin{equation}\label{eq:case_1_gen}
\begin{split}
H_{j}^{x+1} - H_{j}^x  & =  \Pr( \text{viewing product } v \text{ in }\sigma_j^{x+1} ) \cdot (\lambda_{v} r_{v} +c_{v} H_{v+1}^{x_v}- H_{v+1}^{x_{v}})\\
& = \Pr( \text{viewing product } v \text{ in }\sigma_i^{x+1}\setminus\{u\})  \cdot c_u   \big  (\lambda_{v} r_{v} +c_{v} H_{v+1}^{x_v}- H_{v+1}^{x_{v}} \big ) \\
& =   c_u \cdot \big (R(\sigma_j^{x-1}\cup \{v\},x)  - H_j^{x-1} \big ) \\
& \leq  c_u \cdot  (H_j^x - H_j^{x-1}),
\end{split}
\end{equation}
In the first equality, we condition on the event that a customer  views product $v$ in $\sigma_j^{x+1}$.
In the second equality, because $u<v$, product $u$ is displayed before $v$ in $\sigma_{j}^{x+1}$ by Lemma~\ref{lem:fix-m-decrease-r-general}.
In the third equality, after excluding product $u$, we are comparing the revenues of $\sigma_j^{x-1}$ and $\sigma_j^{x-1}\cup \{v\}$.
The inequality due to the suboptimality of $\sigma_{j}^{x-1}\cup \{j\}$ in \eqref{eq:m-1+s+j-m+1_gen}.
By \eqref{eq:r_ast_general}, we have
\begin{align*}
r^* & =  H_{j-1}^x + c_{j-1} (H_j^x - H_j^{x-1}) > H_{j-1}^x + c_u (H_j^x - H_j^{x-1}) \\
& \geq  H_j^x + (H_j^{x+1} - H_j^x) = H_j^{x+1},
\end{align*}
where the first inequality follows from \eqref{eq:c_order}, and the second inequality follows from \eqref{eq:case_1_gen} and $H_{j-1}^x \geq H_j^x$. Therefore, we have $r^* > H_j^{x+1}$ if $u<v$.
\item[(ii)]  Suppose $v < u$.
Similar to \eqref{eq:case_1_gen}, we have
\begin{equation}\label{eq:case_2_gen}
\begin{split}
& R(\sigma_j^{x-1} \cup \{v,u\},x+1) - R(\sigma_j^{x-1} \cup \{v\},x)
=   H_j^{x+1} - R(\sigma_j^{x-1} \cup \{v\},x) \\
& =  \Pr(\text{viewing product } u \text{ in } \sigma_j^{x+1}) \cdot  \big (\lambda_u r_u - (1- c_u ) H_{u+1}^{x_u}  \big )\\
& =  c_{v} \cdot \Pr(\text{viewing product } u \text{ in } \sigma_j^{x+1}\setminus\{v\}) \cdot   \big (\lambda_u r_u - (1- c_u ) H_{u+1}^{x_u}  \big )\
\\
& =  c_{v} \cdot \Big (R(\sigma_j^{x-1} \cup \{u\},x) - H_j^{x-1} \Big) \\
& =  c_{v} \cdot (H_j^x - H_j^{x-1}).
\end{split}
\end{equation}
Again, by \eqref{eq:r_ast_general}, we have
\begin{align*}
r^* & =  H_{j-1}^x + c_{j-1} (H_j^x - H_j^{x-1}) \\
& >   H_{j-1}^x +c_{v} (H_j^x - H_j^{x-1}) \\
& =  H_{j-1}^x + H_j^{x+1} - R(\sigma_j^{x-1} \cup \{v\},x) \ge H_j^{x+1},
\end{align*}
where the first inequality follows from $c_{j-1} > c_v$ in \eqref{eq:c_order}, the second equality is by \eqref{eq:case_2_gen}, and the third inequality follows from $R(\sigma_j^{x-1} \cup \{v\},x)  \leq H_{j-1}^x$ (suboptimality of $\sigma_j^{x-1} \cup \{v\}$ for $\sigma_{j-1}^x$). We conclude that $r^* > H_j^{x+1}$ also holds if $v < u$.
\end{enumerate}
Combining both cases, we have $r^* > H_j^{x+1}$, which yields a contradiction. Therefore, we conclude that $j-1 \in \sigma_{j-1}^{x+1}$.
By the previous analysis, it implies that $\sigma_{j-1}^x\subset \sigma_{j-1}^{x+1}$ and thus $\sigma_{j-1}^{1}  \subset \sigma_{j-1}^{2} \subset \cdots \subset \sigma_{j-1}^{n-j+1}$.
To see $\sigma_{j-1}^{n-j+1} \subset \sigma_{j-1}^{n-j+2}$, note that $\sigma_{j-1}^{n-j+2} = \{j-1,j,\cdots,n\}$ is the full collection of products and the inclusion holds automatically.
This completes the proof.

\QED

Finally, we provide a generalization of Lemma~\ref{lemma:decreased_margin}.
\begin{lemma}\label{lemma:decreased_margin_general}
Given $j\in\{1,\dots, n-2\}$, if $\sigma_j^1 \subset \sigma_j^2 \subset \dots \subset \sigma_j^{n-j+1}$, then we have
$$
H_j^{x+2} - H_j^{x+1} \leq H_j^{x+1} - H_j^{x},
$$
for all $x = 1,2,\ldots, n-j-1$.
\end{lemma}
\textit{Proof of Lemma \ref{lemma:decreased_margin_general}:}
Due to the condition of the lemma, we have $\sigma_j^x \subset \sigma_{j}^{x+1} \subset \sigma_j^{x+2} $.
Let $u$ and $v$ be the two items added to the optimal ranking for attention span $x+1$ and $x+2$ involving products $\{j,j+1,\ldots, n\}$. That is,
$$
\sigma_{j}^{x+1} = \sigma_j^{x} \cup \{u\}\text{ and } \sigma_{j}^{x+2} = \sigma_j^{x+1} \cup \{v\}.
$$
Again, we treat a ranking as a set without ambiguity because of Lemma~\ref{lem:fix-m-decrease-r-general}.
Suppose on the contrary the result does not hold, that is,
\begin{equation}\label{eq:reverse_ineq_gen}
H_j^{x+2} - H_j^{x+1} > H_j^{x+1} - H_j^{x}.
\end{equation}
We consider the two scenarios $u < v$ and $u > v$ separately.
Also recall that $H_j^k = R(\sigma_j^k, k)$ for $k=x,x+1,x+2$.

\begin{itemize}
    \item
Suppose $u < v$ and let $x_v$ be the number of items displayed in $\sigma_j^{x+2}$ after item $v$.
By Lemma~\ref{lem:fix-m-decrease-r-general}, the products displayed after item $v$ are chosen from $\{v+1,v+2,\ldots, n\}$.
Moreover, they must generate the optimal revenue conditional on the event that the customer does not leave after viewing product $v$.
Therefore, the sub-ranking after $v$ is $\sigma_{v+1}^{x_v}$ and the conditional expected revenue is $H_{v+1}^{x_v}$.

Let $P_1$ be the probability that a customer views product $v$ in the ranking $\sigma_j^{x+2}$.
Suppose product $v$ is inserted in the $i_v$-th position of $\sigma_j^{x+2}$, i.e., $\sigma_j^{x+2}(i_v) = v$.
Due to the fact that $\sigma_j^{x+2} = \sigma_j^{x+1} \cup \{v\}$ and $v$ is inserted after $u$ (because $u<v$ and Lemma~\ref{lem:fix-m-decrease-r-general}), the probability of viewing the $i_v$-th item in $\sigma_{j}^{x+1}$ is also $P_1$.
The expected revenues generated before the $i_v$-th product are identical in $\sigma_{j}^{x+1}$ and $\sigma_{j}^{x+2}$.
Thus, we have
\begin{equation}\label{eq:hm+2-hm+1_gen}
H_j^{x+2}- H_j^{x+1}= P_1 \cdot \Big ( \lambda_v r_v  + c_v H_{v+1}^{x_v} - H_{v+1}^{x_v}\Big ) = P_1 \cdot \Big ( \lambda_v r_v  - (1-c_v) H_{v+1}^{x_v}\Big ).
\end{equation}
Moreover, combining \eqref{eq:hm+2-hm+1_gen} with Lemma~\ref{lemma:rev_increase-general}, we can see that $\lambda_v r_v  - (1-c_v) H_{v+1}^{x_v}$ because $P_1>0$.

Let $\tilde \sigma_j^{x+1} = \sigma_j^x \cup \{v\}$.
Note that $\tilde \sigma_j^{x+1}$ includes $x+1$ products and it is not the optimal ranking.
But we arrange the product in the ascending order of their indices as Lemma~\ref{lem:fix-m-decrease-r}.
Let $\tilde P_1$ be the probability of viewing product $v$ in the ranking $\tilde \sigma_j^{x+1}$ for a customer of attention span $x+1$.
Comparing $\tilde \sigma_j^{x+1}$ and $\sigma_j^{x+2}$, the product $u$ is inserted to $\tilde \sigma_j^{x+1}$ to turn it into  $\sigma_i^{x+2}$.
Because $u < v$, we have $P_1 = c_u \cdot \tilde P_1\le \tilde P_1$.
Note that the expected revenue conditional on the event that a customer doesn't purchase after viewing product $v$ in $\tilde \sigma_j^{x+1}$ is still $H_{v+1}^{x_v}$ as the sub-ranking after $v$ remains the same as $\sigma_j^{x+2}$.

Now consider the expected revenue of $\tilde \sigma_j^{x+1}$ and $\sigma_j^x$, which removes product $v$ in the former ranking.
We have
\begin{align*}
 R(\sigma_i^{x} \cup \{v\},x+1) - H_i^x & = \tilde P_1 \Big ( \lambda_v r_v + c_v H_{v+1}^{x_v} -H_{v+1}^{x_v}\Big )= \tilde P_1 \Big ( \lambda_v r_v  -(1-c_v) H_{v+1}^{x_v}\Big ) \\
 & \geq  P_1 \Big ( \lambda_v r_v  -(1-c_v) H_{v+1}^{x_v}\Big ) = H_{j}^{x+2}- H_j^{x+1} > H_{j}^{x+1}- H_j^{x},
\end{align*}
where the first inequality follows from $ P_1 \le \tilde P_1  $ and $\lambda_v r_v  - (1-c_v) H_{v+1}^{x_v} \geq 0$ from \eqref{eq:hm+2-hm+1_gen}, and the last inequality follows from \eqref{eq:reverse_ineq_gen}.
Therefore, we have $R(\sigma_j^x \cup \{v\},x+1) > H_j^{x+1})$, which contracts the optimality of $\sigma_j^{x+1}$.
\item
Next consider the case $v < u$.
Suppose there are $x_v$ items placed after $v$ in $\sigma_j^{x+2}$.
Similarly, denote the sub-ranking as $\sigma_{v+1}^{x_v}$ and the expected reward as $H_{v+1}^{x_v}$.
Define $P_0$ as the probability of viewing item $v$ under $\sigma_j^{x+2}$.
Similar to \eqref{eq:hm+2-hm+1_gen}, we have
$$
H_j^{x+2} - H_j^{x+1}= P_0 \big (\lambda_v r_v -(1-c_v) H_{v+1}^{x_v} \big ).
$$
Consider the ranking $\tilde \sigma_j^{x+1} = \sigma_j^x \cup \{v\}$.
The sub-ranking after $v$ in $\tilde \sigma_j^{x+1}$ is $\sigma_{v+1}^{x_v}\setminus\{u\}$, or equivalently $\sigma_{v+1}^{x_v -1}$.
By Lemma \ref{lemma:rev_increase-general}, we have $H_{v+1}^{x_v -1} \leq H_{v+1}^{x_v}$.
Therefore, similar to \eqref{eq:hm+2-hm+1_gen}, we have
\begin{equation*}
\begin{split}
 \quad R(\sigma_j^x \cup \{v\},x+1) - H_j^x &  =  P_0 \big  (\lambda_v r_v - (1-c_v)  H_{v+1}^{x_v -1}  \big ) \\
 & \geq P_0 \big  (\lambda_v r_v - (1-c_v)  H_{v+1}^{x_v}  \big ) = H_j^{x+2} - H_j^{x+1}> H_j^{x+1} - H_j^{x},\end{split}
\end{equation*}
which implies that $R(\sigma_j^x \cup \{ v\},x+1) > H_j^{x+1}$ and contradicts the $\cL$-optimality of $\sigma_j^{x+1}$.
\end{itemize}
Combining both cases, we conclude $H_j^{x+2} - H_j^{x+1} \leq H_j^{x+1} - H_j^{x} $, which completes the proof.

\QED

Following these results, we can obtain the generalization of Proposition~\ref{prop:nested} and Theorem~\ref{thm:decrease_margin} in this setting following the proofs in Section~\ref{sec:proofs-main}.


\bibliographystyle{chicago}
\bibliography{ref.bib}

\end{appendices}

\end{document}